\documentclass[journal,twoside]{IEEEtran}
\usepackage{amsfonts,bm,booktabs,cite,graphicx,hyperref,multirow,textcomp,url}
\usepackage{algorithm}
\usepackage{subfigure}
\usepackage{algpseudocode}
\usepackage{amsmath}
\usepackage{mathrsfs}
\usepackage{float}
\usepackage{color}
\usepackage[OT1]{fontenc}

\makeatletter 
\renewcommand{\@thesubfigure}{\hskip\subfiglabelskip}
\newcommand{\etal}{{\emph{et al.~}}\@ }
 \makeatother
\begin{document}
\title{Attribute Artifacts Removal for Geometry-based Point Cloud Compression}
\author{
	Xihua Sheng, 
	Li Li, \IEEEmembership{Member, IEEE},
	Dong Liu, \IEEEmembership{Senior Member, IEEE},
	Zhiwei Xiong, \IEEEmembership{Member, IEEE}
\thanks{Date of current version \today.

X. Sheng, L. Li, D. Liu, and Z. Xiong are with the CAS Key Laboratory of Technology in Geo-Spatial Information Processing and Application System, University of Science and Technology of China, Hefei 230027, China (e-mail: xhsheng@mail.ustc.edu.cn; lil1@ustc.edu.cn; dongeliu@ustc.edu.cn; zwxiong@ustc.edu.cn).\par
}
}

\markboth{}{Sheng \MakeLowercase{\textit{et al.}}: Attribute Artifacts Removal for Geometry-based Point Cloud Compression}

\maketitle
\begin{abstract}
Geometry-based point cloud compression (G-PCC) can achieve remarkable compression efficiency for point clouds.
However, it still leads to serious attribute compression artifacts, especially under low bitrate scenarios.
In this paper, we propose a Multi-Scale Graph Attention Network (MS-GAT) to remove the artifacts of point cloud attributes compressed by G-PCC.
We first construct a graph based on point cloud geometry coordinates and then use the Chebyshev graph convolutions to extract features of point cloud attributes.
Considering that one point may be correlated with points both near and far away from it, we propose a multi-scale scheme to capture the short- and long-range correlations between the current point and its neighboring and distant points.
To address the problem that various points may have different degrees of artifacts caused by adaptive quantization, we introduce the quantization step per point as an extra input to the proposed network. 
We also incorporate a weighted graph attentional layer into the network to pay special attention to the points with more attribute artifacts.
To the best of our knowledge, this is the first attribute artifacts removal method for G-PCC.
We validate the effectiveness of our method over various point clouds.
Objective comparison results show that our proposed method achieves an average of 9.74\% BD-rate reduction compared with Predlift and 10.13\% BD-rate reduction compared with RAHT. Subjective comparison results present that visual artifacts such as color shifting, blurring, and quantization noise are reduced.
\end{abstract}
\begin{IEEEkeywords}
Compression artifacts removal, geometry-based point cloud compression, graph attention network, graph convolution, point cloud attribute compression.
\end{IEEEkeywords}
\IEEEpeerreviewmaketitle

\section{Introduction}
Point clouds are sets of 3D discrete points with geometry coordinates and associated attributes such as RGB colors, normals, or reflectances~\cite{schwarz2018emerging}. Without loss of generality, we focus on the color attributes represented as (r, g, b) triplets in RGB color space, as done in existing works~\cite{de2016compression,zhang2014point,huang2008generic,mammou2017video,mammou2018lifting,mekuria2016design}. Point clouds have been widely applied in many scenarios such as 3D immersive  visual communication, cultural heritage, and virtual/augmented reality~\cite{tulvan2016use}. Meanwhile, along with the rapid development of emerging 3D sensing and capture technologies, the data volume of point clouds becomes very large. For example, in a typical point cloud sequence captured by 8i, one frame can reach one million points. As 30 and 24 bits are used to represent the geometry (x, y, z) and the color (r, g, b) of each point, the size of a point cloud frame can reach 6M bytes. For a sequence of 30 frames per second, the bit rate can be as high as 180M bytes per second without compression~\cite{li2019advanced}. Therefore, there is an urgent need to develop some high-efficient point cloud compression methods.\par

To address this problem, the 3 Dimensional Graphics coding group (3DG) under the well-known Moving Picture Experts Group (MPEG) has developed the point cloud compression (PCC) standards, of which the geometry-based PCC (G-PCC) framework is a typical example.
Under the G-PCC framework, an octree-based algorithm~\cite{jackins1980oct,huang2008generic,schnabel2006octree} and a mesh/surface-based algorithm~\cite{anis2016compression,pavez2018dynamic} were proposed for point cloud geometry coding.
A layer-based hierarchical neighborhood prediction with lifting transform (Predlift)~\cite{mammou2017video,mammou2018lifting} and a region-adaptive hierarchical transform (RAHT)~\cite{de2016compression} were designed for point cloud attribute coding. 
Although G-PCC achieves remarkable compression efficiency, we observe that it still leads to serious attribute compression artifacts, e.g, color shifting, blurring, banding, and quantization noise.
Therefore, in this work, we focus on the point cloud attribute compression artifacts removal for G-PCC.\par

To reduce compression artifacts, post-processing methods have been extensively studied on images and videos. 
Traditional filters and sparse-coding-based methods can be used for this purpose~\cite{norkin2012hevc,list2003adaptive,wang2013adaptive,chang2013reducing,jung2012image,rothe2015efficient}. 
Convolutional neural network (CNN) based methods~\cite{dong2015compression,zhang2018dmcnn,guo2016building,chen2016trainable,cavigelli2017cas,guo2017one,guan2019mfqe,wang2020multi} also show impressive results recent years and even significantly outperform the traditional methods.
However, post-processing for point cloud compression has rarely been investigated.
Inspired by the great success of CNN-based image and video post-processing methods, we explore designing a learning-based point cloud attribute artifacts removal method for G-PCC. \par

Due to the irregular sampling of point clouds, typical convolutions which require highly regular input data formats, like image grids, are unsuitable for point clouds. 
Therefore, it is infeasible to apply existing image or video post-processing methods to point cloud attribute compression artifacts removal task. 
In the learning-based point cloud compression task, some researchers~\cite{wang2021multiscale,guarda2020adaptive,quach2019learning} voxelized the point clouds to regular grids and fed them to typical 3D CNNs. 
However, these volumetric methods result in huge memory and computational cost due to 3D convolutions.
In addition, some researchers~\cite{quach2020folding} projected a point cloud into collections of images. 
But the projecting transformation introduces unnecessary quantization artifacts. 
Recently, graph neural network (GNN) has shown great potential in point cloud classification and segmentation tasks~\cite{te2018rgcnn,wang2018local}. It provides structure-adaptive, accurate, and compact representations for point clouds, which is crucial to characterize the underlying topology of point clouds.
Inspired by the natural advantages of GNN for point cloud processing, we explore applying it to the point cloud attribute compression artifacts removal task. \par

\begin{figure*}
  \centering
  \includegraphics[width=\linewidth]{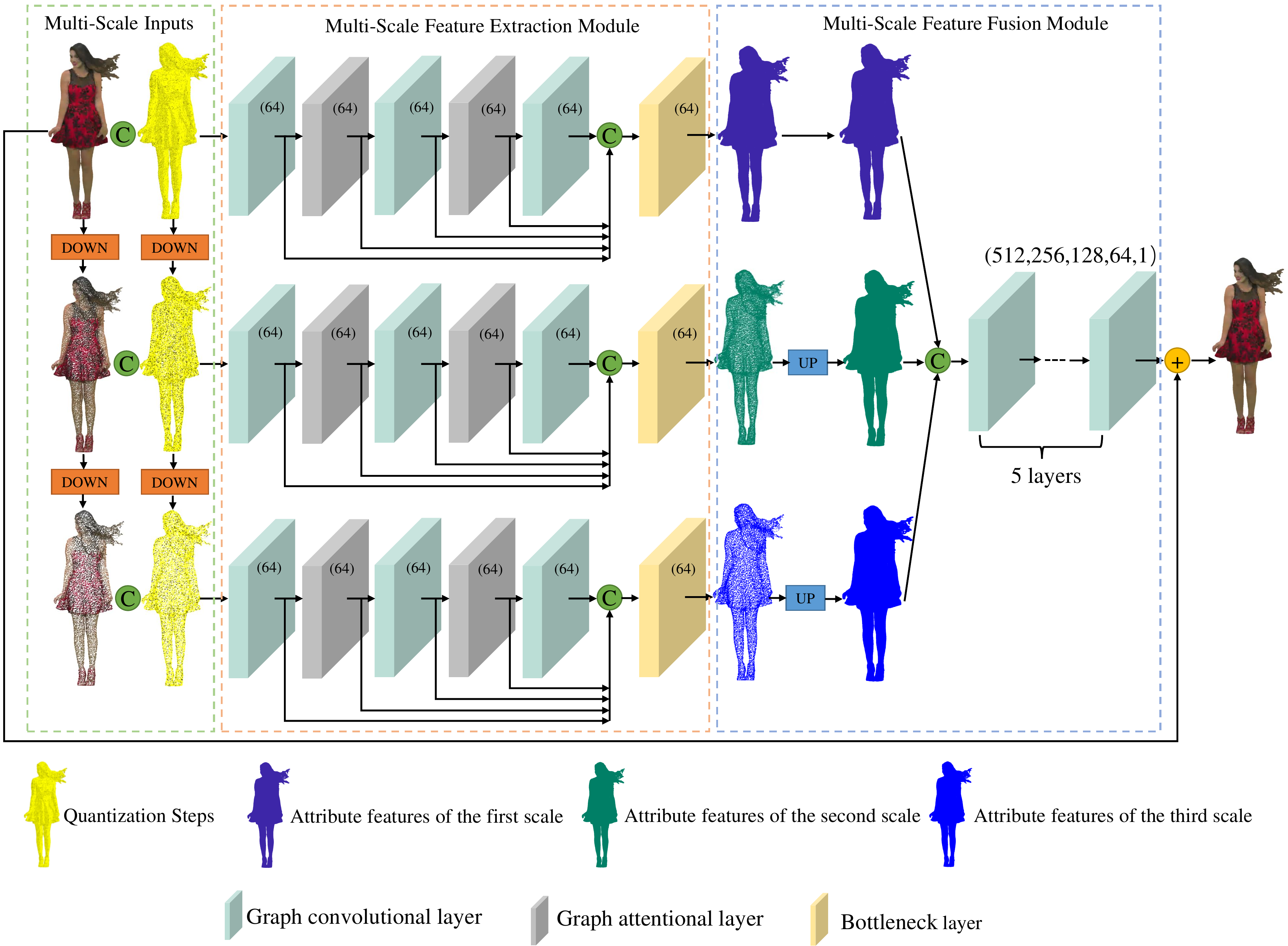}
  \caption{Architecture of our proposed Multi-Scale Graph Attention Network (MS-GAT). ``DOWN'' refers to the farthest point sampling. ``UP'' indicates a geometry distance-based weighted interpolation~\cite{qi2017pointnet++}.  ``C'' and ``+'' indicate concatenation and sum of attribute features of the points with the same coordinates, respectively. The numbers in the parenthesis denote the number of channels of the output features.}
  \label{fig:network}
\end{figure*}

Specifically, we propose a Multi-Scale Graph Attention Network (MS-GAT) to remove the artifacts of point cloud attributes compressed by G-PCC. The architecture of MS-GAT is illustrated in Fig.~\ref{fig:network}. 
We assume that point cloud geometry coordinates can be obtained in the decoder side and are used as auxiliary information to build a graph. 
The compressed point cloud attributes are treated as the graph signals on the vertices of the graph.
We use Chebyshev graph convolutions to extract the features of point cloud attributes. 
Considering that one point may be predicted by the points with both near and far distances in the point cloud attribute predictive coding, we propose a multi-scale scheme to capture the short- and long-range correlations between the current point and near or distant points.
In addition, we observe that various points may have different degrees of artifacts due to adaptive quantization.
To fully utilize this feature, we introduce the quantization steps as extra inputs to guide the network to handle different degrees of compression artifacts. 
Furthermore, we design a weighted graph attentional layer, in which the quantization steps are added into the self-attention, to pay more attention to the points with more artifacts.\par

Our contributions in this paper are summarized as follows:
\begin{itemize}
\item We propose a GNN-based method to remove the artifacts of point cloud attributes compressed by G-PCC. To the best of our knowledge, this is the first exploration of point cloud attribute artifacts removal for G-PCC.
\item We propose a multi-scale scheme to capture the short- and long-range correlations between the current point and near or distant points.
\item We use the quantization step per point as an extra input to guide the network to handle different degrees of artifacts of each point.
\item We design a weighted graph attentional layer, in which the quantization steps are added into the self-attention, to pay more attention to the points with more artifacts.
\end{itemize}
Experimental results show that our proposed MS-GAT leads to significant performance improvement on point cloud attribute compression compared with Predlift and RAHT in the TMC13v12.\par
The remainder of this paper is organized as follows. 
Section~\ref{Relatedwork} gives a brief review of related work. 
Section~\ref{MS-GAT} describes the architecture of our proposed MS-GAT. 
Section ~\ref{experiments} shows the detailed experimental results and analysis. 
Section ~\ref{conclusion} concludes this paper.

\section{Related Work}\label{Relatedwork}

In this section, we introduce the related work from the following three aspects: point cloud attributes compression, compression artifacts removal, and graph neural networks for point clouds.

\subsection{Point Cloud Attributes Compression}
Point cloud attributes compression reports can be roughly divided into four groups: distance-based algorithms~\cite{mammou2018lifting, mammou2017video}, transform-based algorithms~\cite{de2016compression,zhang2014point,huang2008generic}, and codec-based algorithms~\cite{mekuria2016design,graziosi2020overview,li2020efficient}.
In the first class, Mammou \etal\cite{mammou2017video} proposed a layer-based coding method. 
It re-orders the points into several layers according to the distances and used the decoded points from a coarser point cloud to predict those from a finer one.
A lifting scheme~\cite{mammou2018lifting} was further introduced to increase the coding performance. 
These algorithms have developed into a representative point cloud attribute coding method, Predlift, in G-PCC. 
In the second class, Zhang \etal\cite{zhang2014point} built a graph based on geometry coordinates and compressed attributes using a graph transform. 
However, since the graph transform needs to solve a complex eigendecomposition problem of graph Laplacian matrix, it brings a large computational cost. 
Queiroz and Chou~\cite{de2016compression} designed RAHT that resembles an adaptive variation of the Haar wavelet to compress attributes and achieved a better trade-off between rate-distortion (R-D) performance and computational burden, which is also adopted by G-PCC.  
In the last class, Mekuria \etal\cite{mekuria2016design} introduced an image-based point cloud compression scheme.
They projected a point cloud into collections of images and compressed the projected image using JPEG~\cite{wallace1992jpeg}. 
Mammou \etal\cite{mammou2017video} designed a patch-based project method to project the point clouds into a 2D video and compressed it using the state-of-the-art video compression standard High Efficiency Video Coding (HEVC)~\cite{sullivan2012overview}. The method and its improvements~\cite{schwarz2018emerging} are integrated into the video-based PCC (V-PCC) framework.
Since G-PCC is one of the state-of-the-art point cloud compression methods, we try to remove the attribute artifacts for G-PCC in this work.

\subsection{Compression Artifacts Removal}
Compression artifacts removal has been investigated for several decades, and to date still receives a lot of attention from researchers. 
Early reports on compression artifacts reduction~\cite{norkin2012hevc,list2003adaptive,wang2013adaptive,chang2013reducing,jung2012image,rothe2015efficient} mainly aimed at using traditional specially-designed filters and sparse coding.  
Norkin \etal\cite{norkin2012hevc} applied filtering along the boundaries of blocks to reduce blocking artifacts.  
Foi \etal\cite{foi2007pointwise} designed a shape-adaptive discrete cosine transform (DCT) model to reduce the blocking and ringing effects.
Sparse coding-based methods incorporate knowledge on natural images and encode them into an energy function as a prior. 
Chang \etal\cite{chang2013reducing} designed a two-step method to enhance the quality of images using a sparsity prior based on dictionary learning. \par

During the past several years, CNN-based methods~\cite{dong2015compression,zhang2018dmcnn,guo2016building,chen2016trainable,cavigelli2017cas,guo2017one,guan2019mfqe,wang2020multi,fu2019jpeg} have show their superiority. 
Dong \etal\cite{dong2015compression} proposed the first CNN-based JPEG artifacts removal method which demonstrated the effectiveness of CNN on image compression artifacts removal task. 
Guo \etal\cite{guo2016building} designed a dual-domain network, which uses the information from both pixel and DCT domains. 
Fu \etal\cite{fu2019jpeg} combined sparse coding with CNN and used dilated convolutions to handle different degrees of artifacts with a single model. 
Li \etal\cite{li2020learning} introduced the quantization maps as the input of CNN and also learned a single model to deal with a wide range of quality factors of compressed images.\par

For the artifacts removal of compressed videos, Yang \etal\cite{yang2018multi} proposed a multi-frame quality enhancement network to enhance the current frame utilizing its neighboring high-quality frames. 
Li \etal\cite{li2019deep} proposed a multi-frame in-loop filter for HEVC, which removed the artifacts of each encoded frame by leveraging its adjacent frames. Wang \etal\cite{wang2020multi} proposed a novel generative adversarial network-based on multi-level wavelet packet transform to enhance the perceptual quality of compressed videos.
Ding \etal\cite{ding2021patch} combined different spatial-temporal information to adaptively utilized adjacent patches to enhance the current patch.
Zhu \etal\cite{zhu2018convolutional} incorporated learned CNN models to 3D HEVC to reduce the artifacts of synthesized multi-view videos. \par

Although great success has been achieved in image/video compression artifacts removal, point cloud attribute compression artifacts removal has not been explored.    Point cloud attribute compression leads to different kinds of visual artifacts, e.g, color shifting, blurring, banding, and quantization noise. These artifacts are derived from the loss of high-frequency details during the quantization process. It is intuitive to use neural networks to remove the collection of different kinds of compression artifacts. However, since point clouds are not in a regular format, it is unsuitable to apply existing CNN-based image and video compression artifacts removal methods to point cloud attributes. Therefore, we design a GNN-based point cloud attribute artifacts removal method specifically for G-PCC in this work.
\subsection{Graph Neural Networks for Point Clouds}
GNNs have shown great potential in point cloud processing tasks, such as point cloud classification and segmentation tasks.
GNNs for point clouds can be divided into two classes: spatial GNNs~\cite{wang2019dynamic,li2018so,shen2018mining,zhang2019linked,lin2020convolution} and spectral GNNs~\cite{te2018rgcnn,li2018adaptive,wang2018local}. 
Most existing GNN-based methods for geometric data belong to the first class. 
These methods define graph convolutions as aggregating feature information from neighbors, which apply neural networks to every node of the graph. 
Wang~\cite{wang2019dynamic} \etal built a dynamic graph CNN (DGCNN) to extract the structure features of point clouds. 
Zhang \etal\cite{zhang2019linked} improved DGCNN by removing the transformation network and linking the hierarchical features from different layers. 
Shen \etal\cite{shen2018mining} introduced a KCNet to learn local structure information of point clouds based on kernel correlation between the neighboring points of a target point and kernel points. 
Feng \etal\cite{feng2020relation} proposed a relation graph network for 3D object detection in point clouds. 

In the second class, GNNs are employed in the spectral domain using the graph Laplacian. 
Te \etal\cite{te2018rgcnn} introduced a Regularized Graph Convolutional Neural Network (RGCNN) for point cloud segmentation using spectral graph convolution. Wang \etal\cite{wang2018local} proposed local spectral graph convolution on point clouds, that used standard non-parametric Fourier kernels and devised a recursive clustering and pooling strategy for aggregating information from clusters of nodes. 
In this paper, we explore applying spectral GNN to the point cloud attribute artifacts removal for G-PCC.

\begin{figure}
  \centering
  \includegraphics[width=\linewidth]{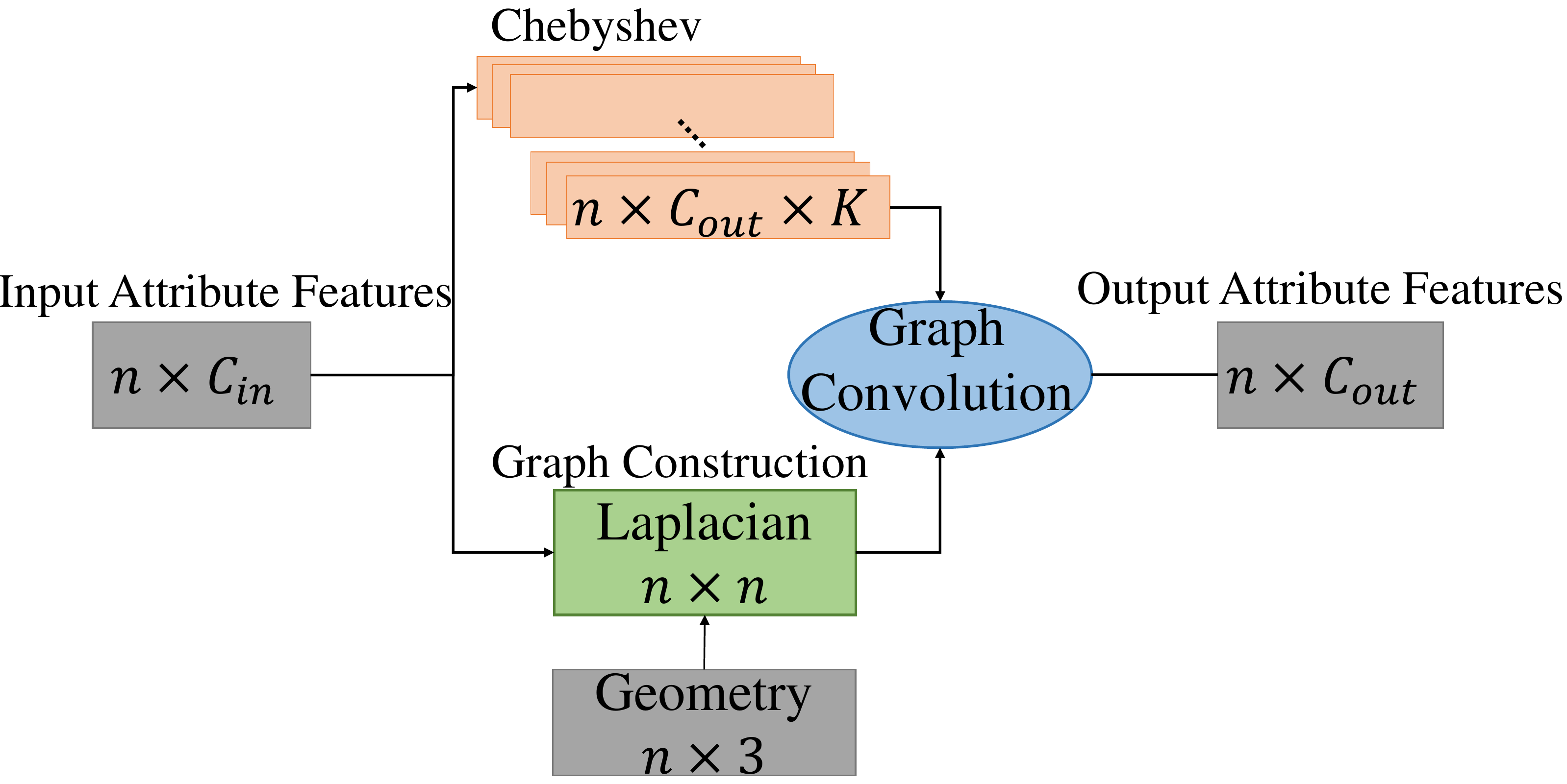}
  \caption{Architecture of the graph convolutional layer.}
  \label{fig:GCN}
\end{figure}

\section{Proposed Multi-Scale Graph Attention Network}\label{MS-GAT}
\subsection{Problem Formulation}\label{statement}
We assume that the point cloud geometry has been compressed via a certain geometry codec. 
So, it is available on the decoder side. 
Given the decoded point cloud geometry coordinates $X=\{x_{1}, x_{2}, \dots, x_{n}\}$ and the compressed point cloud attributes $Y = \{y_{1}, y_{2}, \dots, y_{n}\}$, where $x_i$ and $y_i \in \mathbb{R^{\emph{3}}}$ and $n$ is the number of points, our goal is to remove the attribute compression artifacts. 
That is to obtain restored point cloud attributes $\hat{Y}$ from $Y$ with the assist of $X$,
\begin{equation}
\label{E1}
\hat{Y}=\Psi\left(X,Y\mid \Theta\right),
\end{equation}
where $\Psi$ is our proposed MS-GAT and $\Theta$ are its trainable parameters.
The flowchart of the proposed MS-GAT is shown in Fig.~\ref{fig:network}. 
Each part of the MS-GAT is introduced in detail in the following sections.\par

\subsection{Multi-Scale Inputs of the Network}\label{multi-scale-input}
We observe that in the point cloud attribute coding of G-PCC, the points from coarser-granularity levels are used to predict those from finer-granularity ones. 
Therefore, one point may be related to points in both near or far distances. It is necessary to capture the short- and long-range correlations between the current point and near or distant points. Therefore, we downsample the point cloud using farthest point sampling (FPS)~\cite{qi2017pointnet++} to several scales. 
Point cloud attributes at different scales are all fed into different branches with the same structure. \par

In addition, in the point cloud attribute coding, some points are more influential and used more often for prediction. 
Smaller quantization errors at these points are beneficial for the overall compression performance.
Therefore, an adaptive quantization scheme is designed in G-PCC. 
The scheme allocates different quantization steps $Q=\{q_{1}, q_{2}, \dots, q_{n}\}$, $q_i \in \mathbb{R}$, for different points. 
As a result, the attributes of various points suffer different degrees of compression artifacts.
Therefore, the quantization steps $Q$ are essential priors that can be utilized for compression artifacts removal.
We feed the quantization steps into the network as side information to guide the network to handle different degrees of compression artifacts.
The illustration of quantization steps is also shown in Fig.~\ref{fig:network}. 
A point with darker color indicates a smaller quantization step is associated with it. 
In this way, the formulation of our proposed MS-GAT changes to:
\begin{equation}
\label{E2}
\hat{Y}=\Psi\left(X, Y, Q\mid \Theta\right).
\end{equation}
Note that corresponding to the downsampled point cloud attributes, the associated quantization steps are also downsampled using the same way as the compressed attributes to several scales.

\subsection{Multi-Scale Feature Extraction Module}
To capture the short- and long-range correlations between the current point and near or distant points, we design a multi-scale feature extraction module to extract attribute features of the input point cloud attributes and quantization steps from different scales.
The main components of the module are graph convolutional layers and graph attentional layers. These layers are connected in a hierarchical feature transmission way. 
We introduce these layers with more details in the following sections.\par

\subsubsection{Graph Convolutional Layer}
Due to the irregular sampling of point clouds, GNNs have the natural advantage for point cloud processing. We use a graph convolutional layer to extract the attribute features of point clouds as depicted in Fig.~\ref{fig:GCN}.
We first connect each point with all the other points in the point cloud to construct an undirected graph $\mathcal{G}=\{\mathcal{V},  \mathcal{E}\}$~\cite{te2018rgcnn}. 
$\mathcal{V}$ is the vertex set, $\mathcal{E}$ refers to the edge set, and $A$ represents a weighted adjacency matrix. 
The adjacency weight $a_{i,j}$ of $A$ is computed based on the distances between the coordinates of points:
\begin{equation}
\label{E3}
a_{i,j} = e^{-\left\|x_{i}-x_{j}\right\|_{2}^{2}}.
\end{equation}
Then, we can obtain the graph Laplacian matrix~\cite{te2018rgcnn,hu2012depth,hu2014multiresolution,shen2010edge} by $L_{c}:=D-A$. $D$ is the degree matrix, which is a diagonal matrix with $d_{i, i}=\sum_{i=1}^{n} a_{i, j}$. We normalize the graph Laplacian matrix as $L=D^{-\frac{1}{2}} L_{c} D^{-\frac{1}{2}}$.\par

We treat the point cloud attribute features and quantization steps as the graph signals on the vertices of a graph. 
The graph convolution is applied to the graph signals in the spectral domain~\cite{bruna2013spectral,kipf2016semi,defferrard2016convolutional}. 
We compute the eigenvectors of the Laplacian matrix $L$ to obtain the basis $U$.  
The spectral filtering of the graph signals $H=\{h_{1}, h_{2}, \ldots, h_{n}\}$, $h_i \in \mathbb{R}^{C_{in}}$, by $g_{\theta}$ is calculated as
\begin{equation}
\label{E4}
H^{\prime}=g_{\theta}(L) H=g_{\theta}\left(U \Lambda U^{T}\right) H=U g_{\theta}(\Lambda) U^{T} H,
\end{equation}
where $H^{\prime}=\{h^{\prime}_{1}, h^{\prime}_{2}, \ldots, h^{\prime}_{n}\}$, $h^{\prime}_i \in \mathbb{R}^{C_{out}}$ and $g_{\theta}$ is a function of the eigenvalues of Laplacian matrix $L$.
We use the truncated Chebyshev polynomials to approximate the spectral filtering to reduce the high computational complexity of computing the eigendecomposition of L and aggregate local features~\cite{kipf2016semi, defferrard2016convolutional}. 
The $K$-localized filtering operation is defined as:
\begin{equation}
\label{E5}
H^{\prime}=g_{\theta}(L) H=\sum_{k=0}^{K-1} Z_{k}(L) H\Theta_{k},
\end{equation}
where $Z_{k}(L)$ is the Chebyshev polynomial of the $k$th order.
It can be recurrently calculated by $Z_{k}(L)=2 LZ_{k-1}(L)-Z_{k-2}(L)$, where $Z_{0}(L) = 1$ and $Z_{1}(L) = L$. 
$\Theta_{k}$ refers to the $k$-th learnable Chebyshev coefficient matrix.

\subsubsection{Weighted Graph Attentional Layer}
\begin{figure}
  \centering
  \includegraphics[width=\linewidth]{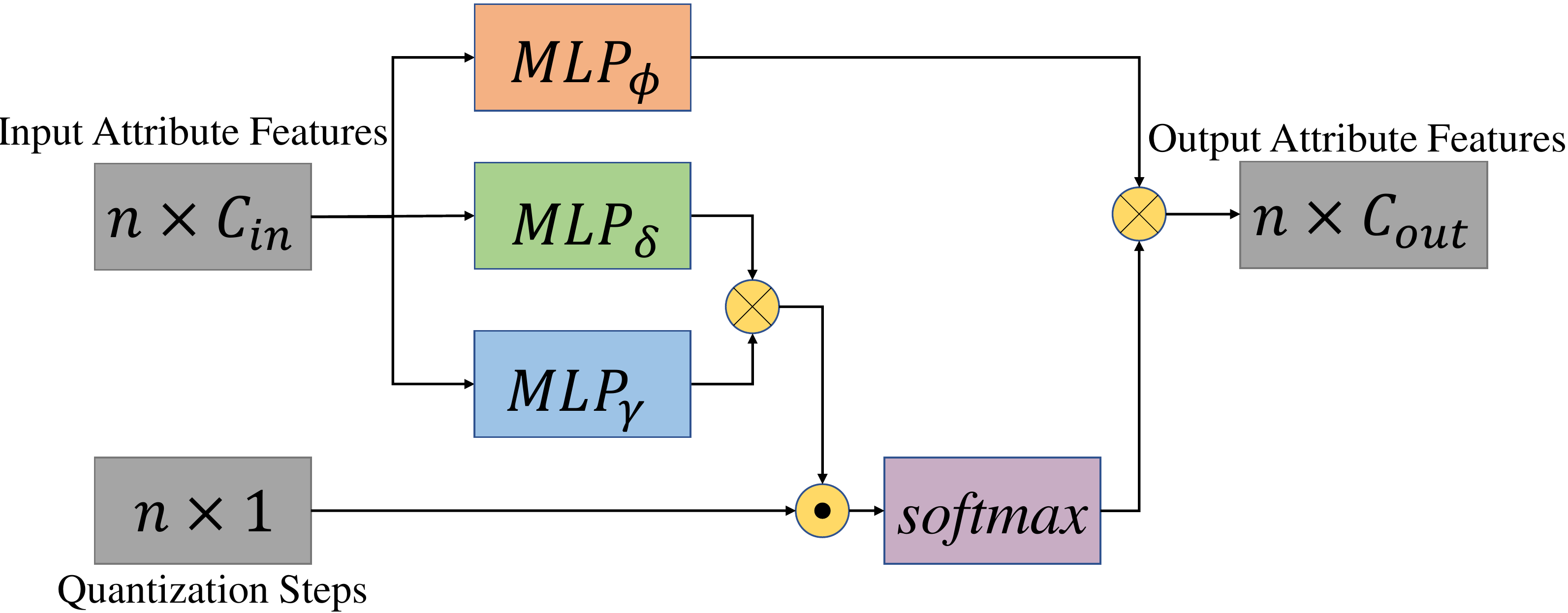}
  \caption{Architecture of the weighted graph attentional layer.}
  \label{fig:GA}
\end{figure}
Since the attributes of various points suffer different degrees of compression artifacts, the network should pay more attention to the points with more artifacts.
Inspired by the recent success of attentional layers~\cite{cheng2020learned,shan2021glow}, we design a weighted graph attentional layer to learn adaptive weights for different points, as presented in Fig.~\ref{fig:GA}. 
Given the input attribute features $H=\{h_{1}, h_{2}, \ldots, h_{n}\}$, $h_i \in \mathbb{R}^{F_{in}}$, the similarity matrix $S\in \mathbb{R}^{n\times n}$, which builds pairwise relationships for all points, is calculated as:
\begin{equation}
\label{E6}
S = {MLP_{\delta}(H)}\times {MLP^{\mathrm{T}}_\gamma(H)},
\end{equation}
where $MLP_{\delta}$ and $MLP_{\gamma}$ are $N$-layer perceptrons. Each layer is followed by the non-linear activation function ReLU. 
$(\cdot) \times(\cdot)$ is matrix multiplication. 
$MLP^{\mathrm{T}}_\gamma$ refers to the transpose of the output features of $MLP_\gamma$. \par
Considering that the quantization steps introduced in the Section~\ref{multi-scale-input} can give the network global context information to express the importance of each point, we add them into attentional layers as the global guiding weights. The global prior can help the network learn weighted attention to further focus on the points with more artifacts.
\begin{equation}
\label{E7}
M = Q\odot S,
\end{equation}
$\odot$ represents a Hadamard product by repeating $Q$ to perform element-wise multiplication as follows: 
\begin{equation}
\label{E8}
Q \odot S=\left(q_{i} \cdot s_{i j}\right)=\left(\begin{array}{ccc}
q_{1} \cdot s_{11} & \cdots & q_{1} \cdot s_{1 n} \\
\vdots & \ddots & \vdots \\
q_{n} \cdot s_{n 1} & \cdots & q_{n} \cdot s_{n n}
\end{array}\right).
\end{equation}
$M$ is then normalized with softmax function and assigned to each point. 
The output features of the weighted graph attentional layer $H^{\prime}=\{h_{1}^{\prime}, h_{2}^{\prime}, \ldots, h_{n}^{\prime}\}$, $h^{\prime}_i \in \mathbb{R}^{F_{out}}$, are computed as:
\begin{equation}
\label{E9}
H^{\prime} = softmax(M)\times{MLP_{\phi}(H)}.
\end{equation}
$MLP_{\phi}$ is also a $N$-layer perceptron. Each layer is followed by the non-linear activation function ReLU.

\subsubsection{Hierarchical Feature transmission Structure}
To avoid the disappearance  of gradients of attribute features during back-propagation, we use a hierarchical feature transmission structure similar to~\cite{li2018multi}.
We send all the output attribute features of the graph convolutional layers and graph attentional layers to the end of the feature extraction module. 
Then, we use a $N$-layer perceptron $MLP_{b}$ as a bottleneck layer to reduce the redundant information of all these attribute features and computational complexity.
The output attribute features $H_{b}=\{h_{b1}, h_{b2}, \ldots, h_{bn}\}$ are computed as:
\begin{equation}
\label{E10}
H_{b} = MLP_{b}([H_0,H_1,\dots,H_n]),
\end{equation}
where $H_i$ represents the output of the $i^{th}$ of the graph convolutional layer or weighted graph attentional layer. $[\cdot]$ refers to the concatenation of the attribute features of the points with the same  coordinates.

\subsection{Multi-Scale Feature Fusion Module}
We downsample the point clouds into several scales and extract the attribute features of point clouds with different scales in the multi-scale feature extraction module. The attribute features in different scales may contain different information. 
Therefore, we design a multi-scale feature fusion module to fuse the attribute features in all scales. 
We first upsample the downsampled point clouds to their original resolutions via the ``UP'' module in Fig.~\ref{fig:network}. The ``UP'' module is implemented by a geometry distance weighted interpolation~\cite{qi2017pointnet++}. 
We assume that the number of points of the downsampled point cloud is $n_d$ and that of the upsampled point cloud is $n$ (with $n_d < n$). The attribute features $h_{bi}^{\prime}$ of the $i^{th}$ upsampled point are calculated as follows:
\begin{equation}
\label{E11}
h_{bi}^{\prime}\left(x_{i}\right)=\frac{\sum_{j=1}^{K}\psi_{j}\left(x_{i}\right) h_{bj}}{\sum_{j=1}^{K}\psi_{j}\left(x_{i}\right)},
\end{equation}
where $\psi_j(x_i)=\frac{1}{d\left(x_i,x_j\right)^{2}}$, $i \in \{1,\cdots, n\}$, and $K$ is the number of neighboring points among the $n_d$ points in a low scale for each upsampled point. 
We set $K$ to 3 in our experiments. 
Then, all the upsampled attribute features are concatenated together and sent to five consecutive graph convolutional layers. 
Each layer is followed by a non-linear activation function ReLU except the last layer. 
To reduce the redundancy and restore the attributes, the number of feature maps of the five layers decreases gradually. 
Inspired by \cite{zhang2017beyond}, we also corporate the residual learning to simplify the learning process. 
Finally, the formulation of our proposed MS-GAT changes from~\eqref{E2} to:
\begin{equation}
\label{E12}
\hat{Y}=Y + \Psi\left(X, Y, Q\mid \Theta\right).
\end{equation}
The output of $\Psi$ is the learned residual of point cloud attributes, which contains the high-frequency attribute information.

\subsection{Loss Function}
In our work, we focus on point cloud attribute compression while the geometry coordinates of point clouds are assumed unchanged.
The attributes $Y_g$ of the original point clouds are regarded as ground truth. 
We convert the input attributes (r, g, b) into (y, u, v). Considering that the characteristics of the three components of compressed point cloud attributes are quite different, three models with the same structure are trained for each component of (y, u, v).
Each component is processed independently and identically. 
We compute the point-wise differences between each component of the restored point cloud attributes $\hat{Y}$ and the corresponding ground truth $Y_g$.
The formulation of the loss function for each component of (y, u, v) is identical and is depicted as:
\begin{equation}
\label{E13}
D\left(Y^{i}_g,\hat{Y}^{i}\right)=\sum_{j=1}^{n}\|y^{(i)}_{gj}-\hat{y}^{(i)}_{j}\|_{2}^{2},
\end{equation}
where $Y^{i}_g = \{y^{i}_{g1}, y^{i}_{g2}, \dots, y^{i}_{gn}\}$ is the set of ground truth attributes of the $i^{th}$ component of (y, u, v). 
$\hat{Y}^{i} = \{\hat{y}^{i}_{1}, \hat{y}^{i}_{2}, \dots, \hat{y}^{i}_{n}\}$ is the set of restored attributes of the $i^{th}$ component of (y, u, v). $y^{i}_{gj}$ stands for the $i^{th}$ attribute component of the $j^{th}$ point in the ground truth point cloud and $\hat{y}^{i}_{j}$ stands for the $i^{th}$ attribute component of the $j^{th}$ point in the restored point cloud.\par

\subsection{Block Partition and Combination}
\begin{figure}
  \centering
  \includegraphics[width=\linewidth]{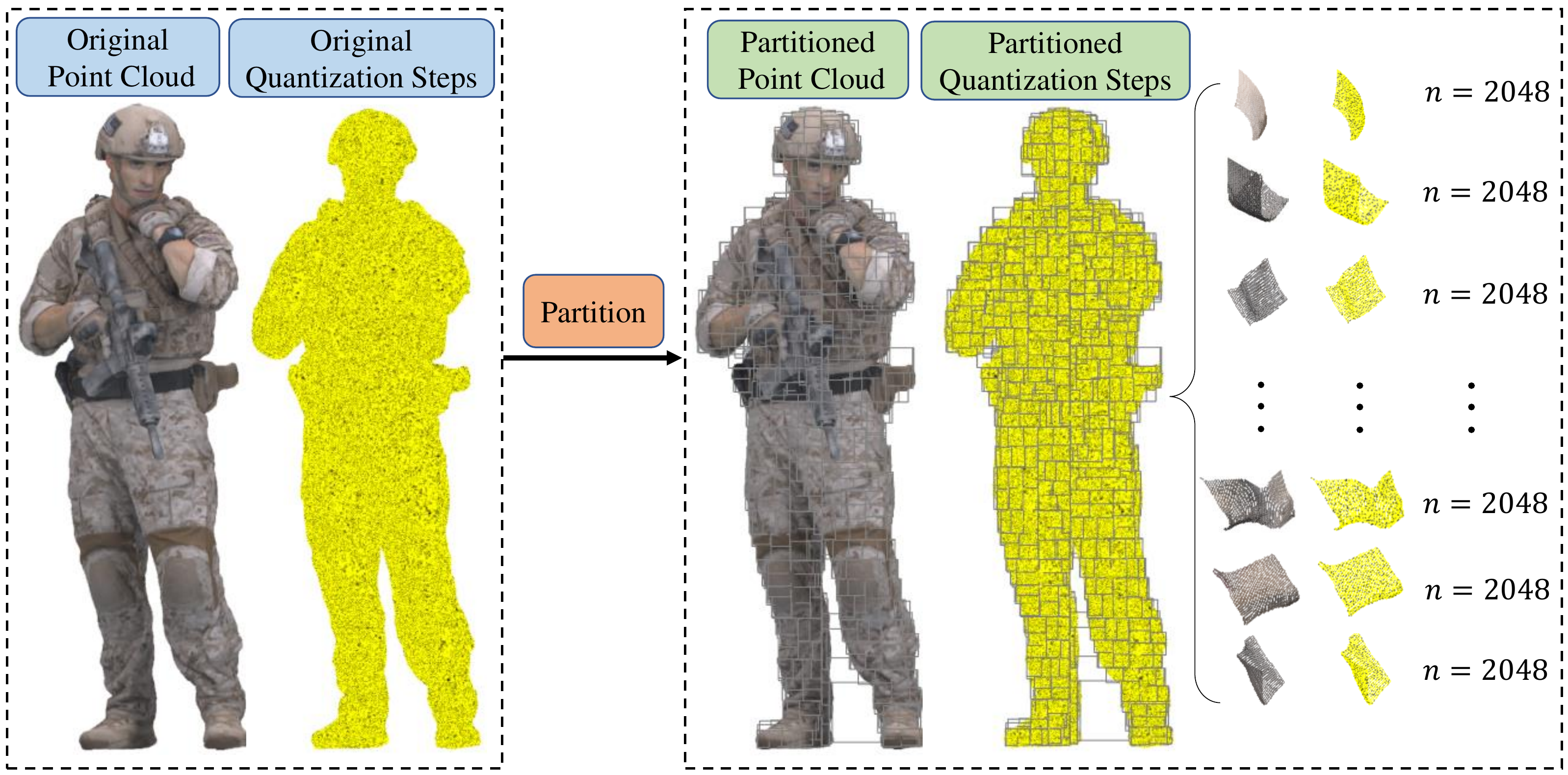}
  \caption{Partition the point clouds and quantization steps into smaller blocks. Each block contains a fixed number of points. In our experiments, the fixed number of points is set to 2048.}
  \label{fig:partition}
\end{figure}
The number of a point cloud captured by 8i can easily reach one million. 
Therefore, it is unfeasible to restore all the attributes of a complete point cloud at once due to memory constraints. To address this problem, we use a divide-and-conquer approach in the same way as~\cite{sheng2021deep}. Specifically, we partition an entire point cloud having an arbitrary number of points $N$ into local blocks with a fixed number of points $n$, as shown in Fig.~\ref{fig:partition}. The $n$ points are selected sequentially from the first point to form a local block.
Meanwhile, the quantization steps are partitioned using the same method. 
Since $N$ is usually not divisible by $n$, the number of points in the last block is smaller than $n$. Therefore, a point padding operation
is used to add a certain number of points $n_p$ ($n_p < n$) to the
last block to make $N$ divisible by $n$. Specifically, we choose to repeat the coordinates and attributes of the last point to achieve point padding as~\cite{sheng2021deep}.
After the block partition, each block is regarded as a basic unit and restored independently and identically.  
Corresponding to the block partition, the restored blocks are recombined back into a complete point cloud. Note that as
$n_p$ points are added into the original point cloud for block partition, we need to remove those added points~\cite{sheng2021deep}.\par
\begin{figure}
  \centering
  \includegraphics[width=\linewidth]{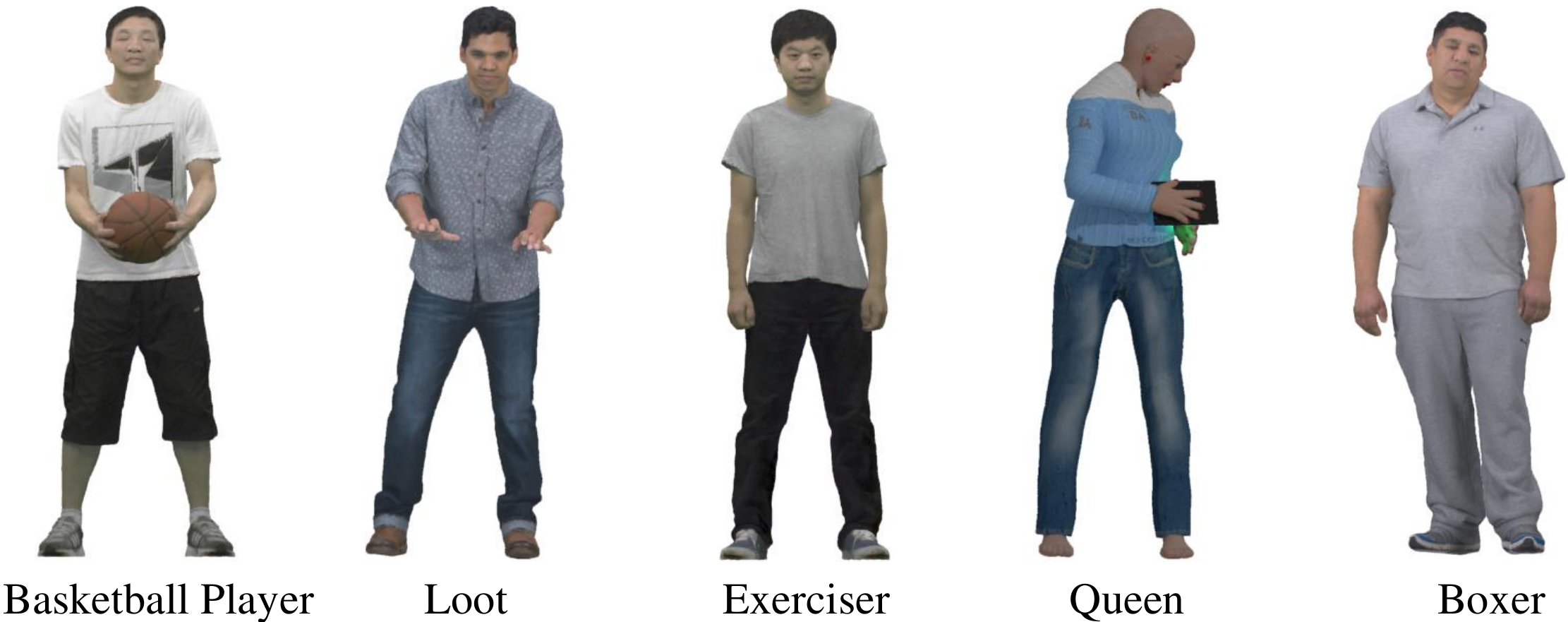}
  \caption{Training dataset includes ``Basketball Player'', ``Loot'', ``Exerciser'', ``Queen'', and ``Boxer''.}
  \label{fig:training}
\end{figure}
\begin{figure}
  \centering
  \includegraphics[width=\linewidth]{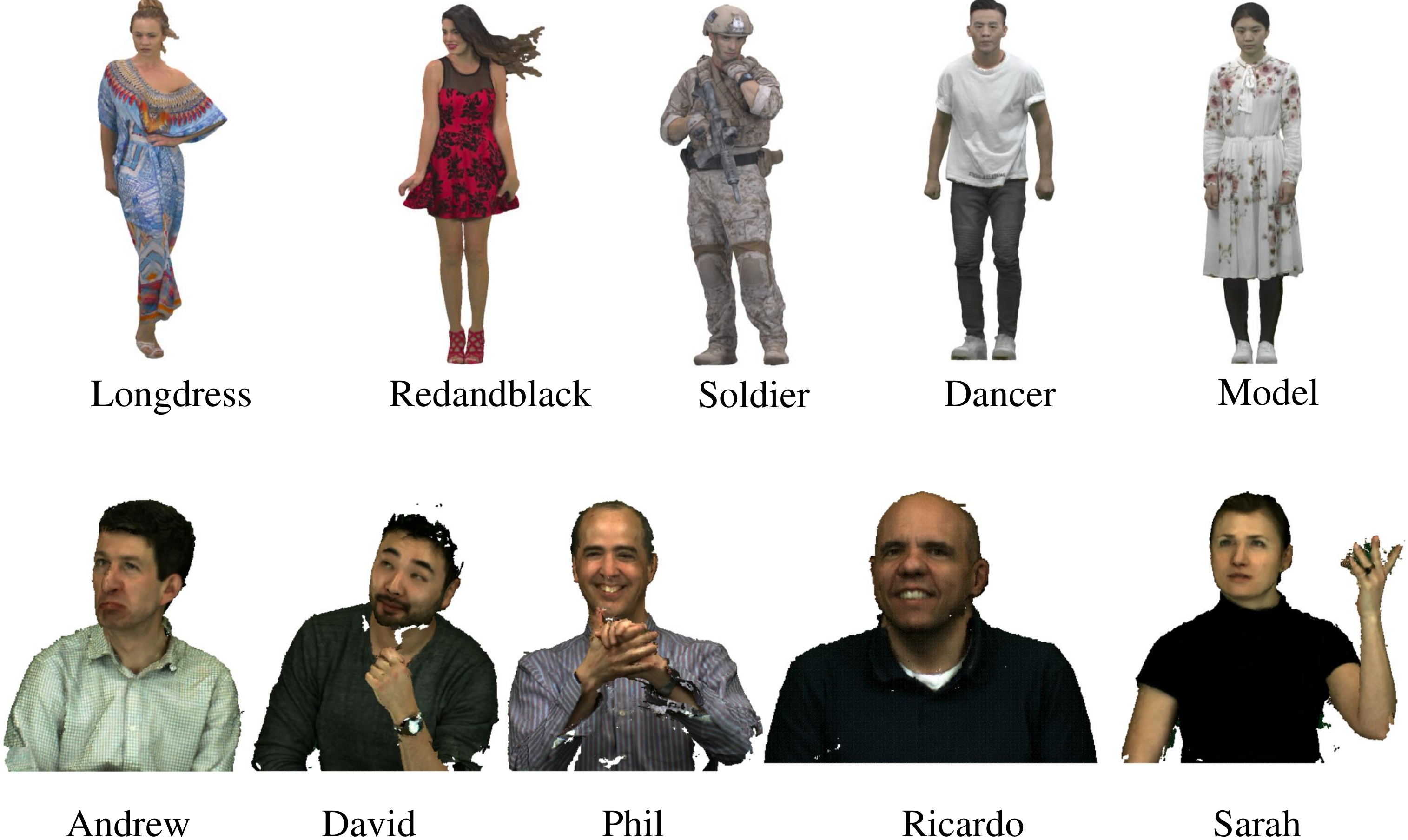}
  \caption{Test dataset includes ``Longdress'', ``Redandblack'', ``Soldier'', ``Dancer'', ``Model'', ``Andrew'', ``David'', ``Phil'', ``Ricardo'', and ``Sarah''.}
  \label{fig:testing}
\end{figure}

\begin{figure*}
\centering
  \includegraphics[width=0.234\linewidth]{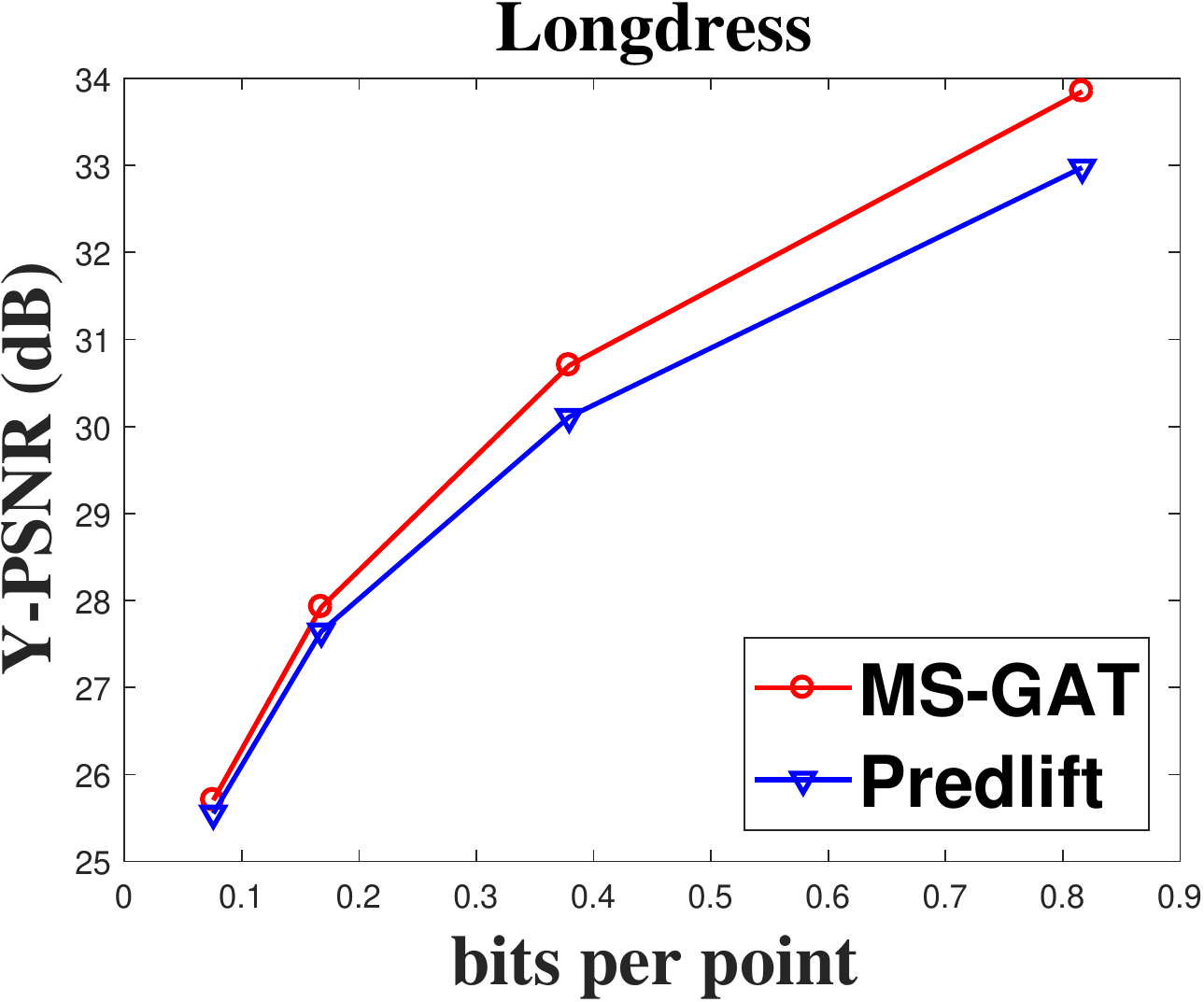}
  \includegraphics[width=0.234\linewidth]{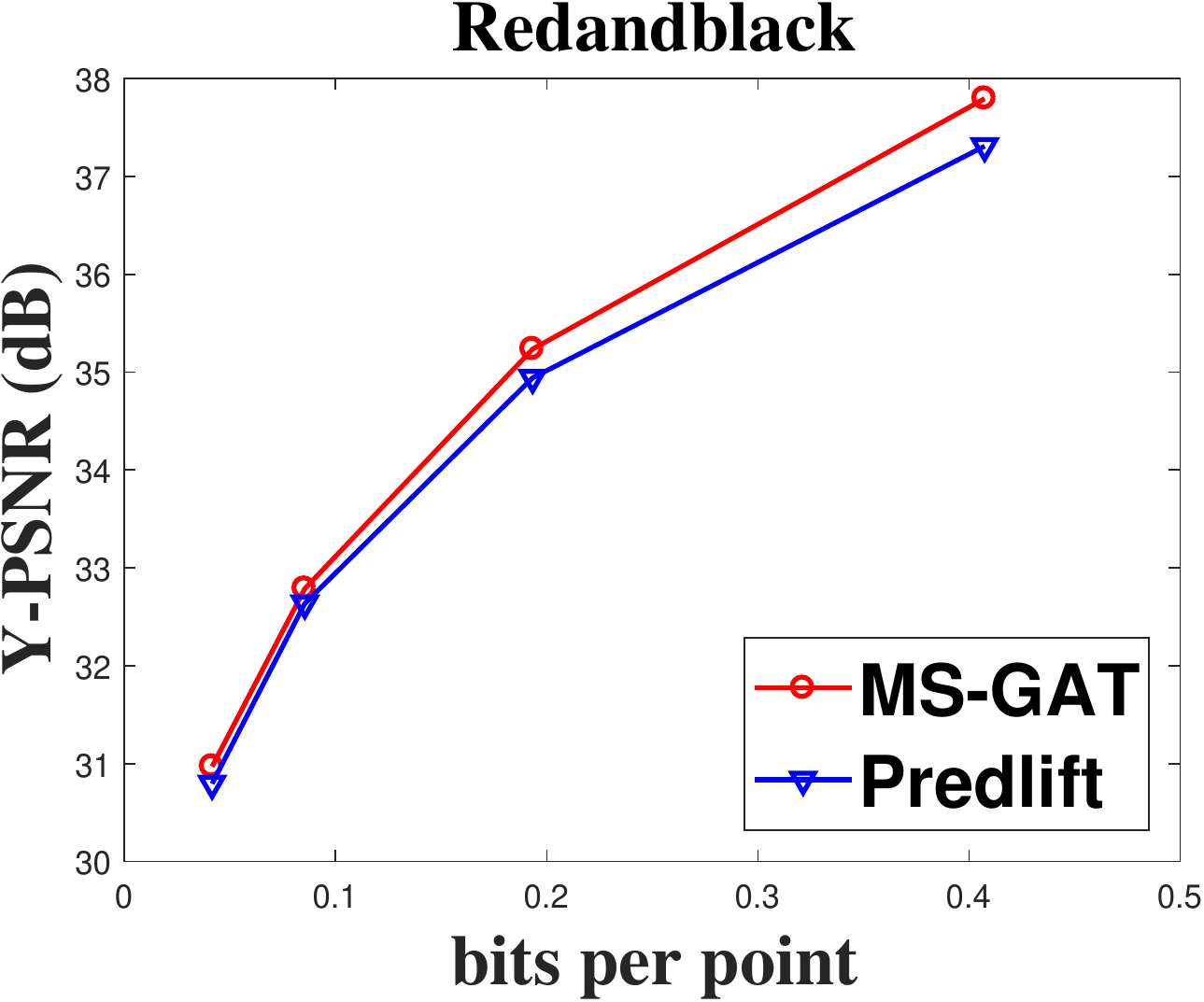}
  \includegraphics[width=0.234\linewidth]{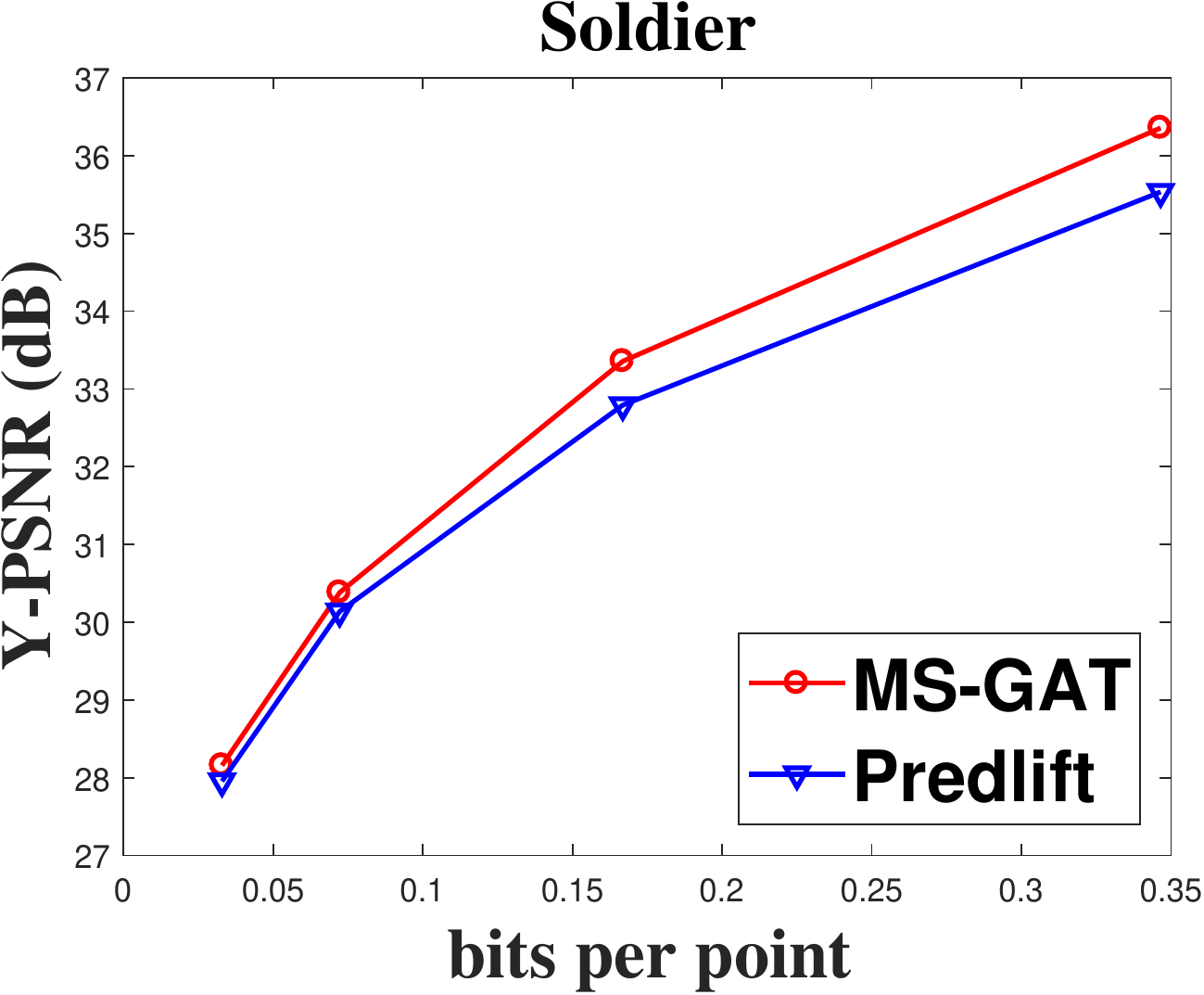}
  \includegraphics[width=0.234\linewidth]{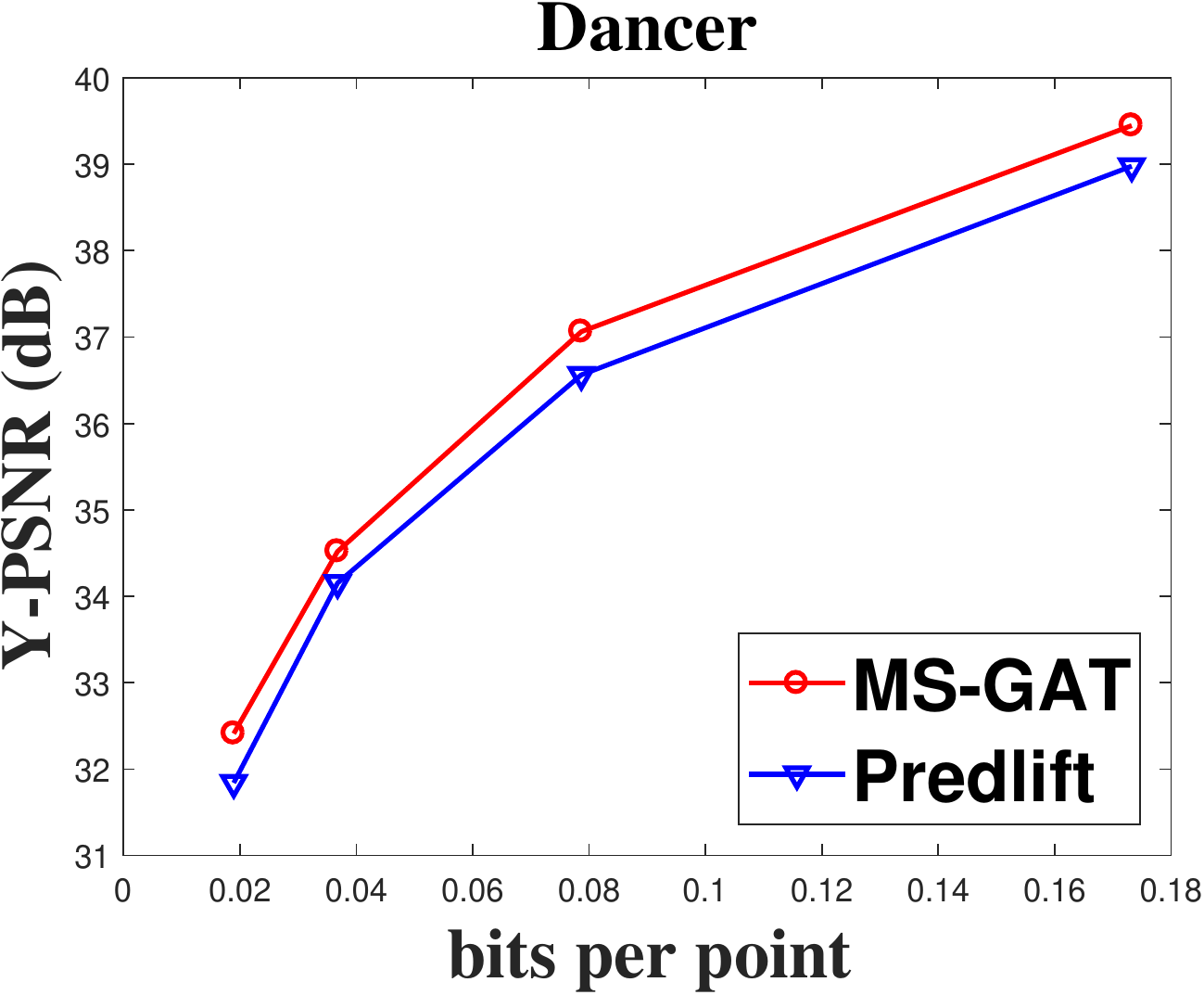}

\vspace{1mm}
  \includegraphics[width=0.234\linewidth]{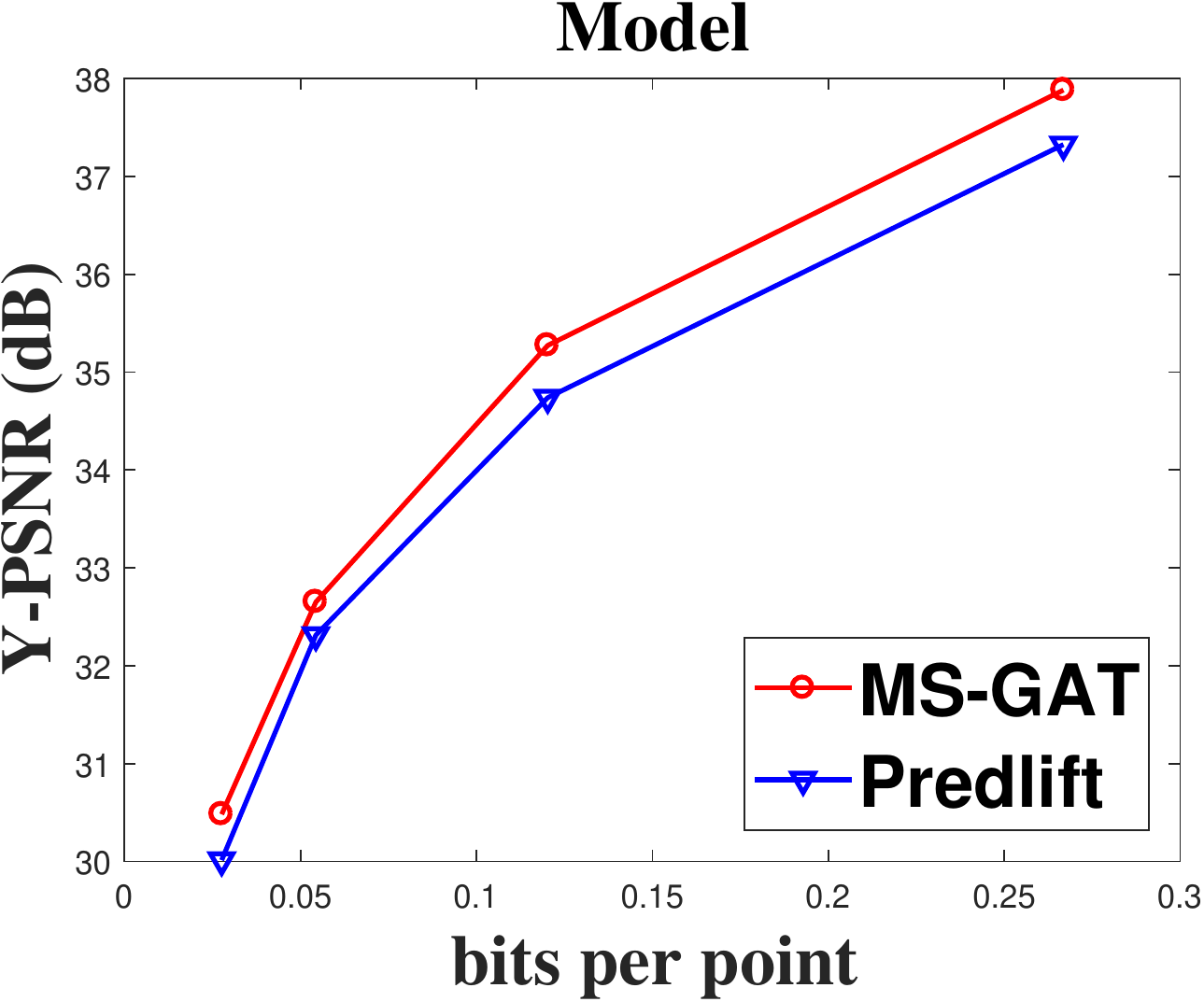}
  \includegraphics[width=0.234\linewidth]{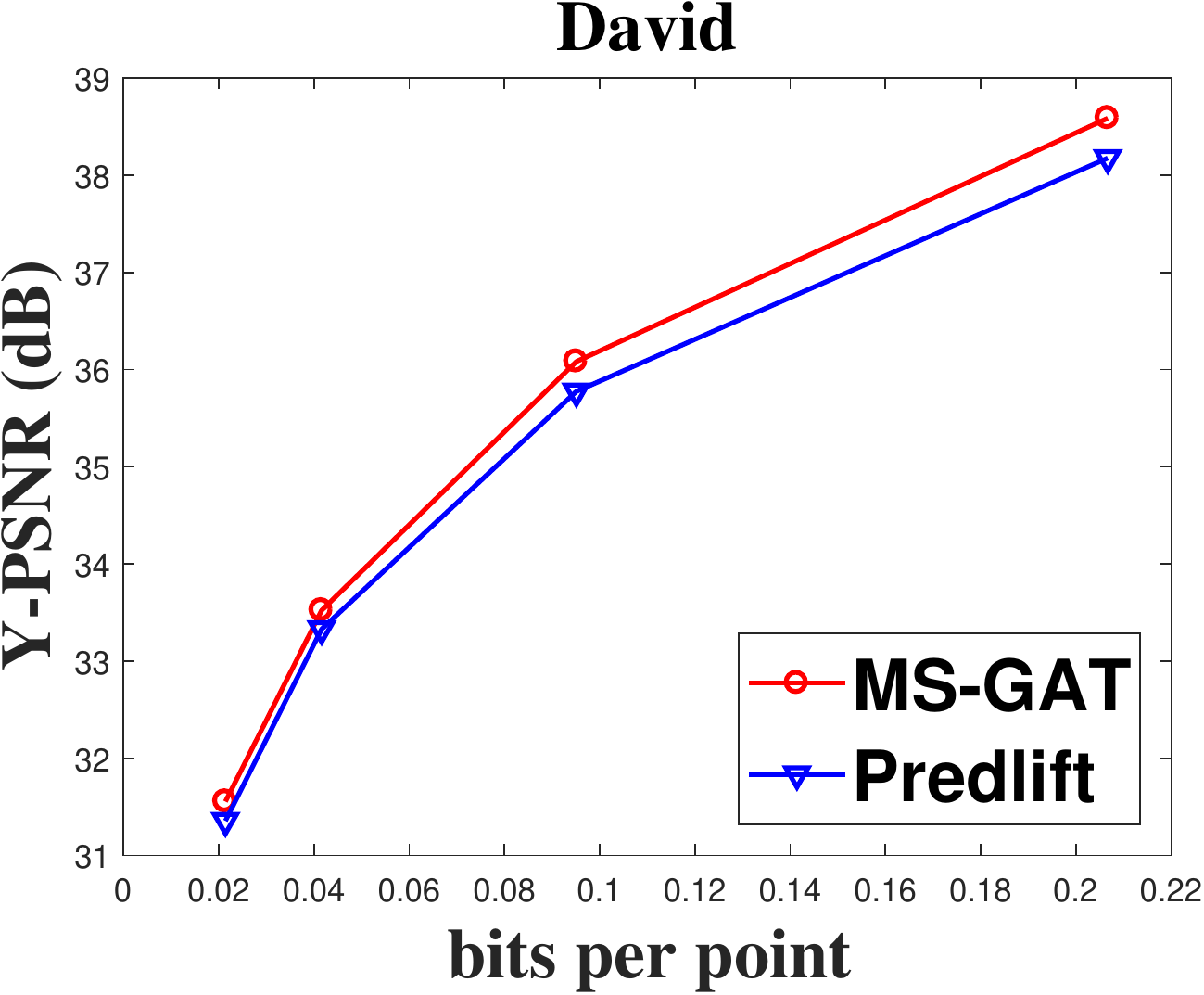}
  \includegraphics[width=0.234\linewidth]{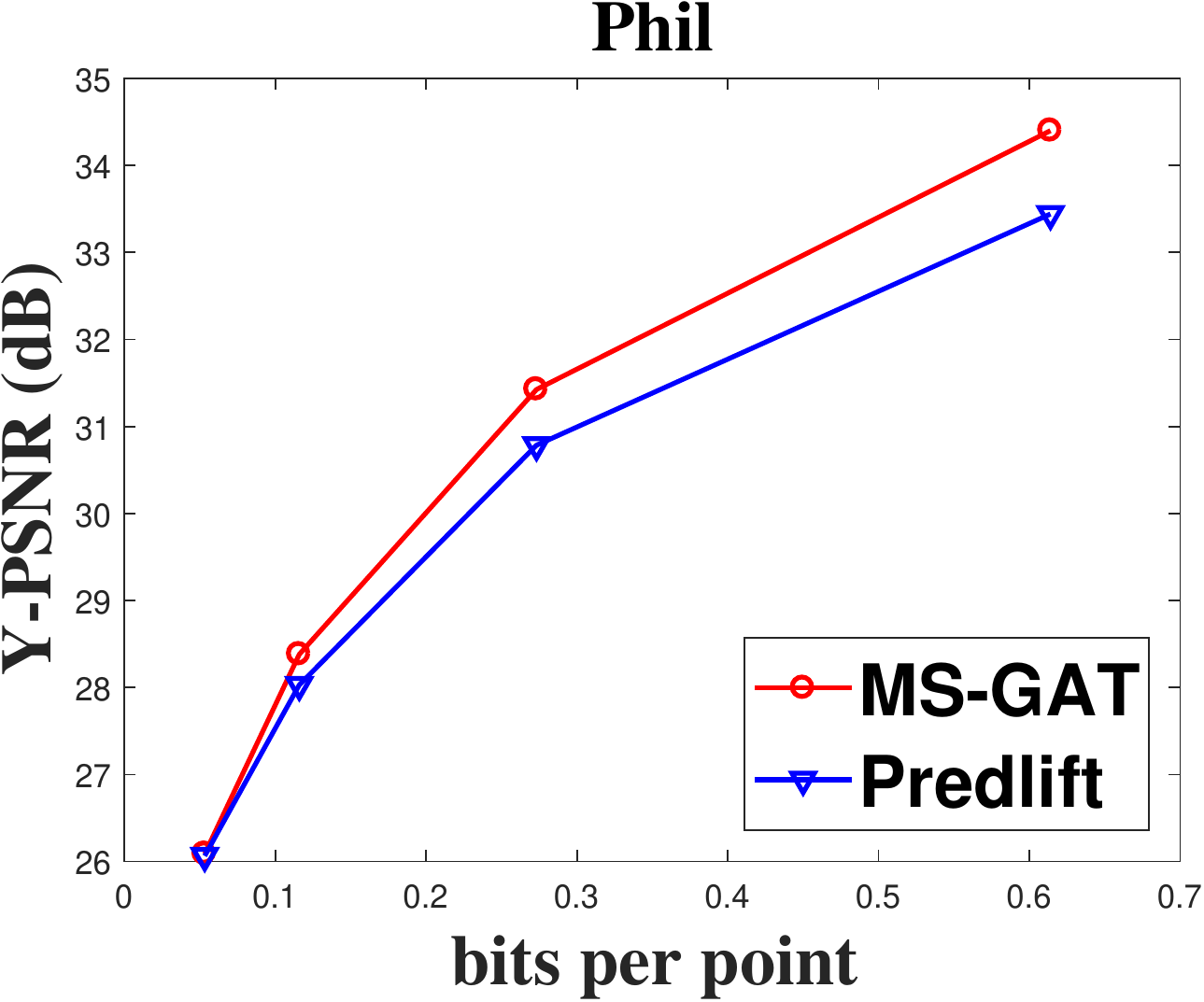}
  \includegraphics[width=0.234\linewidth]{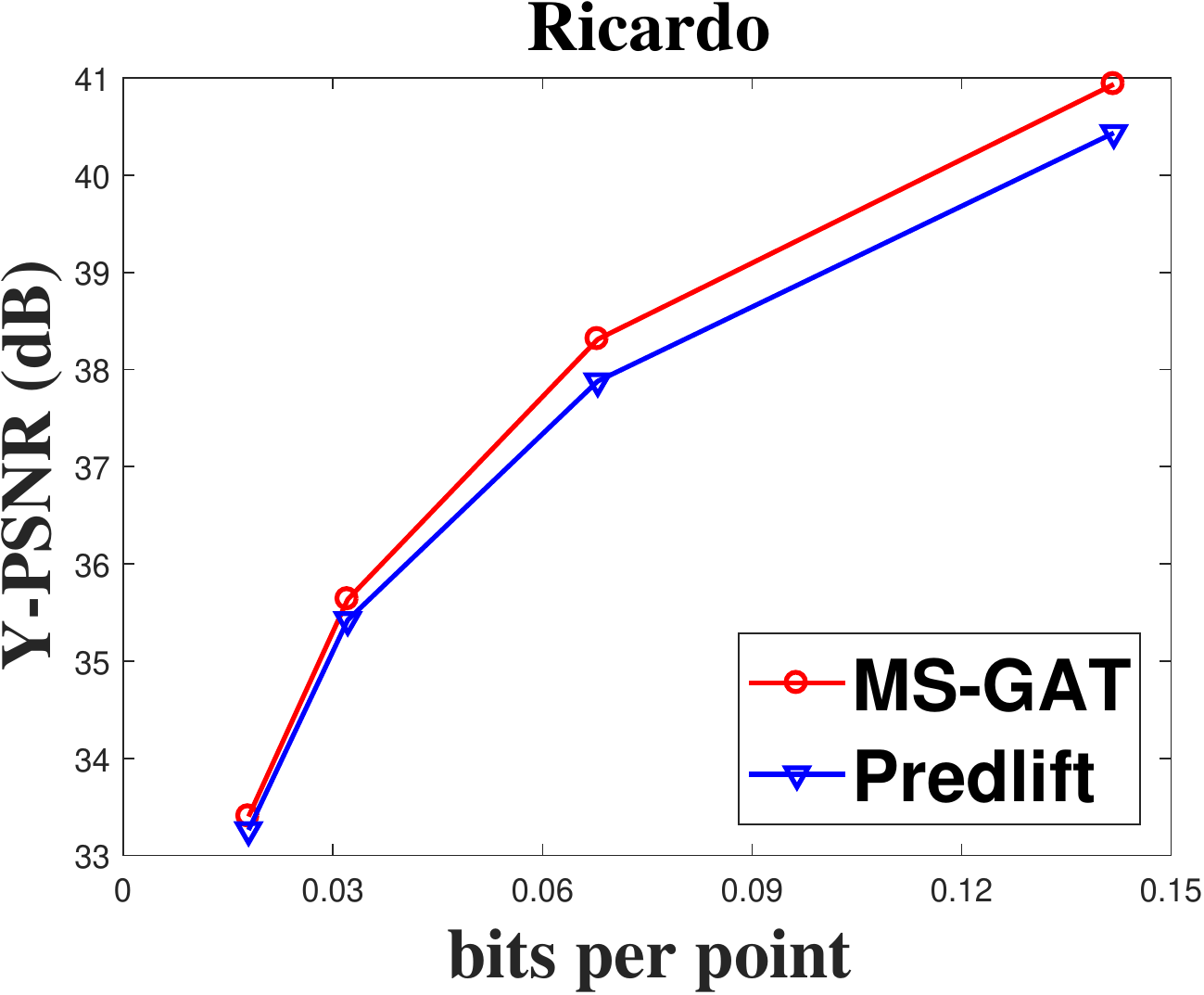}
  \caption{Rate-distortion curves for Predlift with and without our proposed algorithm on the test dataset. ``MS-GAT" denotes the rate-distortion curve of point cloud attributes restored by our proposed algorithm. ``Predlift" denotes the rate-distortion curve of Predlift-compressed attributes. The quality is measured by the peak signal-to-noise ratio (PSNR) of the Y component.}
  \label{fig:RDcurve_Y_Predlift}

\end{figure*}
\begin{figure*}
\centering
  \includegraphics[width=0.234\linewidth]{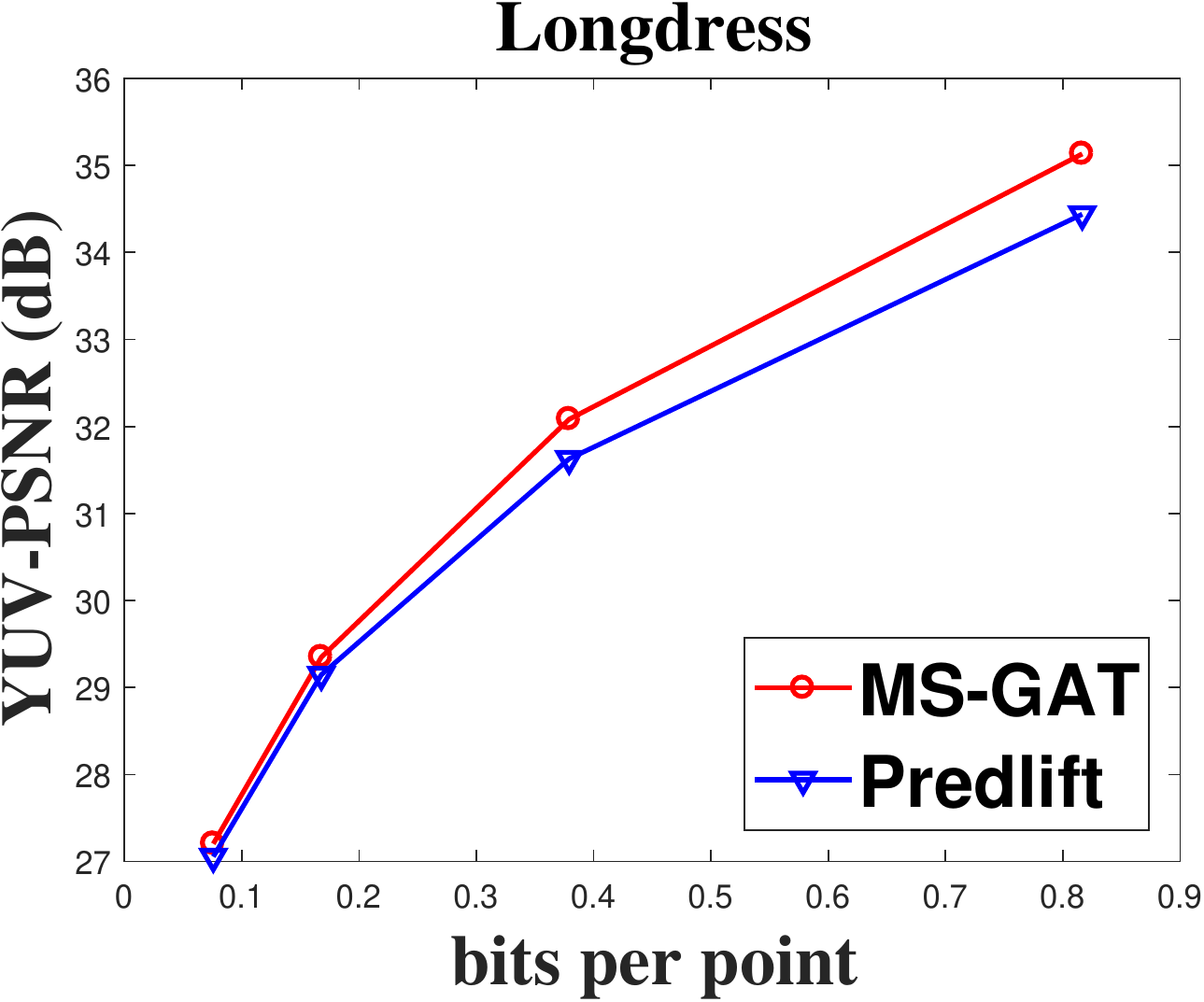}
  \includegraphics[width=0.234\linewidth]{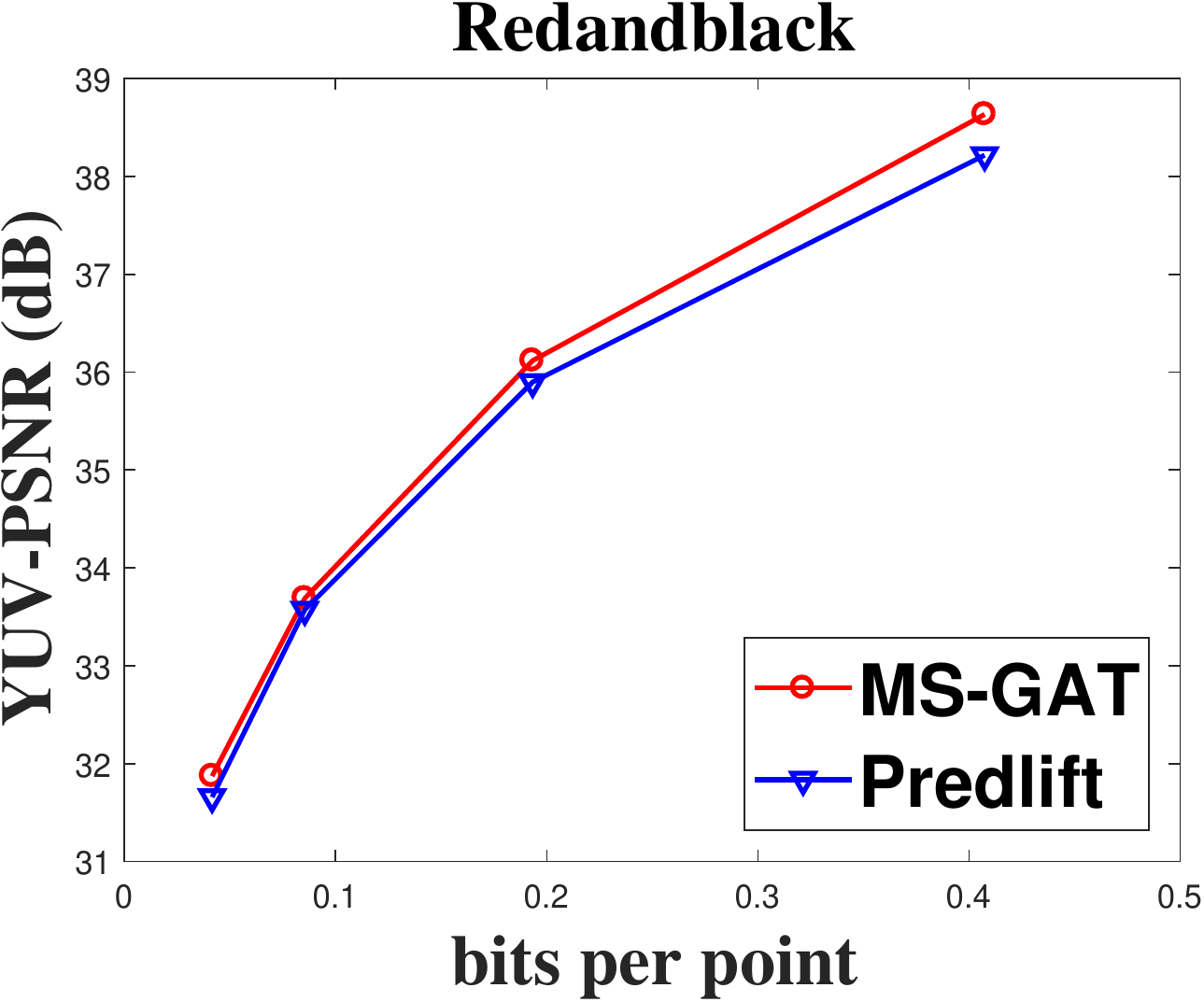}
  \includegraphics[width=0.234\linewidth]{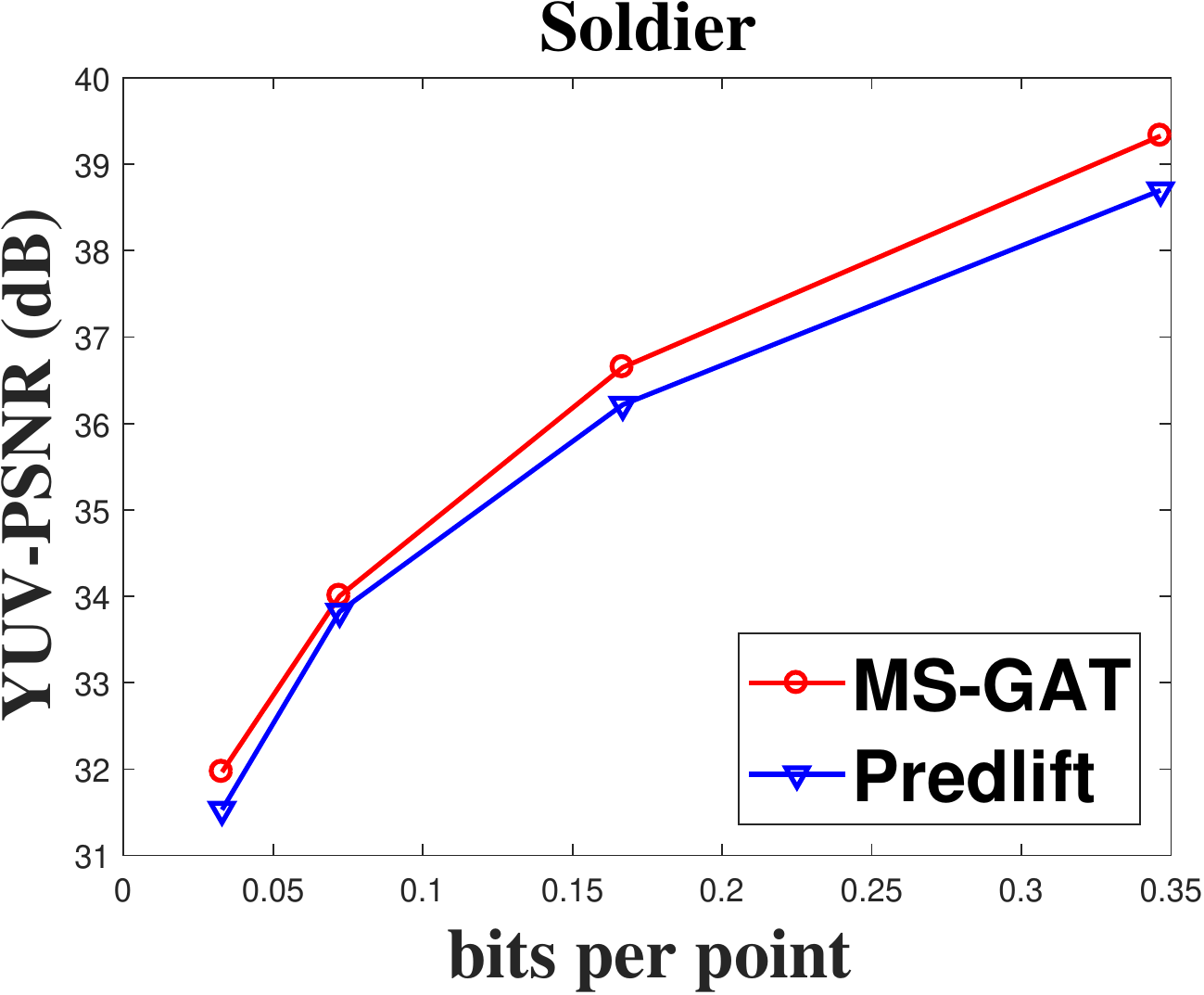}
  \includegraphics[width=0.234\linewidth]{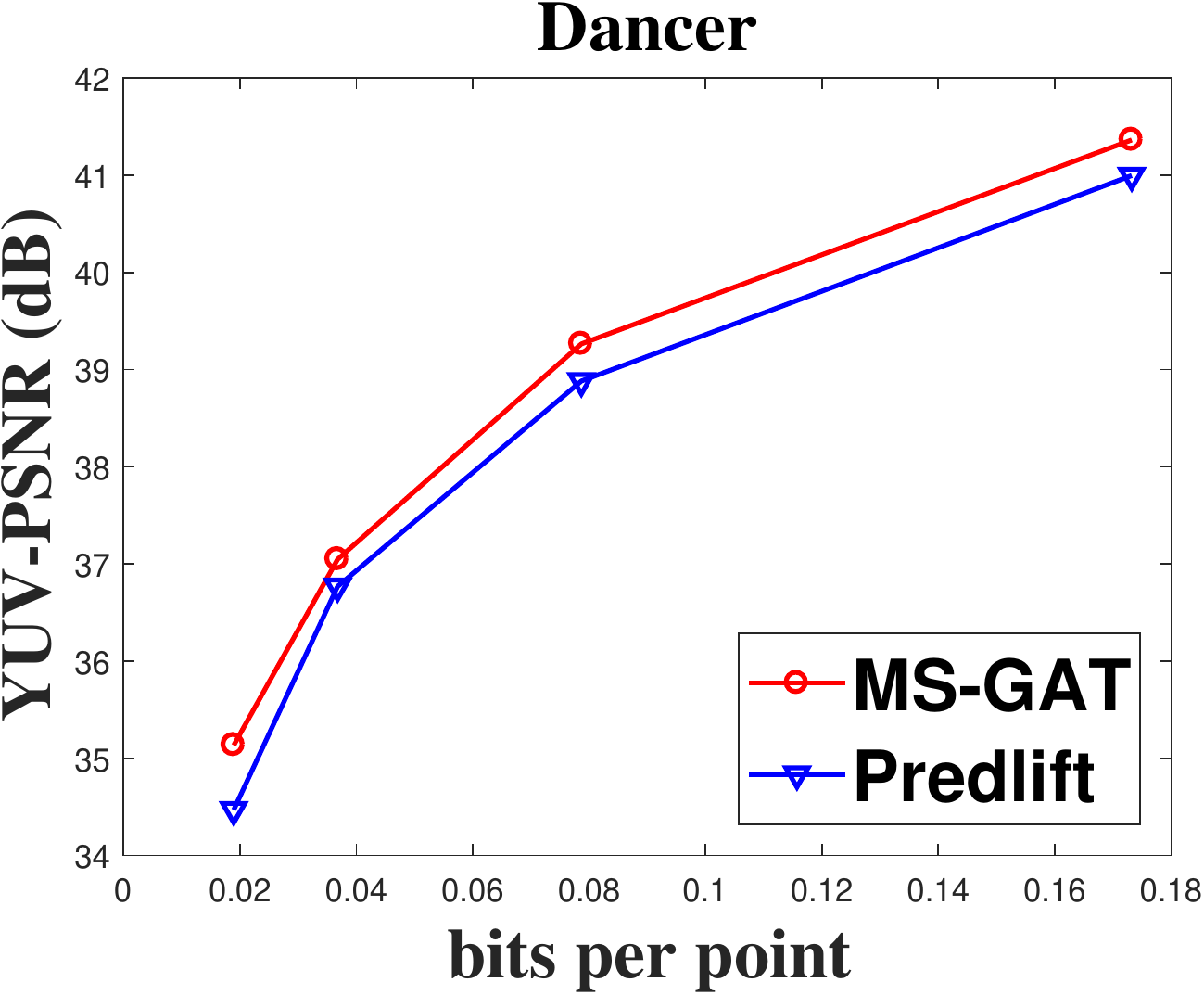}

\vspace{1mm}
  \includegraphics[width=0.234\linewidth]{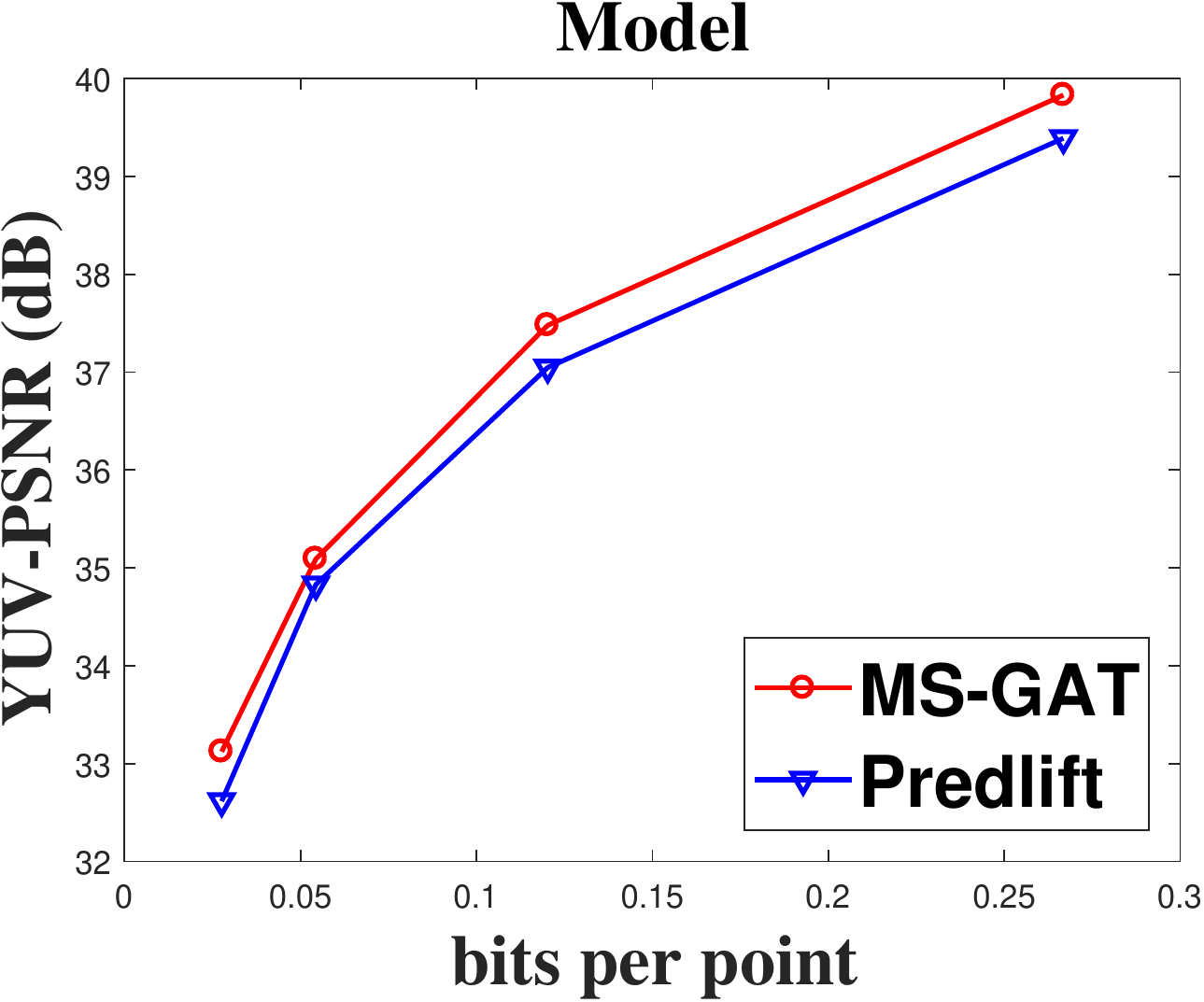}
  \includegraphics[width=0.234\linewidth]{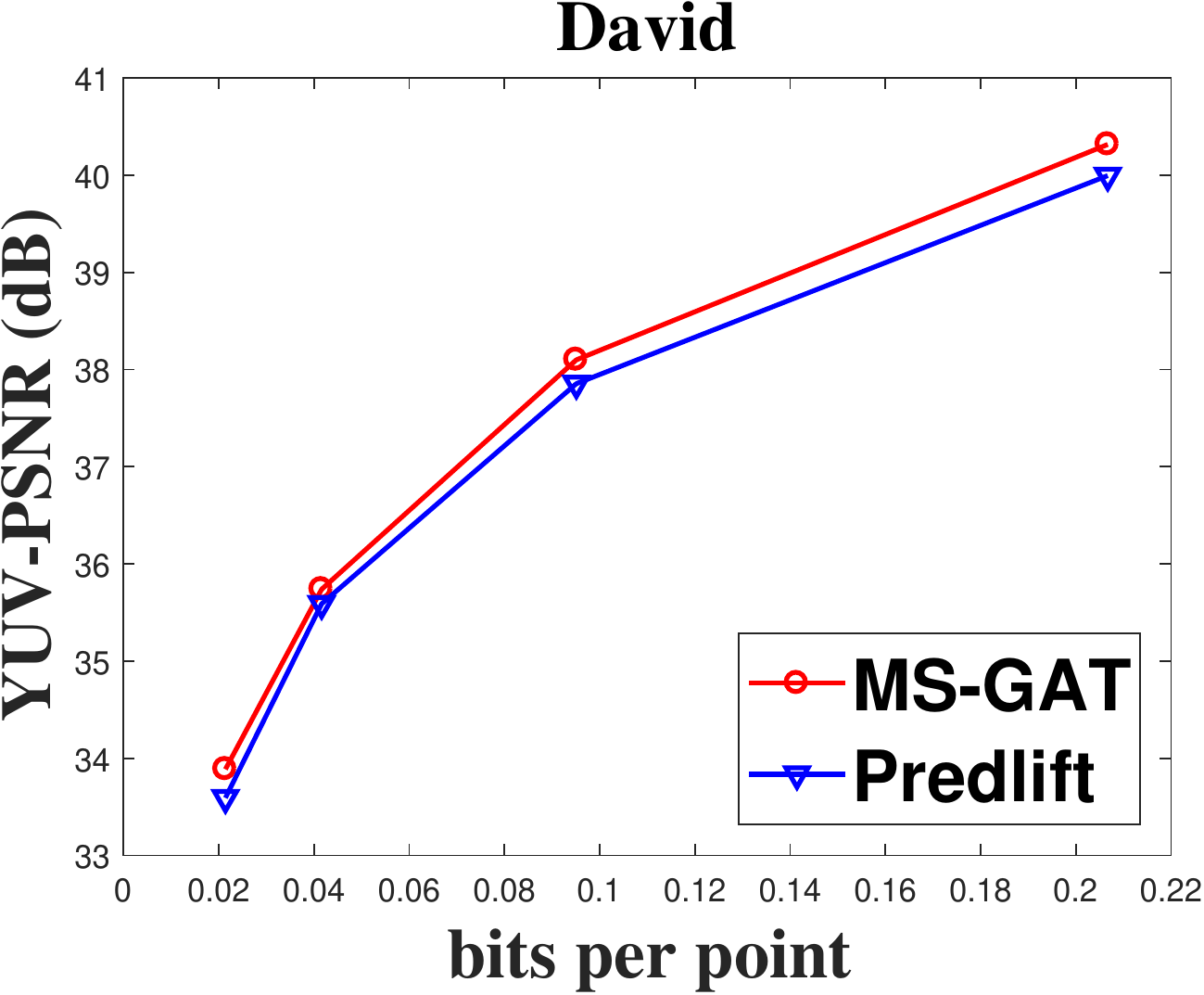}
  \includegraphics[width=0.234\linewidth]{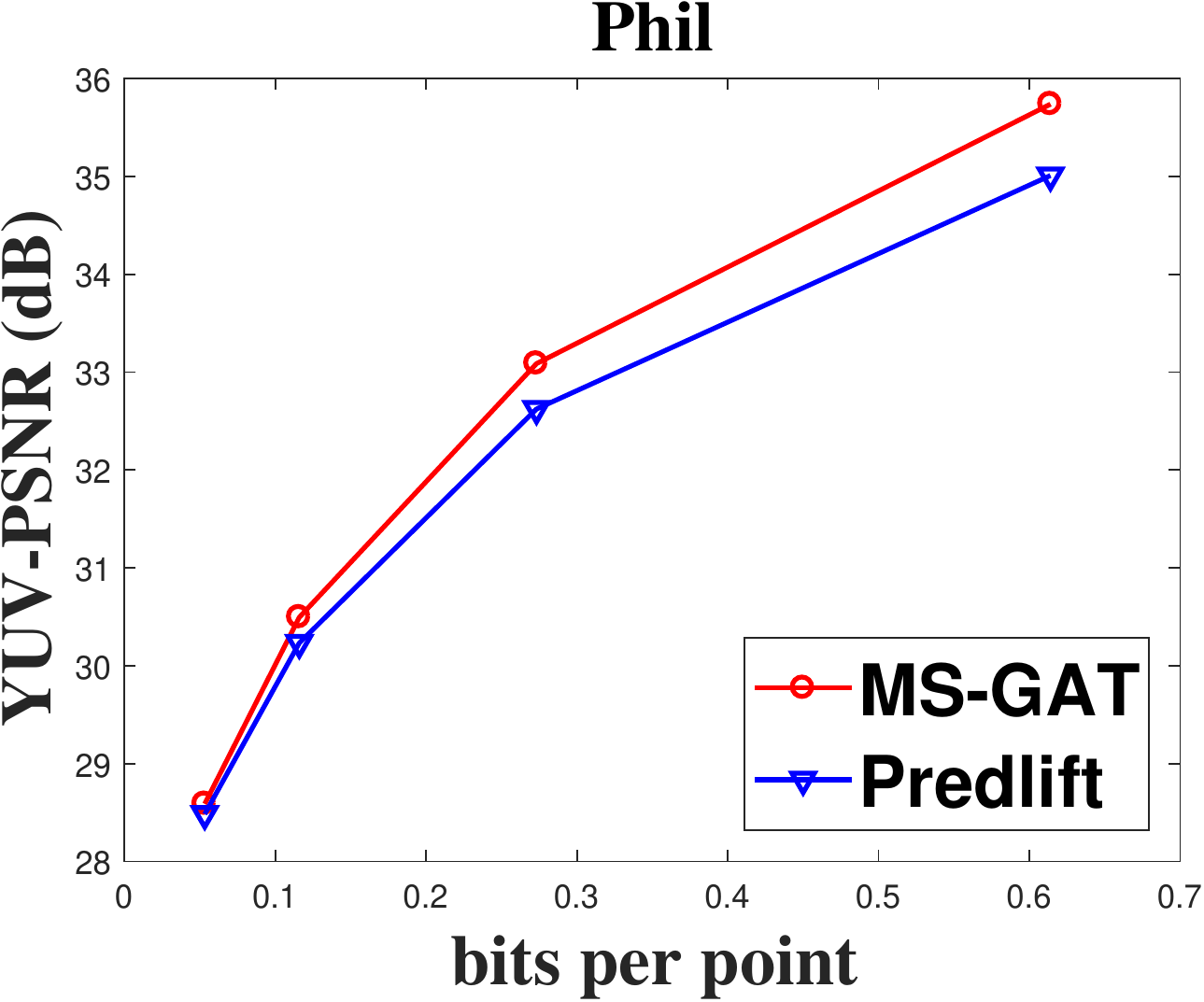}
  \includegraphics[width=0.234\linewidth]{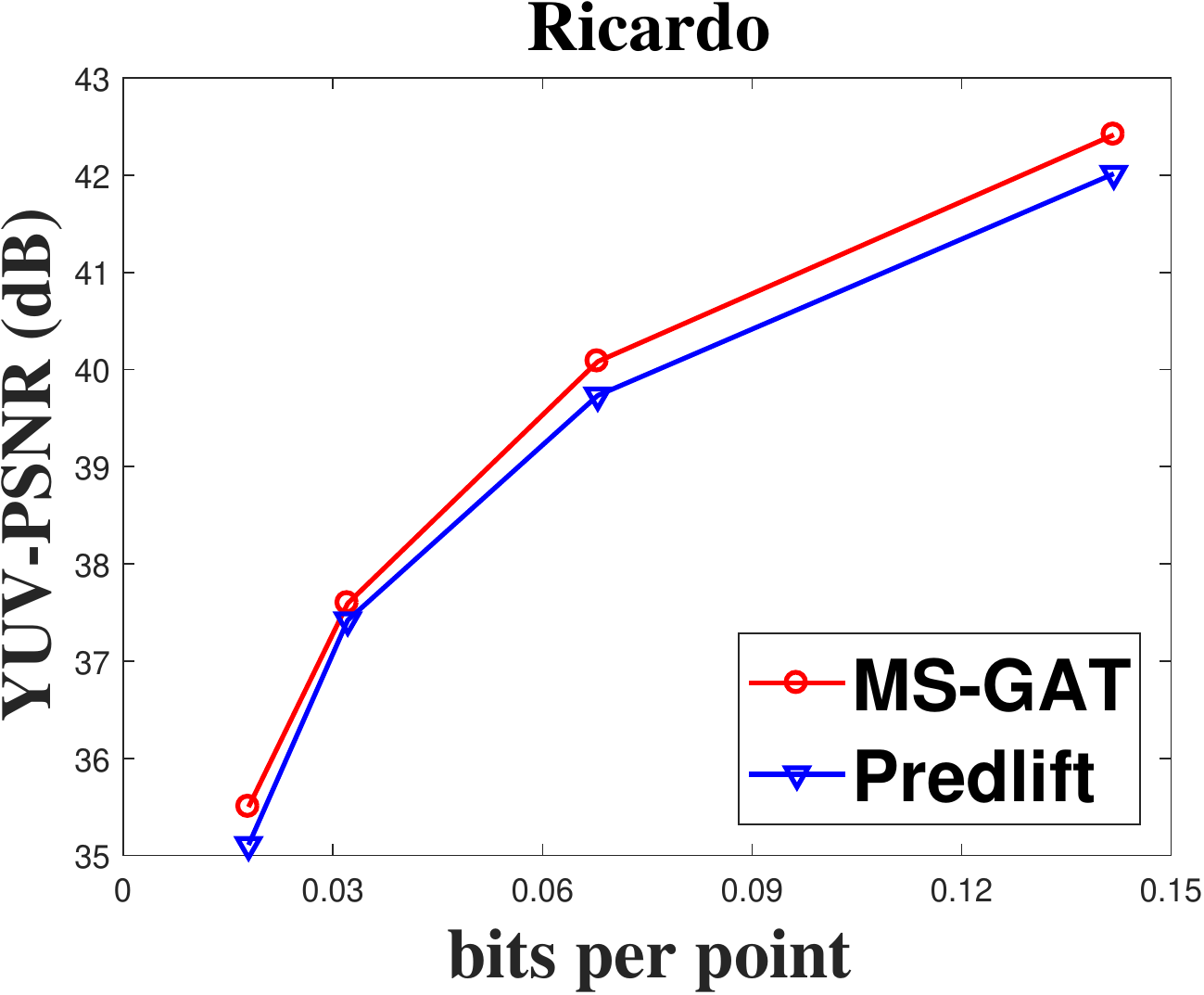}
  \caption{Rate-distortion curves for Predlift with and without our proposed algorithm on the test dataset. The quality is measured by YUV PSNR~\cite{ohm2012comparison}.}
  \label{fig:RDcurve_YUV_Predlift}
\end{figure*}

\begin{table}[]
\centering
\caption{BD-rate Comparison between Predlift with and without the proposed algorithm }
\label{table:bdrate_Predlift}
\begin{tabular}{c|c|c|c|c}
\hline
PointCloud  & BDBR-Y & BDBR-U & BDBR-V & BDBR-YUV \\ \hline
Longdress   & -12.73 & 0.15   & -2.73  & -10.17    \\ 
Redandblack & -8.008 & 1.06   & -3.04  & -6.64    \\ 
Soldier     & -11.72 & -1.28  & -1.45  & -10.79   \\ 
Dancer      & -13.18 & -2.26  & -1.71  & -11.89   \\
Model       & -13.23 & -3.87  & -4.40  & -11.94   \\ 
Andrew      & -7.02  & -4.37  & -5.66  & -6.67    \\ 
David       & -8.33  & -3.37  & -2.12  & -7.56    \\
Phil        & -13.73 & -0.53  & 0.68   & -12.27   \\
Sarah       & -12.75 & -3.52  & -3.16  & -10.84    \\
Ricardo     & -8.93  & -3.05  & -4.55  & -8.39    \\ \hline
Average     & -10.97 & -2.10  & -2.81  & -9.74    \\ \hline
\end{tabular}
\end{table}

\section{Experimental Results}\label{experiments}
\subsection{Experimental Setup}
\subsubsection{Datasets}
As far as we know, there is no widely used training dataset of compression artifacts removal for point cloud attributes. 
We use five static point clouds defined in the MPEG PCC common test condition~\cite{schwarz2018common} as training data, including ``Basketball Player'', ``Loot'', ``Exerciser'', ``Queen'', and ``Boxer'', as shown in  Fig.~\ref{fig:training}. 
We partition the selected point clouds into small blocks with each of them having $n$ points. 
Considering the limitations of the GPU memory, we set $n$ to $2048$.
We compress the point cloud attributes with Predlift and RAHT in TMC13v12 to generate two training datasets for Predlift and RAHT, respectively.
In total, we collect $5145$ small blocks in each training dataset. The two training datasets are used to train two models with the same structure for Predlift and RAHT, respectively.
To demonstrate the effectiveness of our proposed MS-GAT, we test it on ten point clouds as shown in Fig.~\ref{fig:testing}, including ``Longdress'', ``Redandblack'', ``Soldier'', ``Dancer'', ``Model'', ``Andrew'', ``David'', ``Phil'', ``Ricardo'', and ``Sarah''.
\begin{figure*}
\centering
  \includegraphics[width=0.234\linewidth]{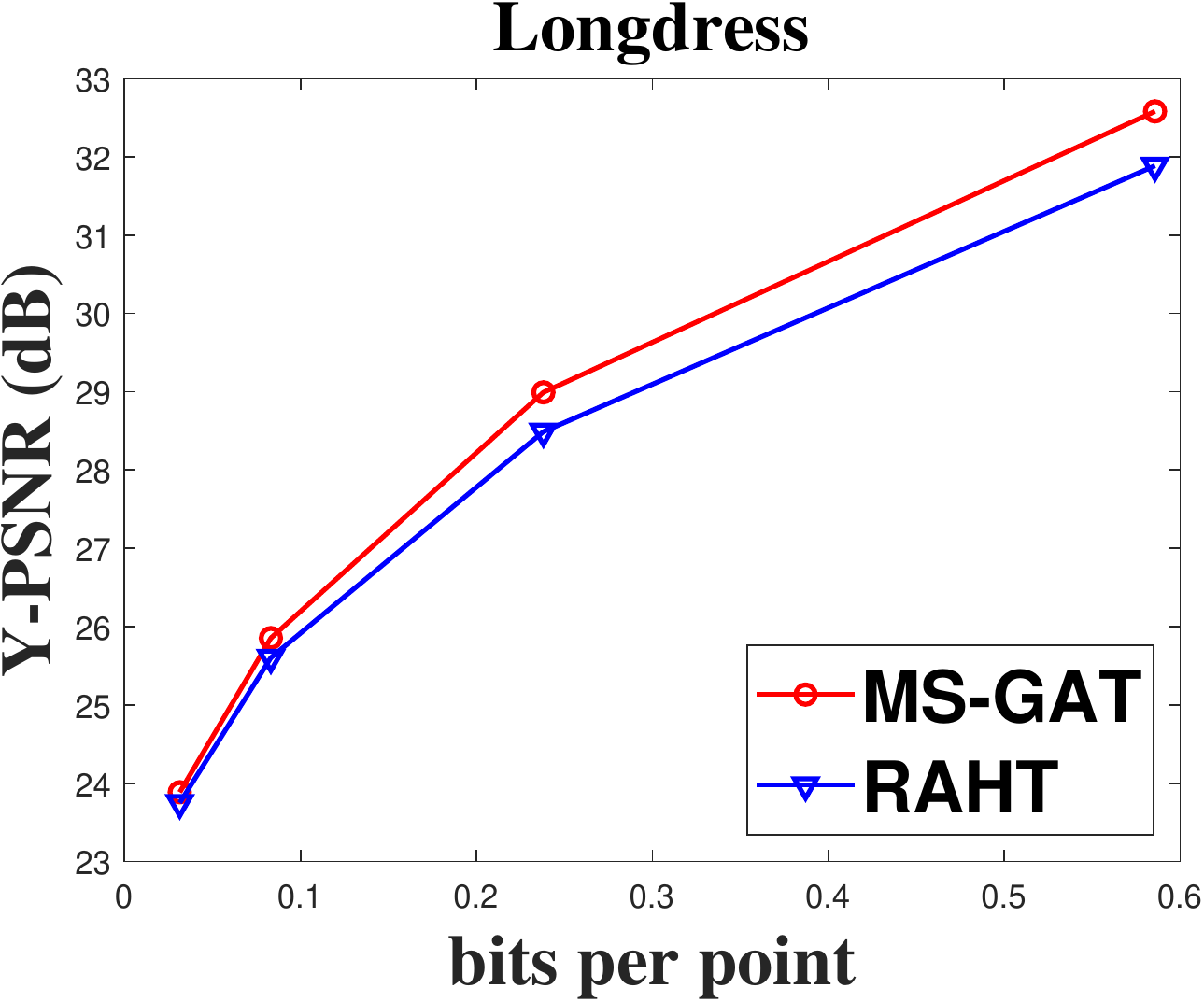}
  \includegraphics[width=0.234\linewidth]{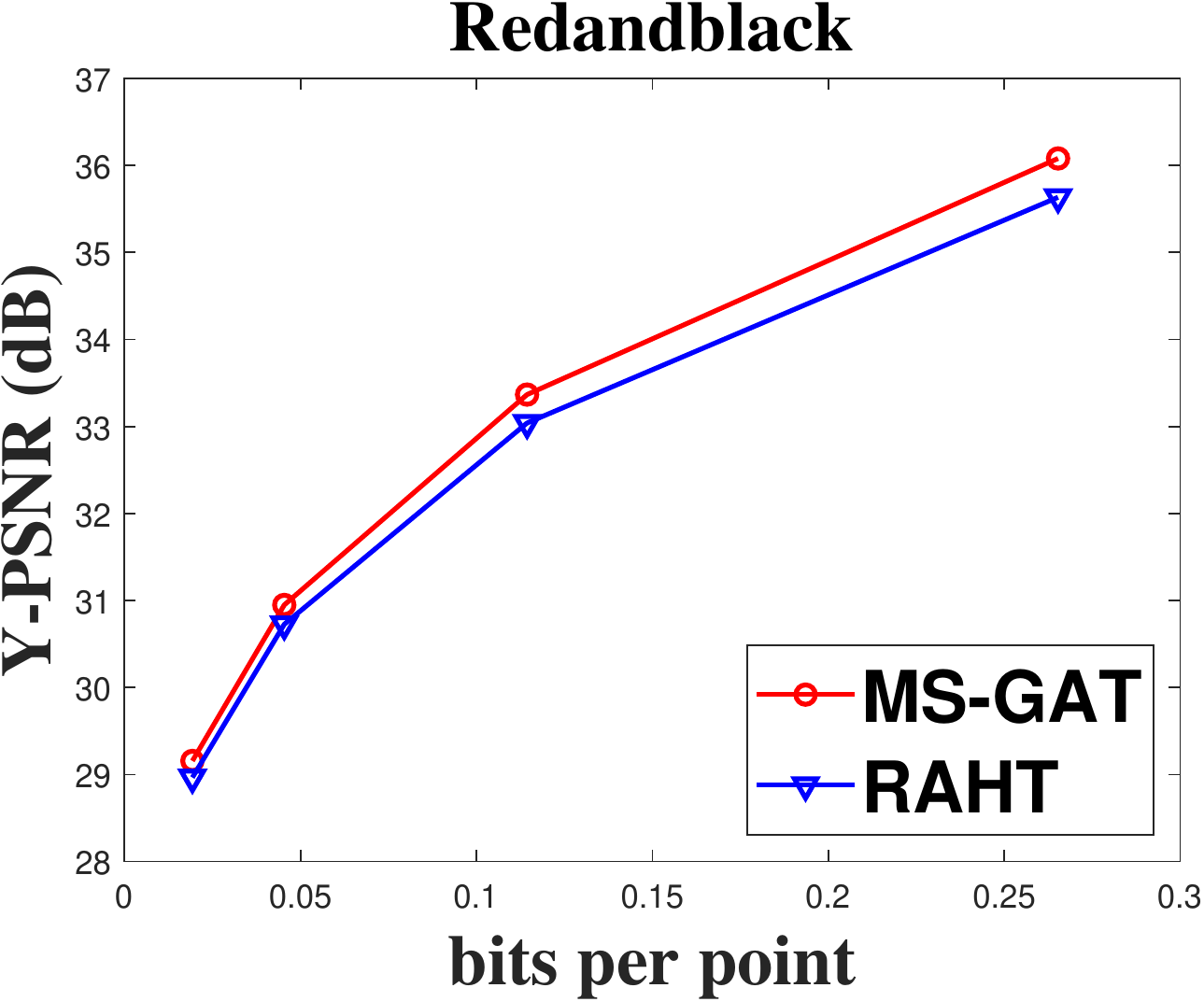}
  \includegraphics[width=0.234\linewidth]{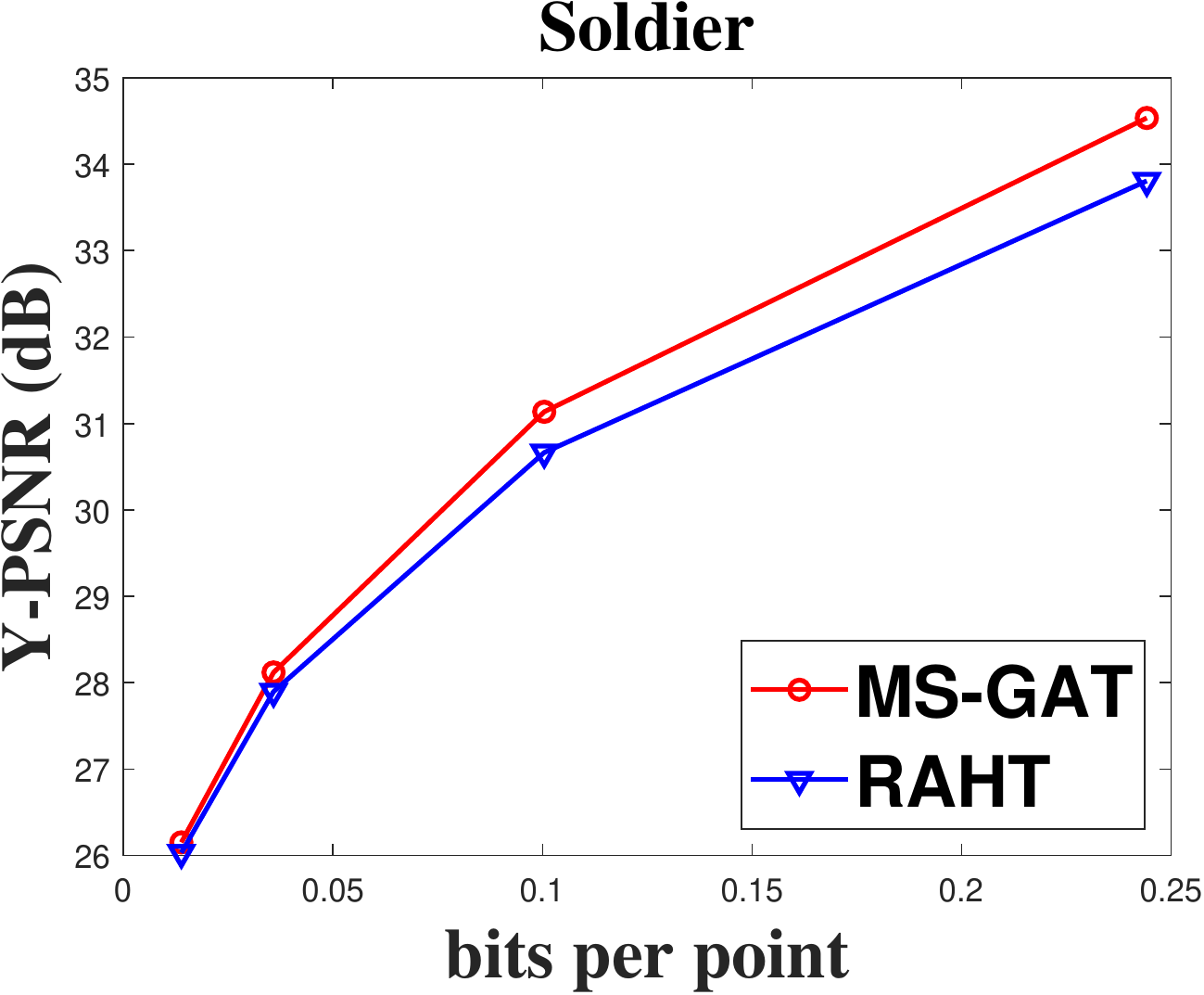}
  \includegraphics[width=0.234\linewidth]{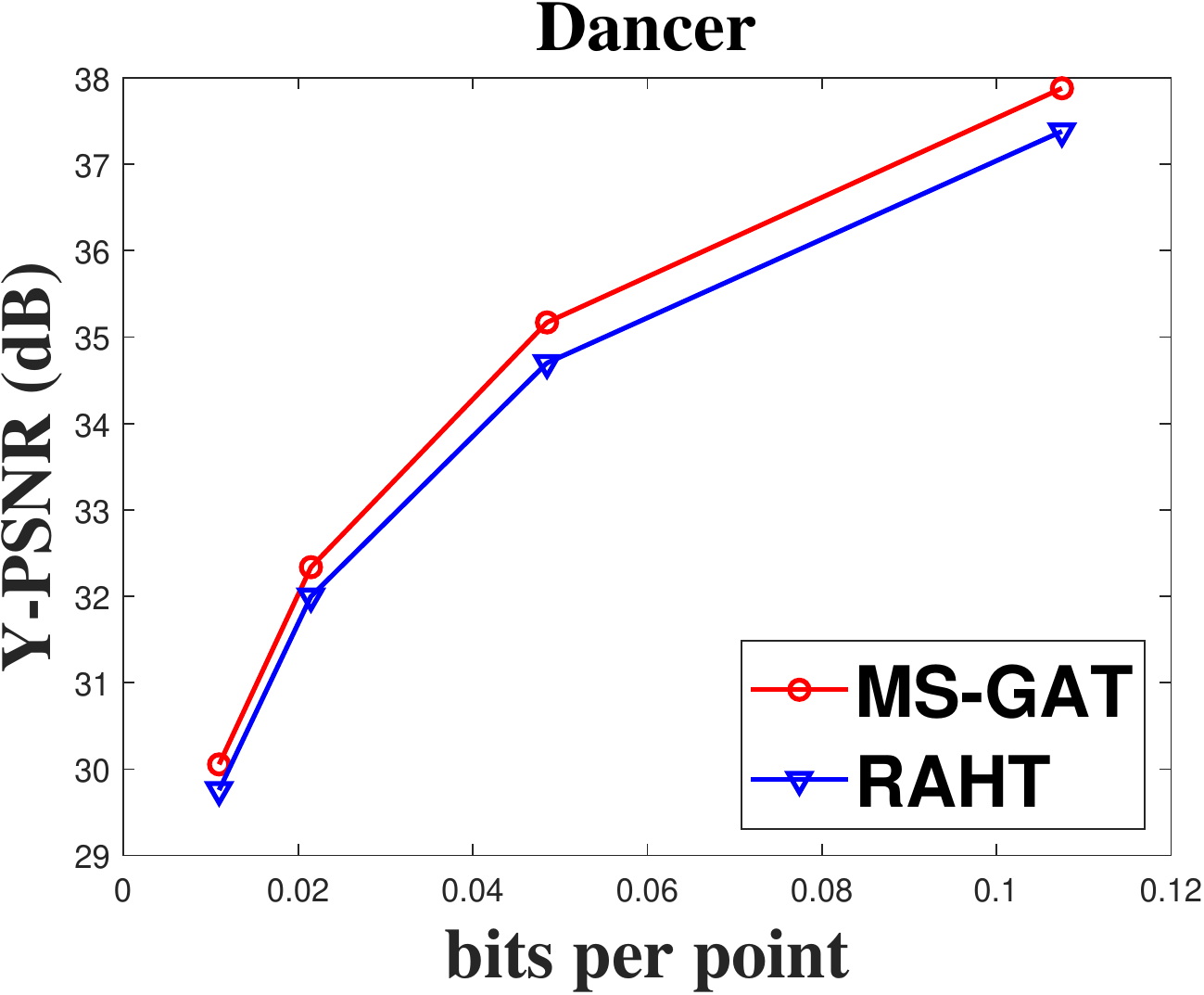}

\vspace{1mm}
  \includegraphics[width=0.234\linewidth]{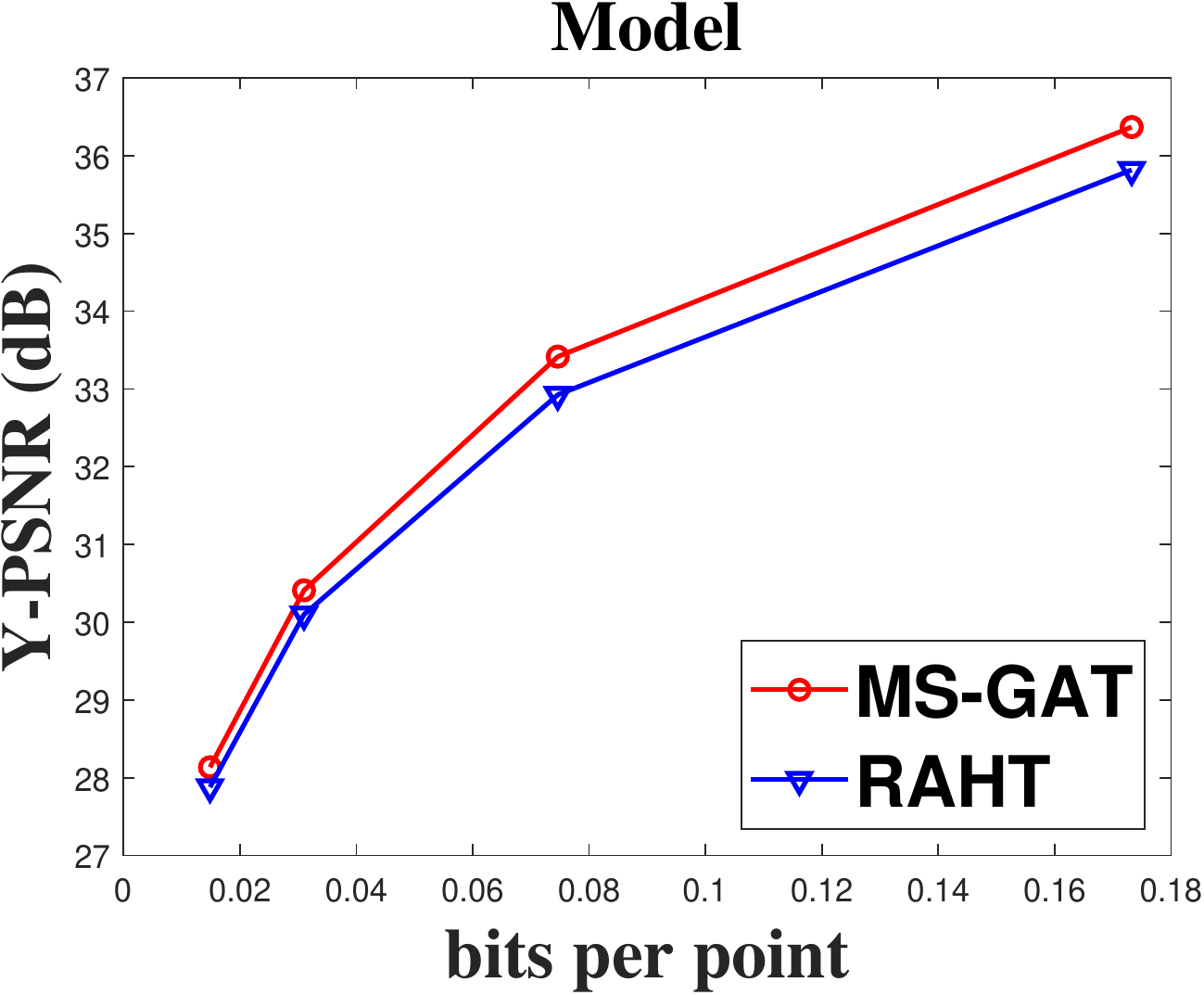}
  \includegraphics[width=0.234\linewidth]{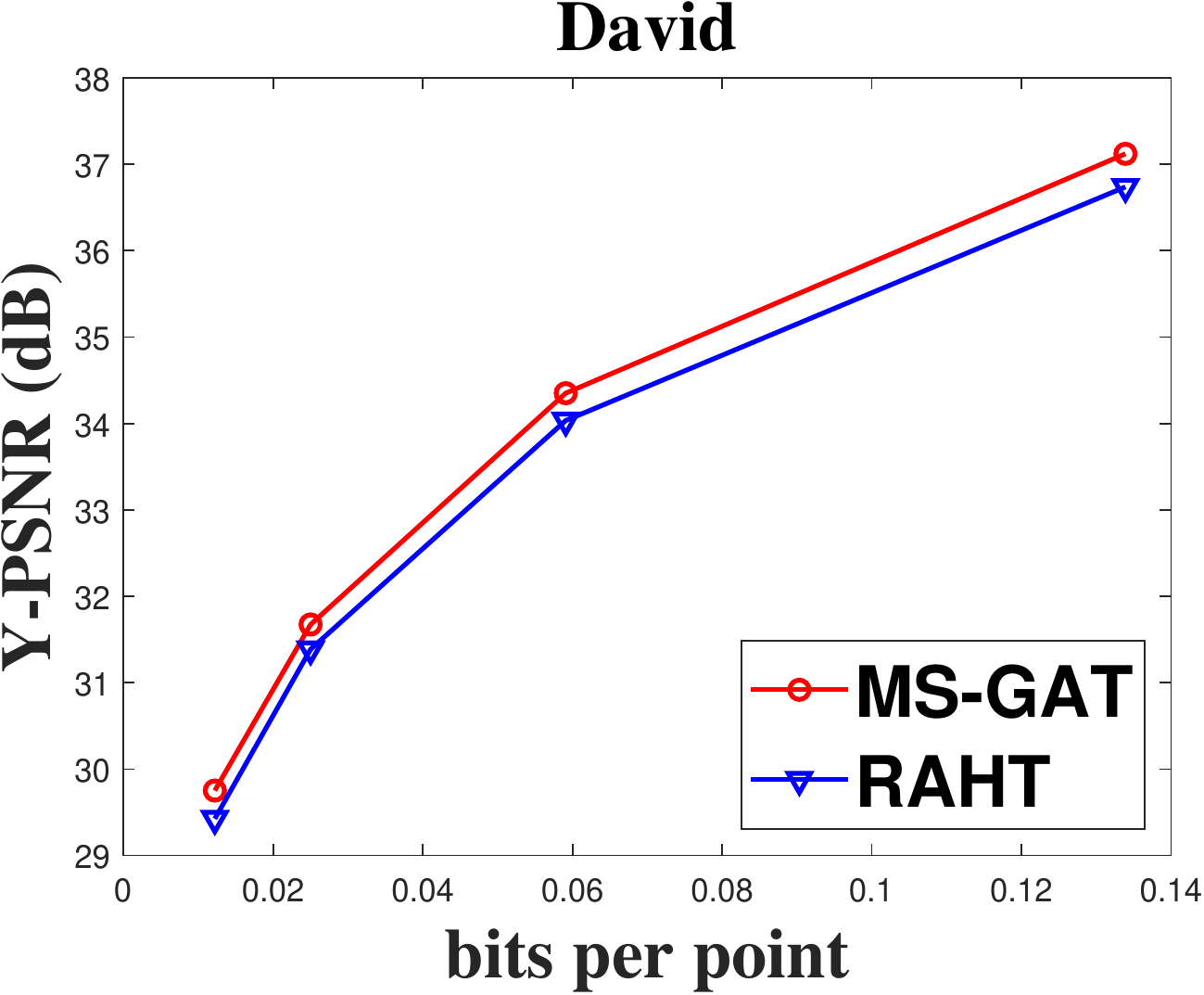}
  \includegraphics[width=0.234\linewidth]{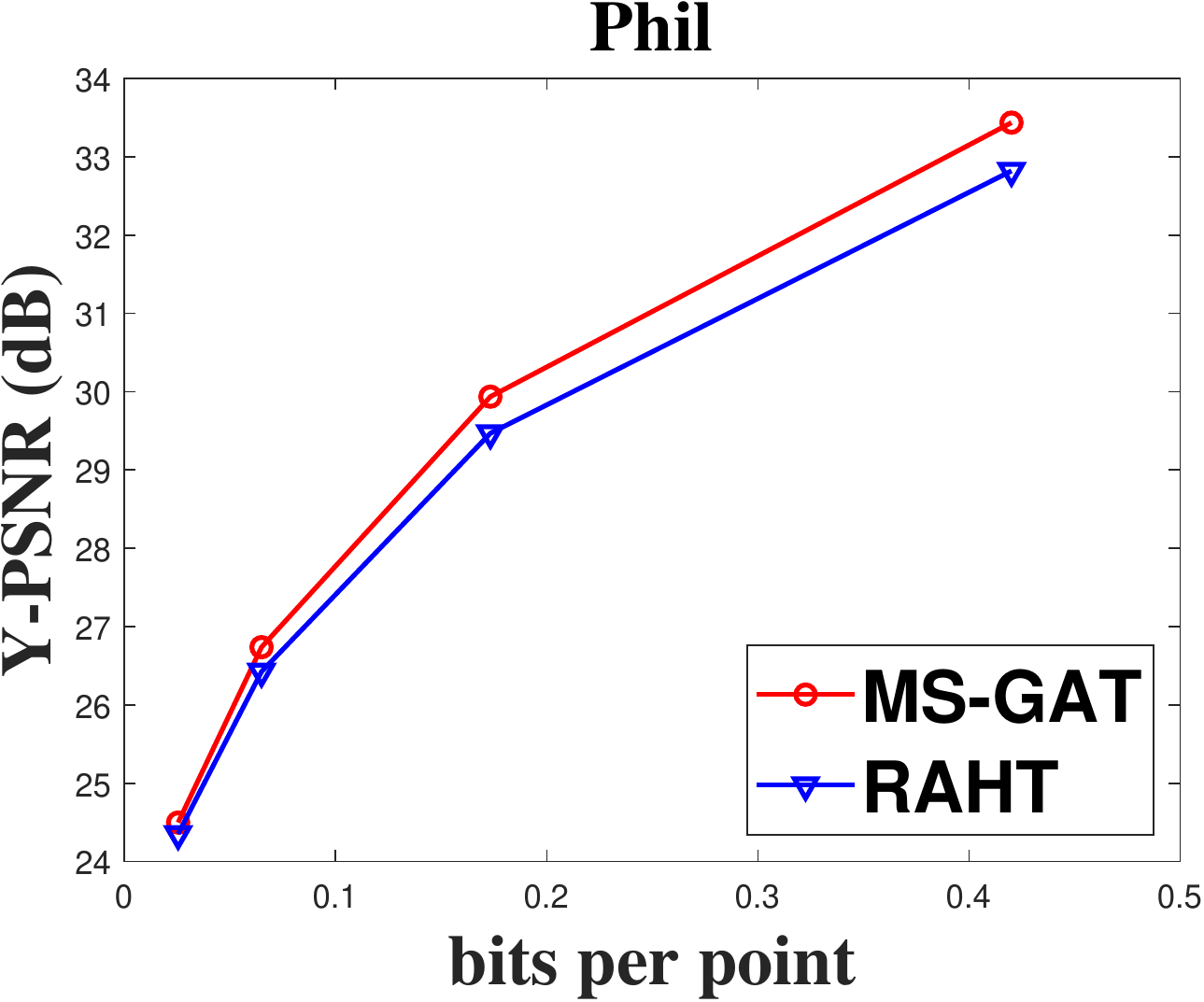}
  \includegraphics[width=0.234\linewidth]{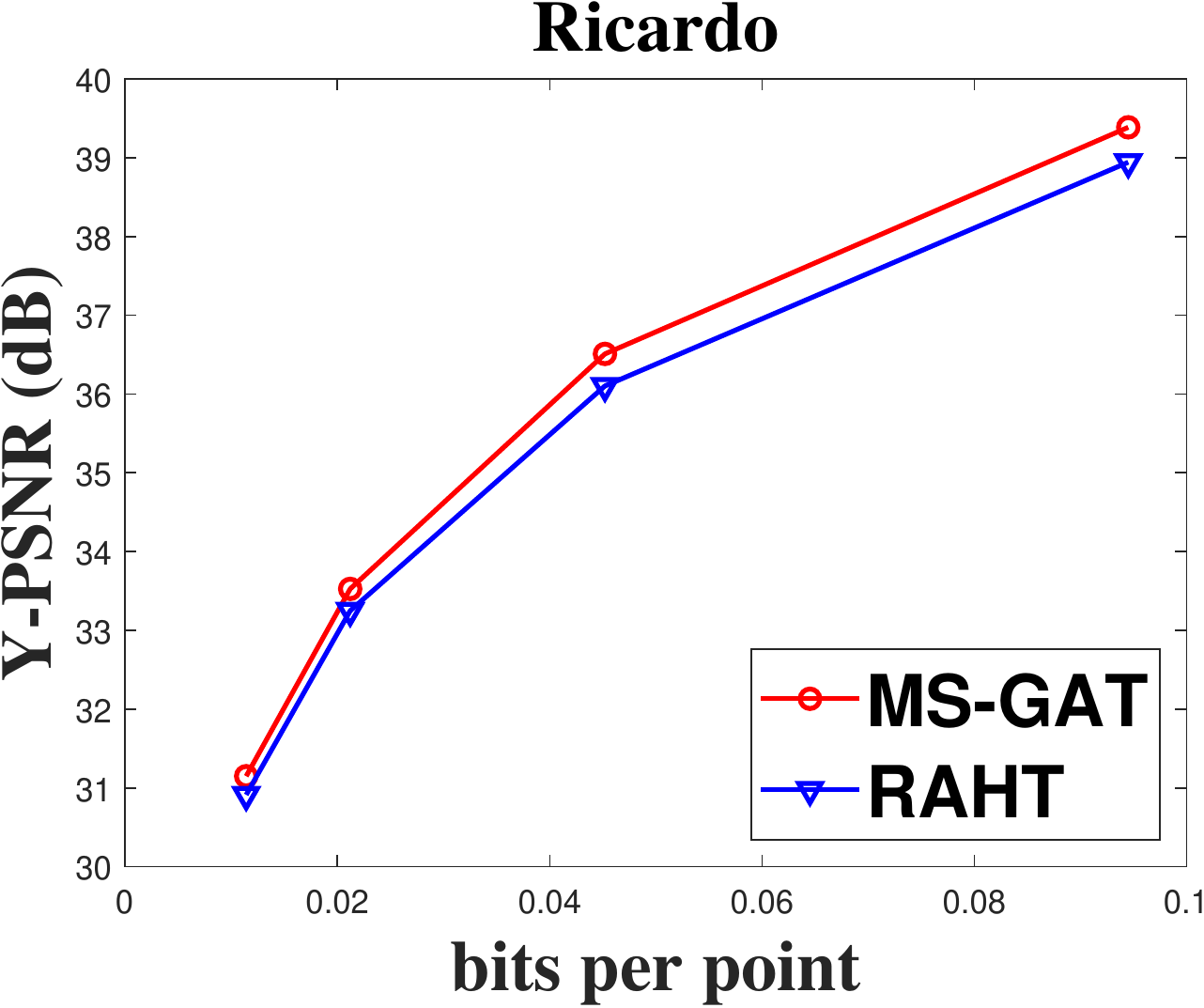}

  \caption{Rate-distortion curves for RAHT with and without our proposed algorithm on the test dataset. ``MS-GAT" denotes the rate-distortion curve of point cloud attributes restored by our proposed algorithm. ``RAHT" denotes the rate-distortion curve of RAHT-compressed attributes. The quality is measured by Y PSNR.}
  \label{fig:RDcurve_Y_RAHT}
\end{figure*}

\begin{figure*}
\centering
  \includegraphics[width=0.234\linewidth]{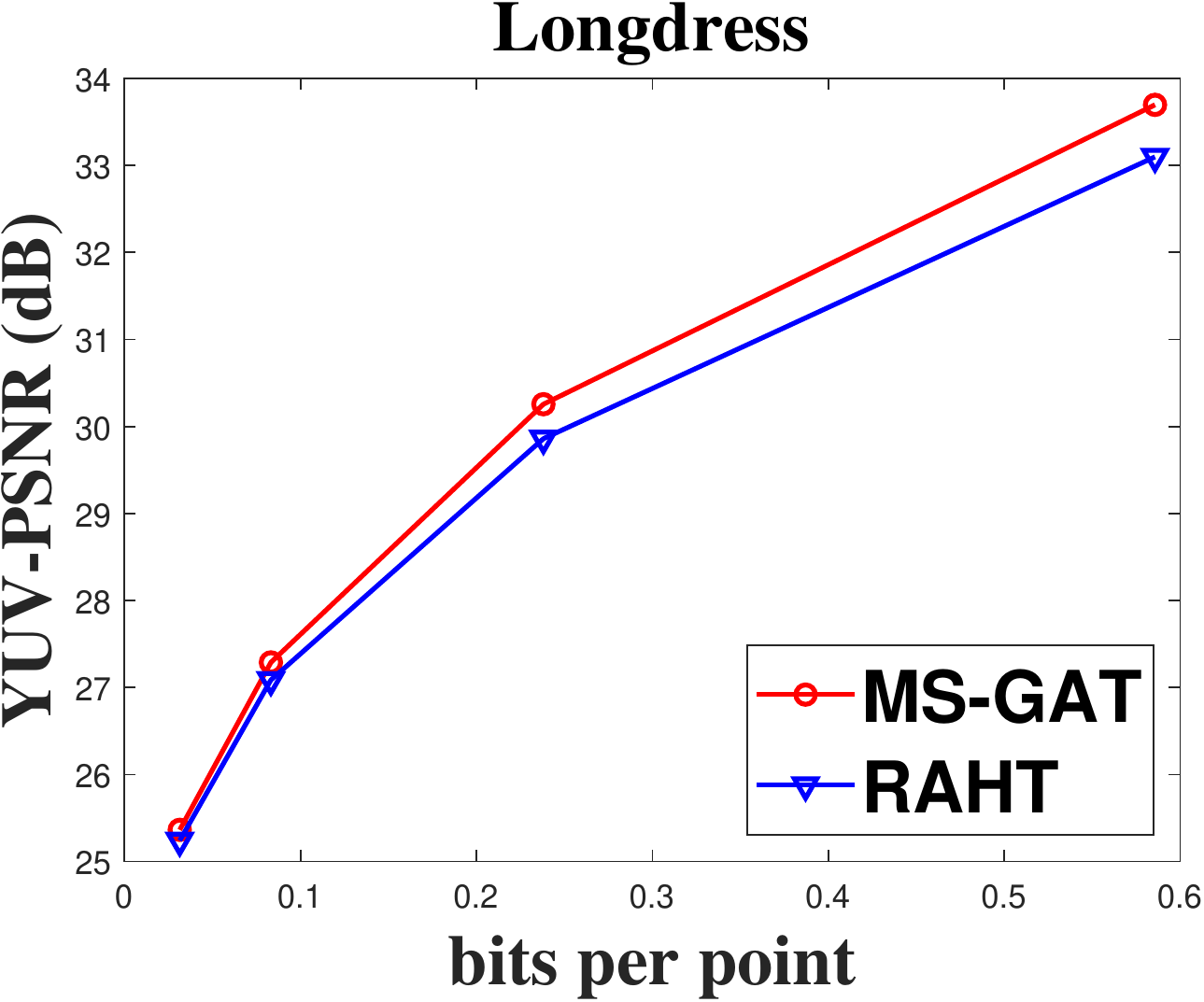}
  \includegraphics[width=0.234\linewidth]{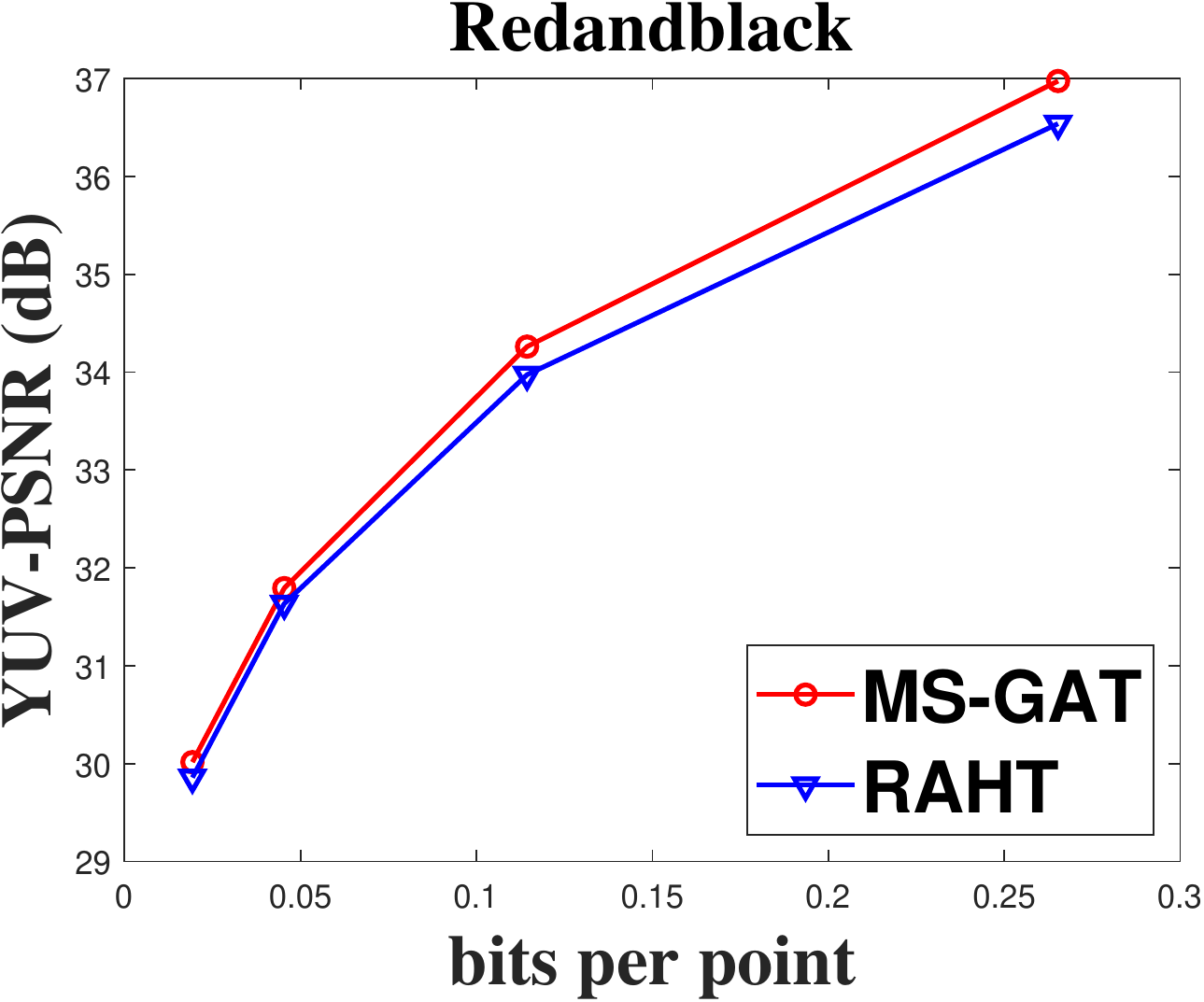}
  \includegraphics[width=0.234\linewidth]{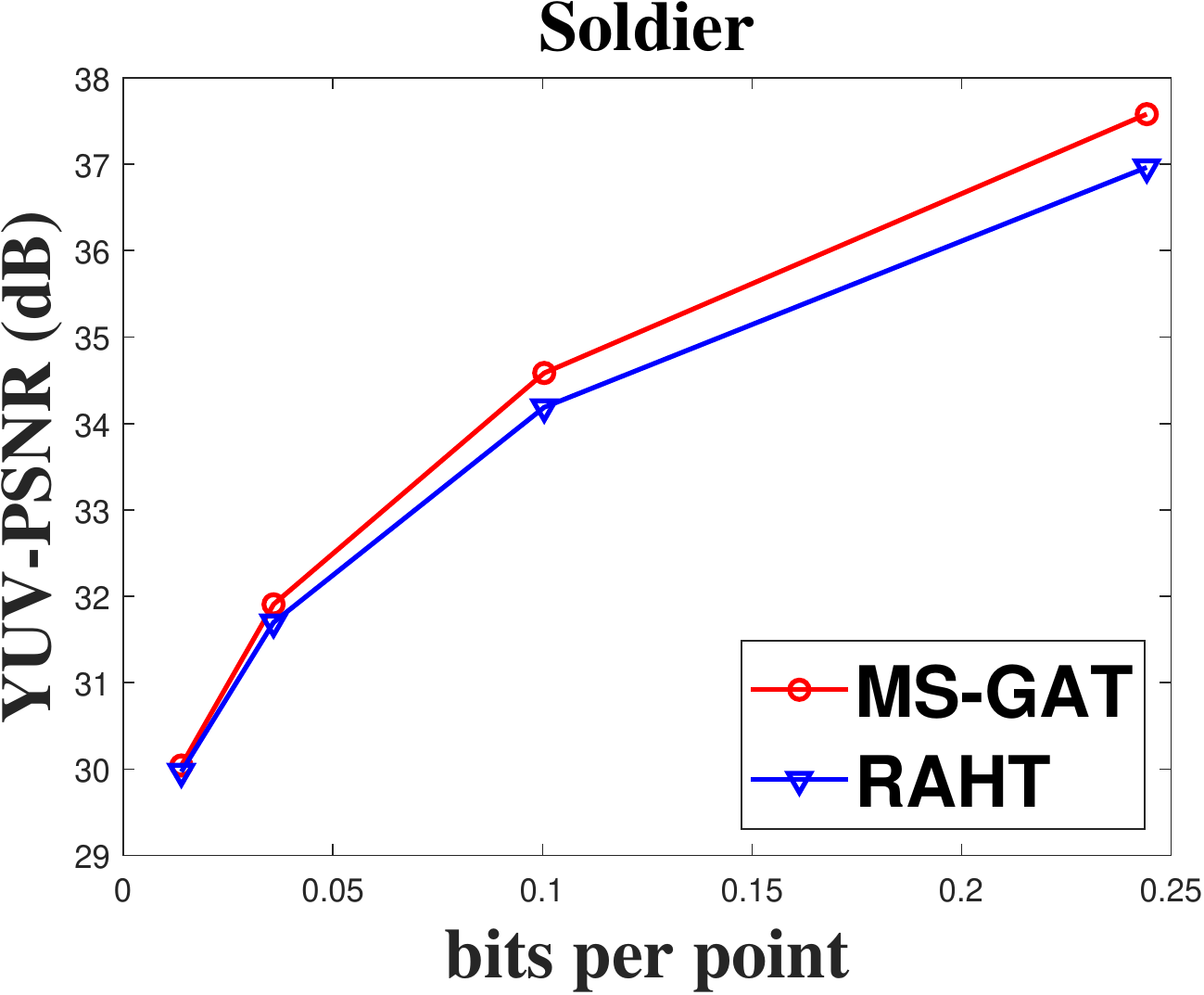}
  \includegraphics[width=0.234\linewidth]{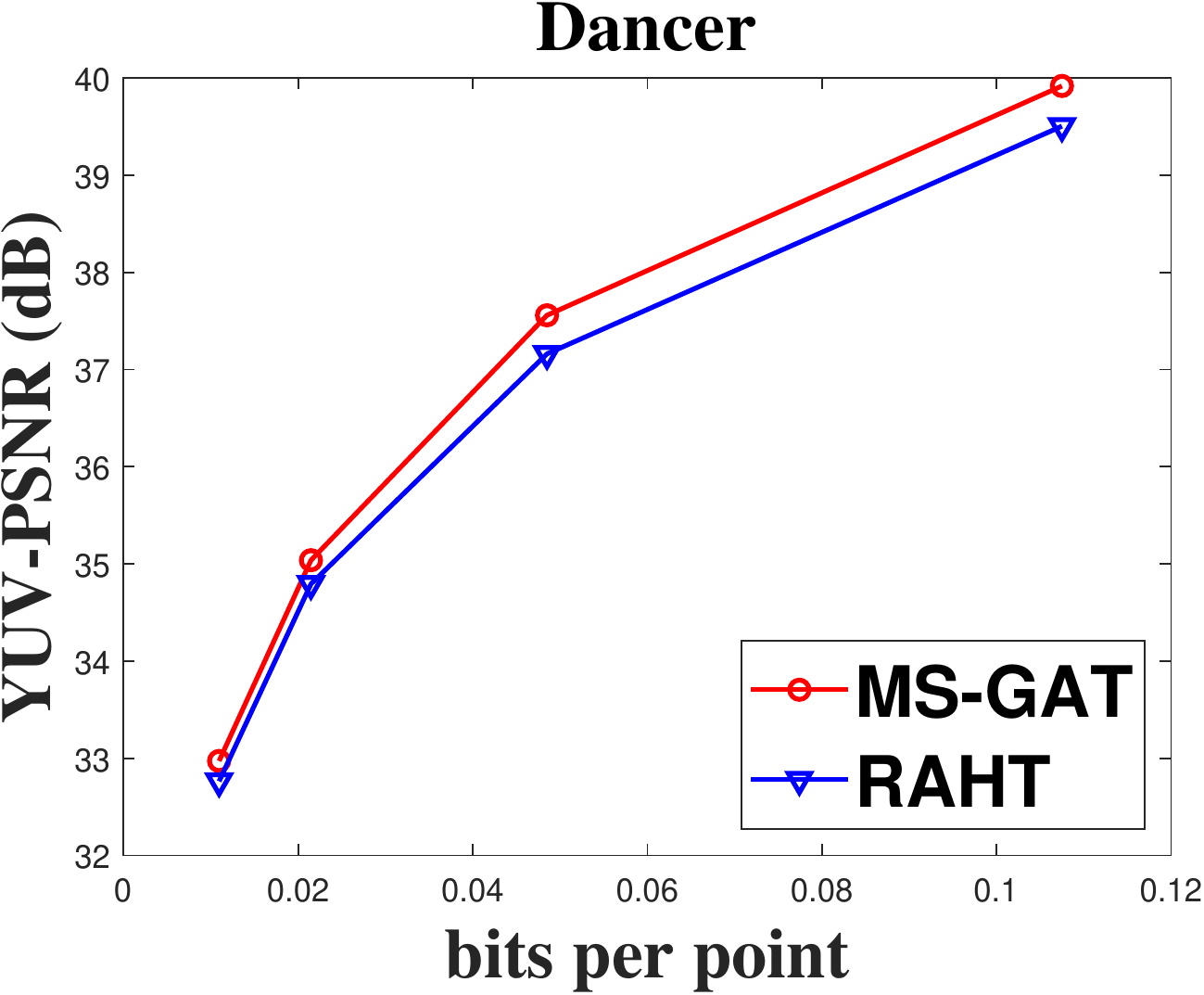}

\vspace{1mm}
  \includegraphics[width=0.234\linewidth]{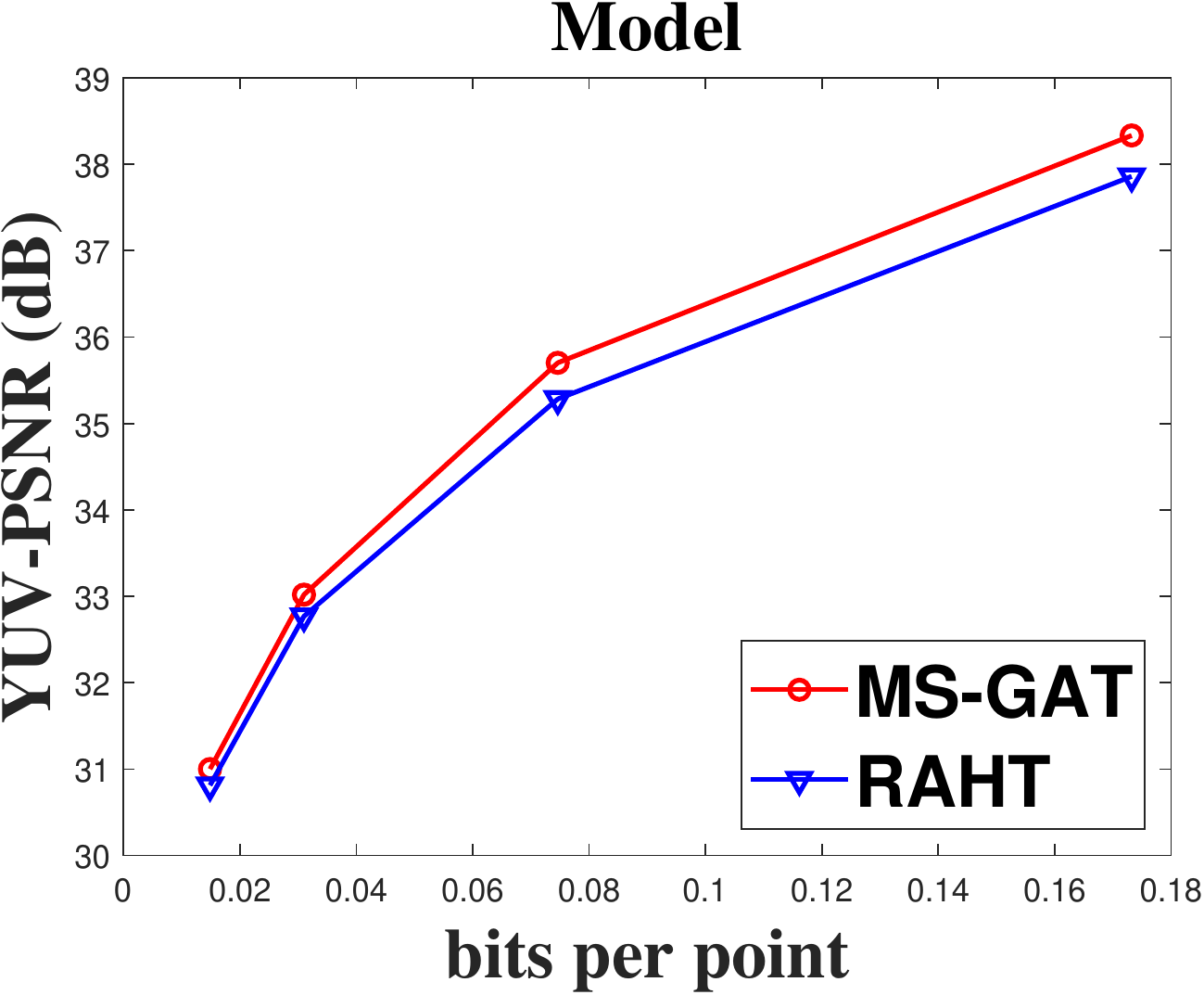}
  \includegraphics[width=0.234\linewidth]{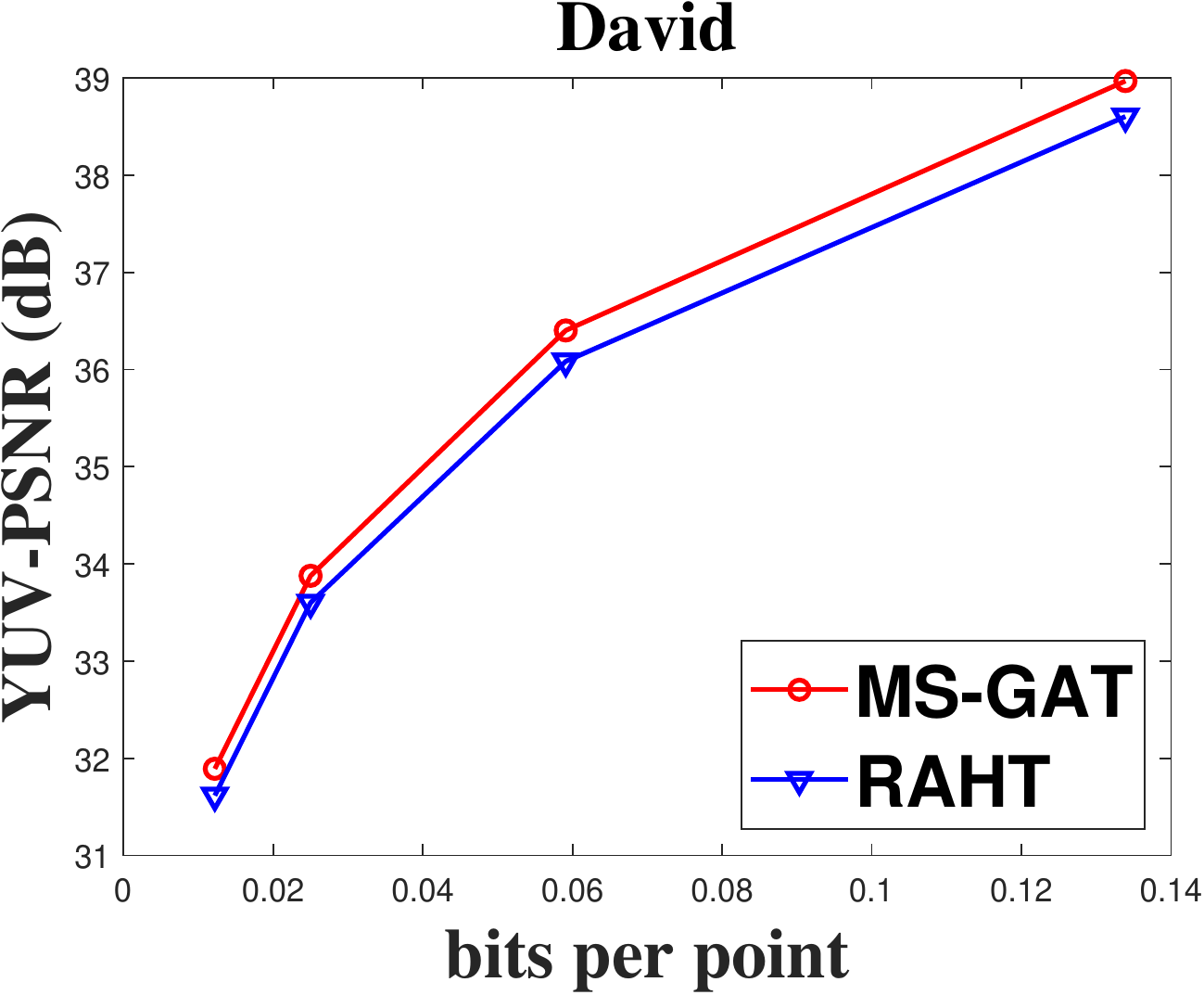}
  \includegraphics[width=0.234\linewidth]{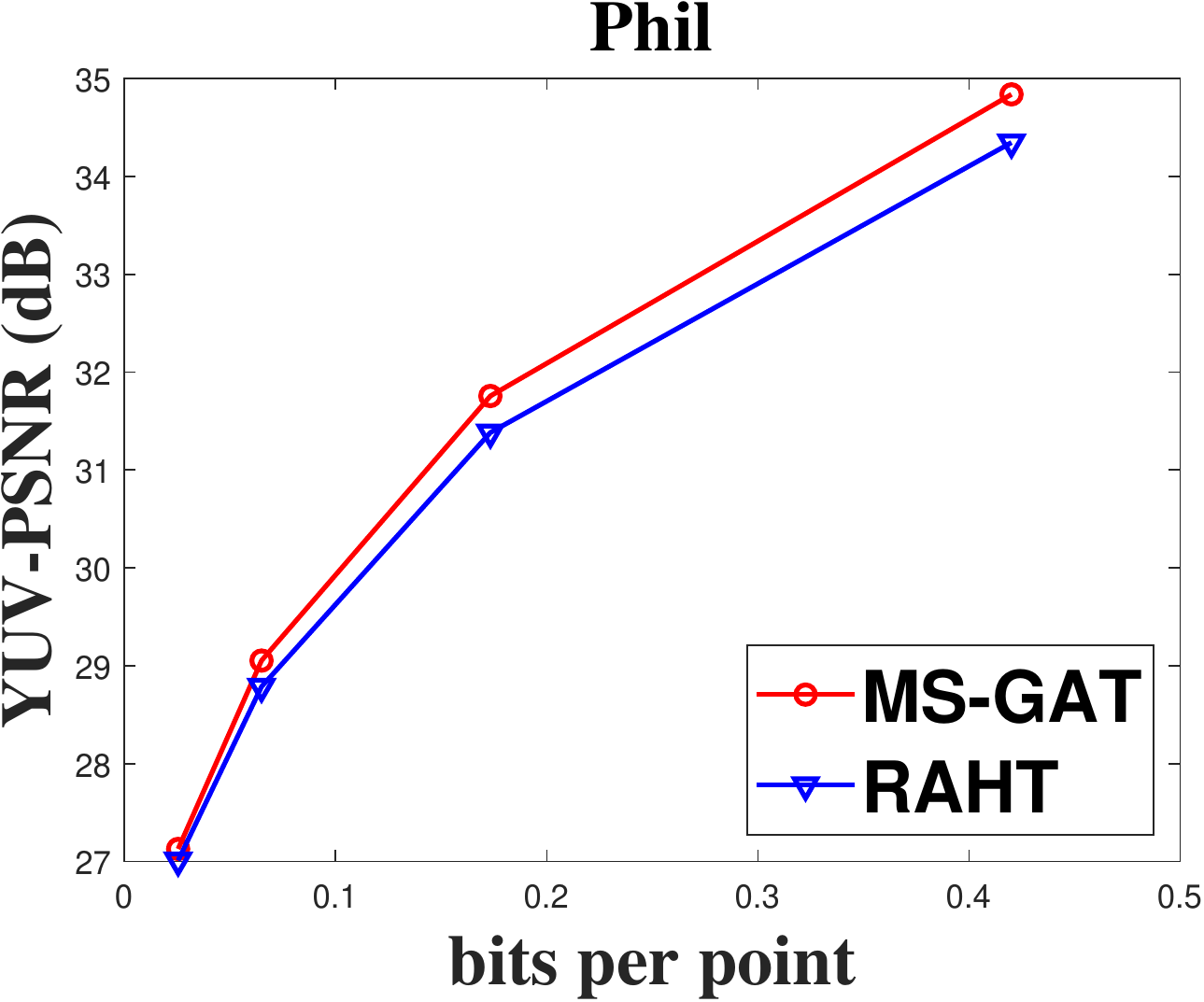}
  \includegraphics[width=0.234\linewidth]{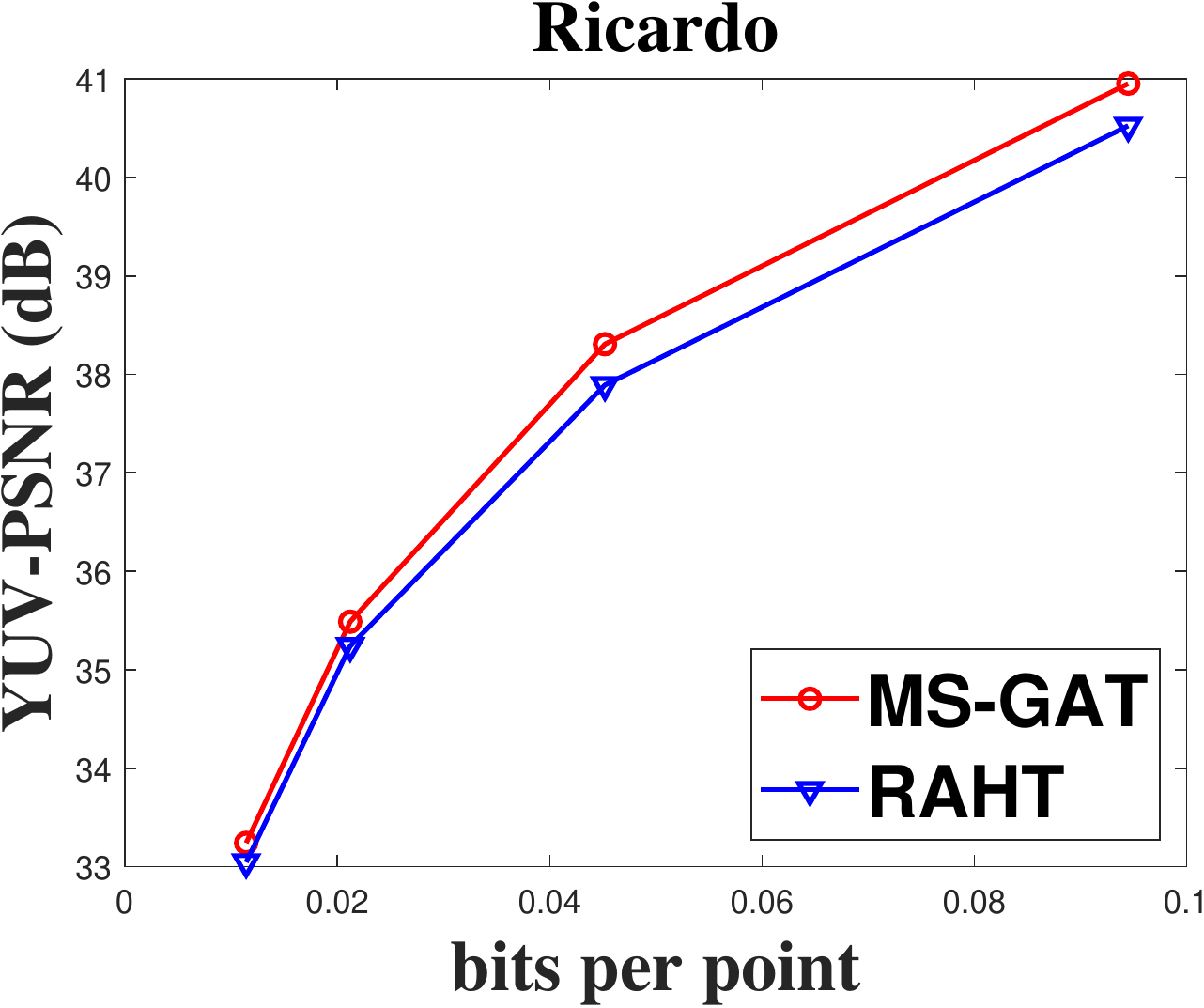}

  \caption{Rate-distortion curves for RAHT with and without our proposed algorithm on the test dataset. The quality is measured by YUV PSNR.}
  \label{fig:RDcurve_YUV_RAHT}
\end{figure*}
\begin{table}[]
\centering
\caption{BD-rate Comparison between RAHT with and without the proposed algorithm }
\label{table:bdrate_RAHT}
\begin{tabular}{c|c|c|c|c}
\hline
PointCloud  & BDBR-Y & BDBR-U & BDBR-V & BDBR-YUV \\ \hline
Longdress   & -12.37 & -2.58   & -4.56  & -10.56    \\ 
Redandblack & -10.40 & -2.56   & -4.58  & -8.81    \\ 
Soldier     & -11.93 & -6.60  & -10.46  & -11.31   \\ 
Dancer      & -11.37 & -2.10  & -4.84  & -10.26   \\
Model       & -11.66 & -3.72  & -8.59  & -10.94   \\ 
Andrew      & -7.87  & -9.03  & -11.45  & -7.81    \\ 
David       & -9.74  & -9.19  & -11.23  & -9.84    \\
Phil        & -11.54 & -6.84  & -5.47   & -10.91   \\
Sarah       & -12.06 & -10.83  & -12.74  & -12.02   \\
Ricardo     & -8.64  & -7.97  & -11.25  & -8.87    \\ \hline
Average     & -10.76 & -6.14  & -8.83  & -10.13    \\ \hline
\end{tabular}
\end{table}
\begin{table*}[]
\centering
\caption{Run time Analysis (In Seconds). }
\label{table:time}
\begin{tabular}{c|c|c|c|c|c|c|c}
\hline
\multirow{2}{*}{Point Cloud} & Predlift & RAHT & \multicolumn{5}{c}{MS-GAT} \\ \cline{2-8} 
             & Eec\&Dec      & Eec\&Dec & Part      & Y        & U        & V         & Comb  \\ \hline
Longdress   &0.72            &0.47 & 1.53      & 57.54    & 56.86    & 56.95     & 0.26\\
Redandblack &0.69            &0.44 & 1.46      & 56.07    & 58.13    & 54.71     & 0.23\\
Soldier     & 1.03           &0.64 &2.12       & 83.05    & 82.17    & 83.14     & 0.07\\
Dancer      & 2.62           &1.42 & 5.00      & 173.68   & 173.79   & 174.28    & 0.25 \\
Model       & 2.31           &1.44 & 4.93      & 173.23   & 173.83   & 174.00    & 0.17 \\
Andrew      &0.27            &0.15 & 0.50      & 23.30    & 23.08    & 21.08     & 0.10  \\
David       &0.33            &0.20 & 0.65      & 25.24    & 26.69    & 26.96     & 0.04  \\
Phil        &0.36            &0.22 & 0.86      & 29.26    & 27.37    & 27.88     & 0.03  \\
Ricardo     &0.20            &0.19 & 0.43      & 19.09    & 17.61    & 17.34     & 0.06 \\
Sarah       &0.30            &0.17 & 0.59      & 24.96    & 24.25    & 24.18     & 0.03  \\ \hline
Average     &0.88            &0.53& 1.81      & 66.54    & 66.38    & 66.05     & 0.12  \\ \hline
\end{tabular}
\end{table*}
\subsubsection{Implementation Details}
We convert the input attributes (r, g, b) into (y, u, v). 
Three models with the same structure are trained for each component of (y, u, v). 
Each component is processed independently and identically. 
Three restored components are concatenated together and converted back to (r, g, b). 
As illustrated in Fig~\ref{fig:network}, the number of output feature maps of all the layers of the multi-scale feature extraction module is set to 64. 
The numbers of output feature maps of the five graph convolutional layers in the attribute reconstruction module are set to 512, 256, 128, 64, and 1, respectively. 
We train four models with four compression qualities. 
The quantization parameters are set to 51, 46, 40, and 34, respectively, as the common test condition defined in ~\cite{schwarz2018common}. 
The learning rate is set to $10^{-5}$. 
The batch size is set to $8$. 
The network is optimized by Adam~\cite{kingma2014adam} and the parameters $\beta_1$ and $\beta_2$ are set to $0.9$ and $0.999$, respectively.

To validate the efficiency of our proposed algorithm, we apply it to the point cloud attribute compression algorithms Predlift and RAHT in the G-PCC reference software TMC13v12. For Predlift-compressed point clouds, we use the model trained by the training dataset generated by Predlift. For RAHT-compressed point clouds, we use the model trained by the training dataset generated by RAHT.
We use the bits per point (bpp) to measure the total bit cost of Y, U, and V components for each point.
The peak signal-to-noise ratio (PSNR) of Y, U, and V components~\cite{tian2017evaluation} and a compound YUV PSNR of all components~\cite{ohm2012comparison} are used as the quality metrics.
We use the Bjøntegaard delta bit rate (BDBR)~\cite{bjontegaard2001calculation} to compare the respective performance.
 \begin{figure*}
\begin{center}
   \subfigure[Model]{
  \includegraphics[width=0.3\linewidth]{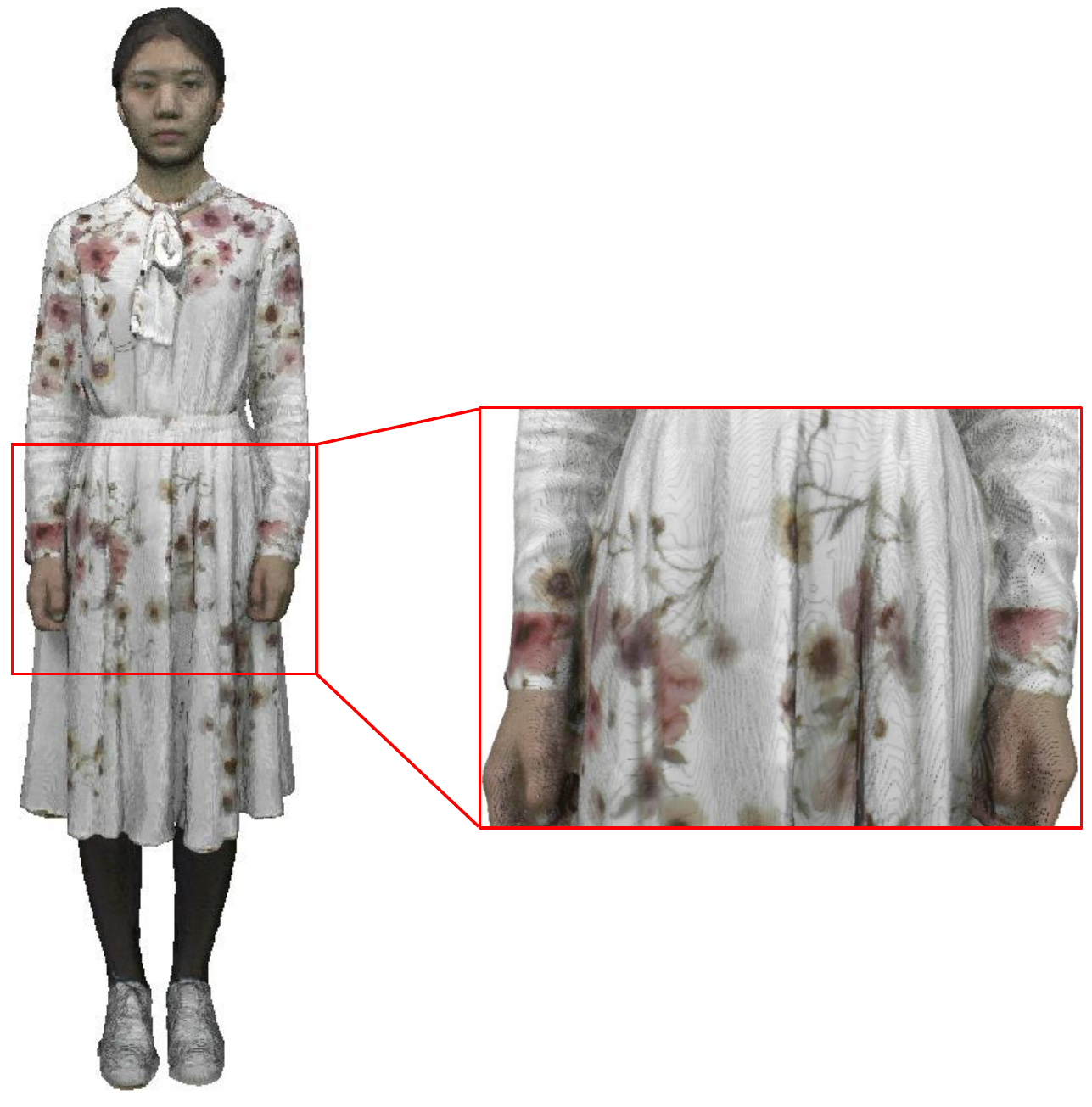}
  }
   \subfigure[bitrate: 0.027 bpp, Y-PSNR: 30.01 dB]{
  \includegraphics[width=0.3\linewidth]{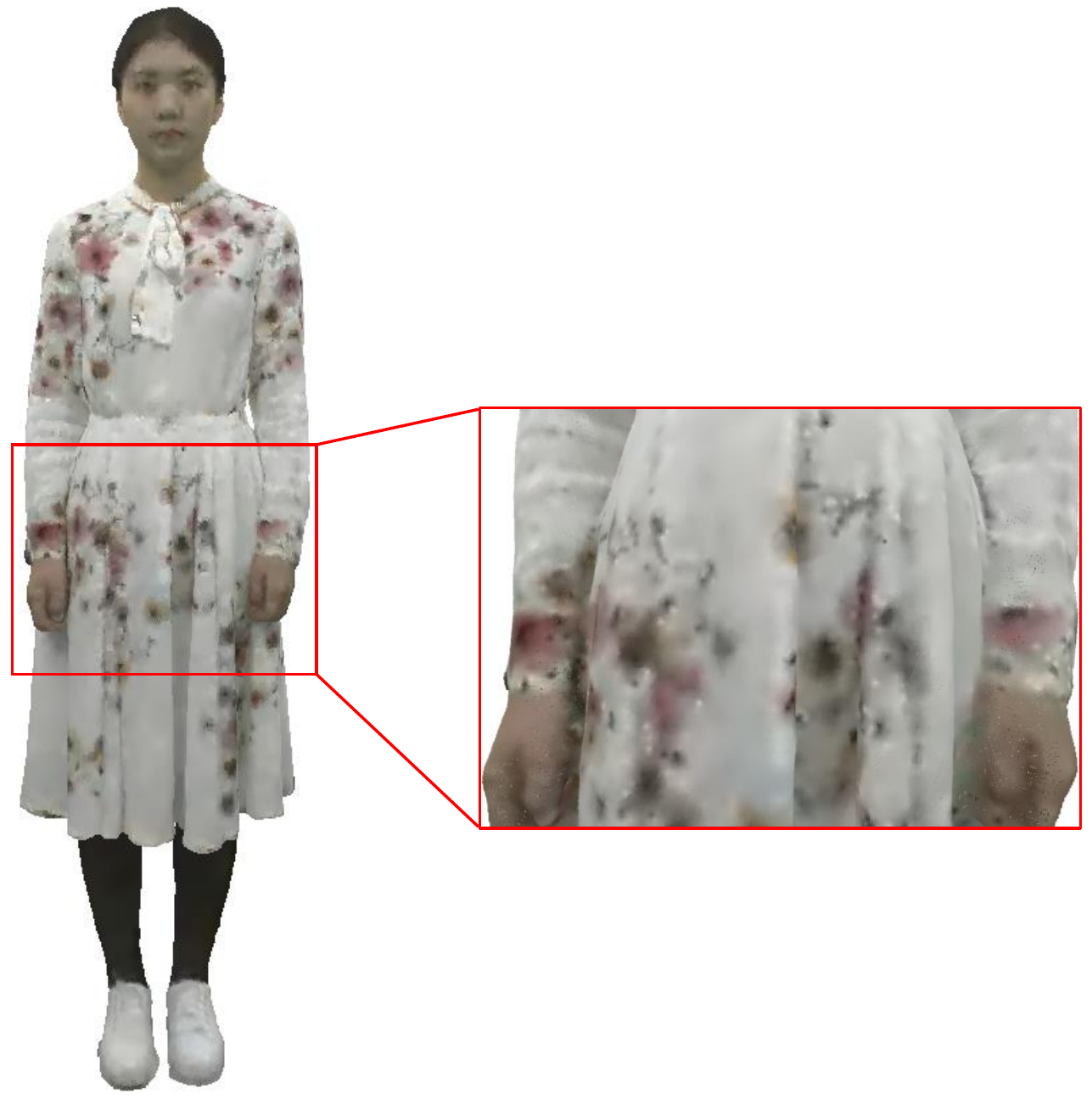}
  }
   \subfigure[bitrate: 0.027 bpp, Y-PSNR: 30.43 dB]{
  \includegraphics[width=0.3\linewidth]{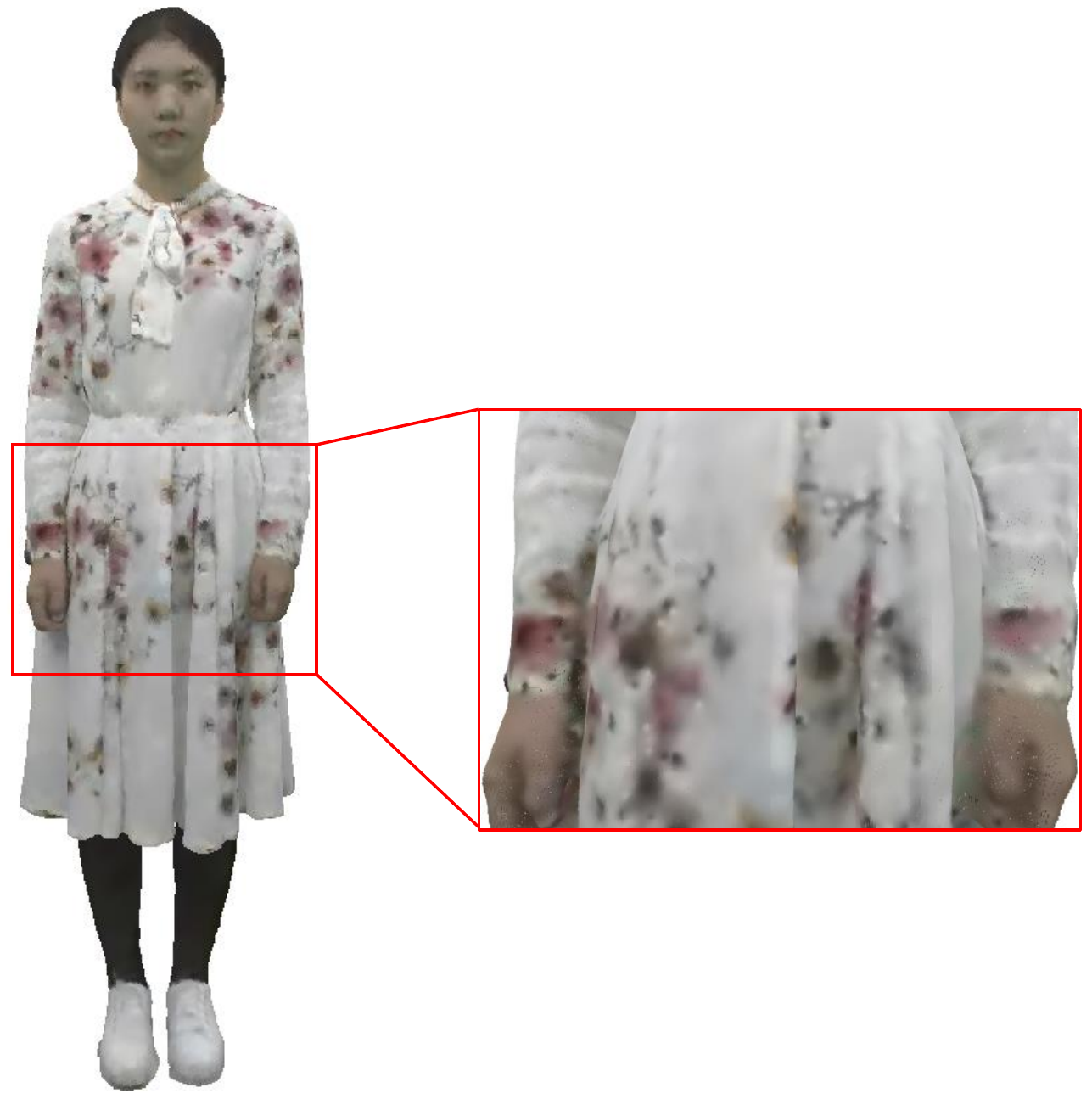}
  }

   \subfigure[Dancer]{
  \includegraphics[width=0.3\linewidth]{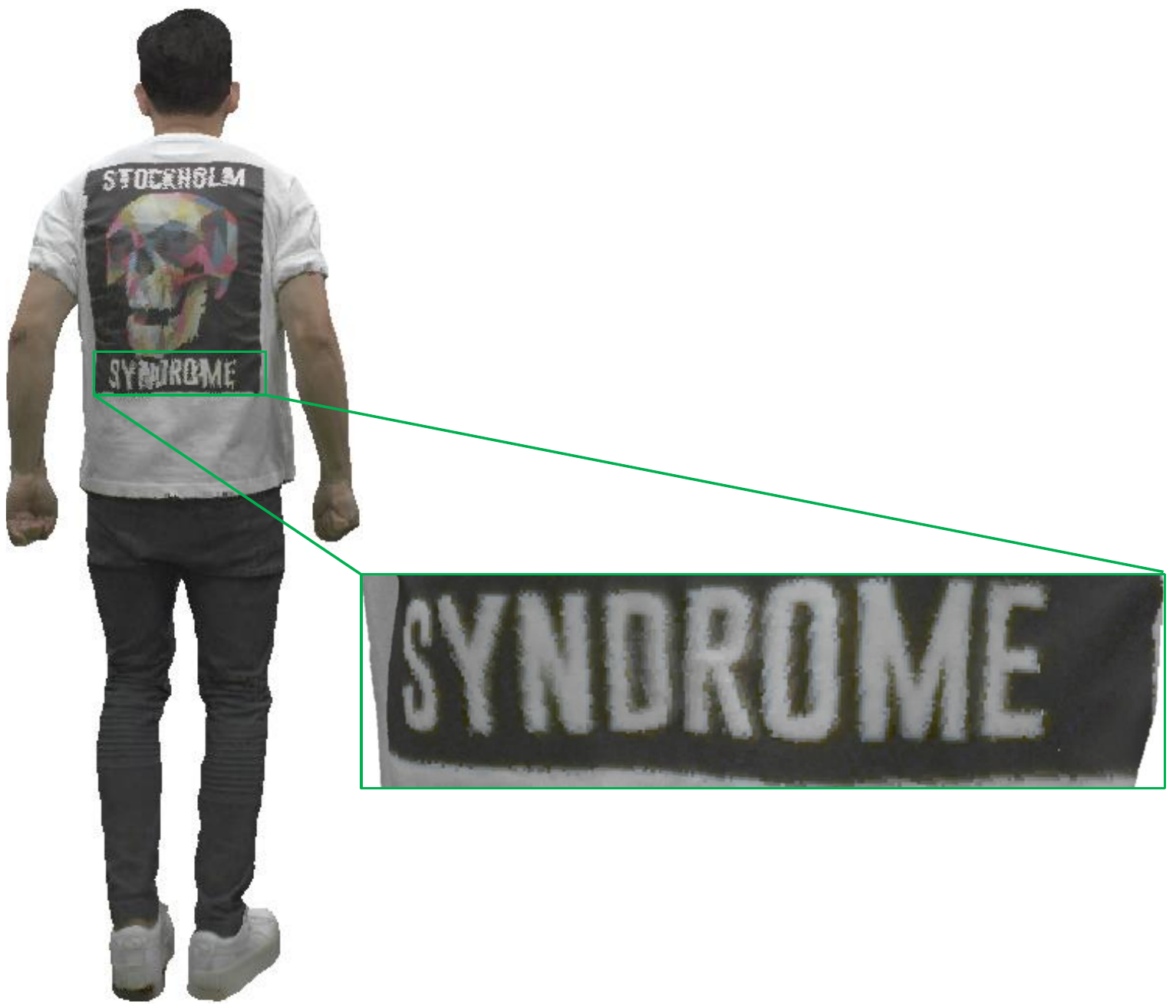}
  }
   \subfigure[bitrate: 0.019 bpp, Y-PSNR: 31.84 dB]{
  \includegraphics[width=0.3\linewidth]{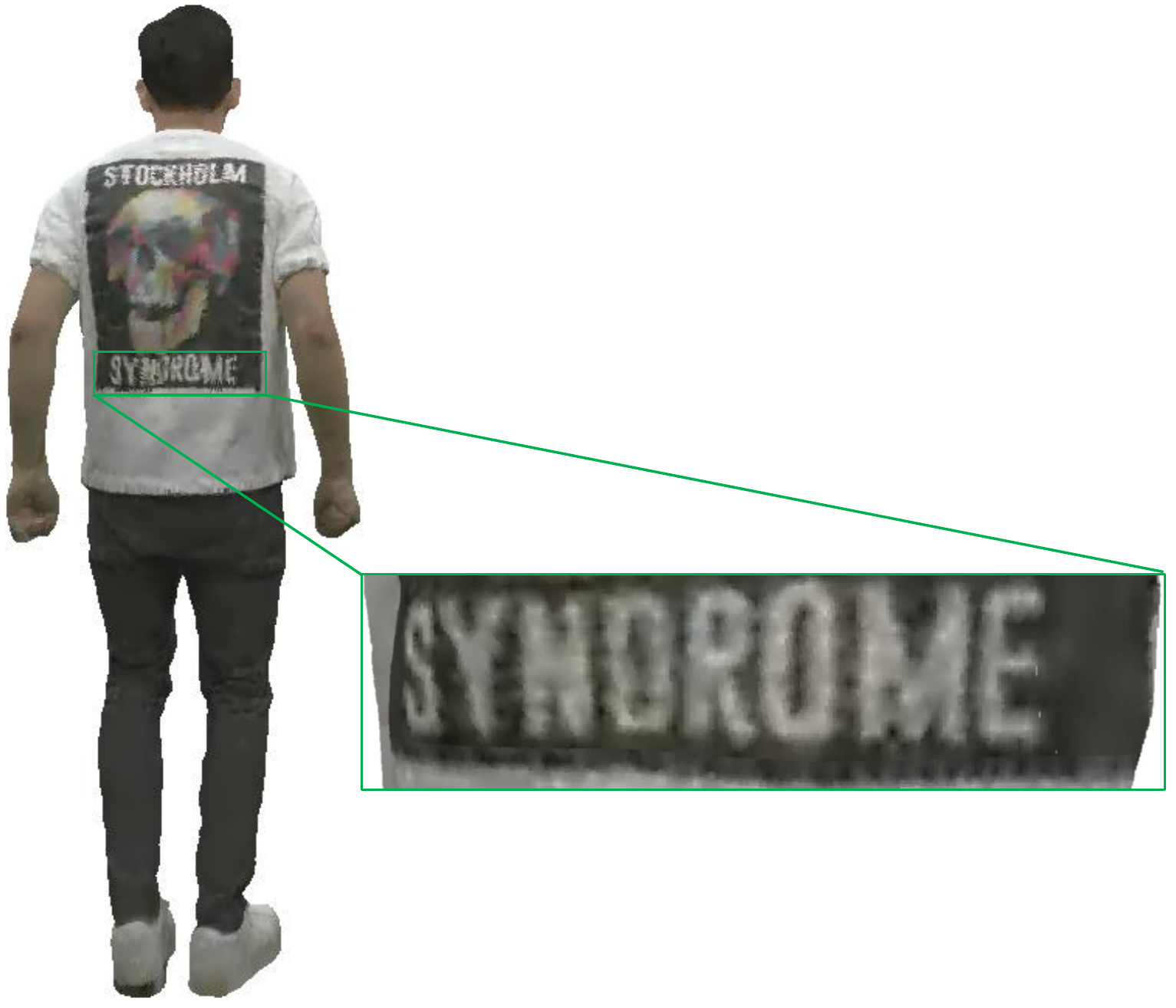}
  }
   \subfigure[bitrate: 0.019 bpp, Y-PSNR: 32.37 dB]{
  \includegraphics[width=0.3\linewidth]{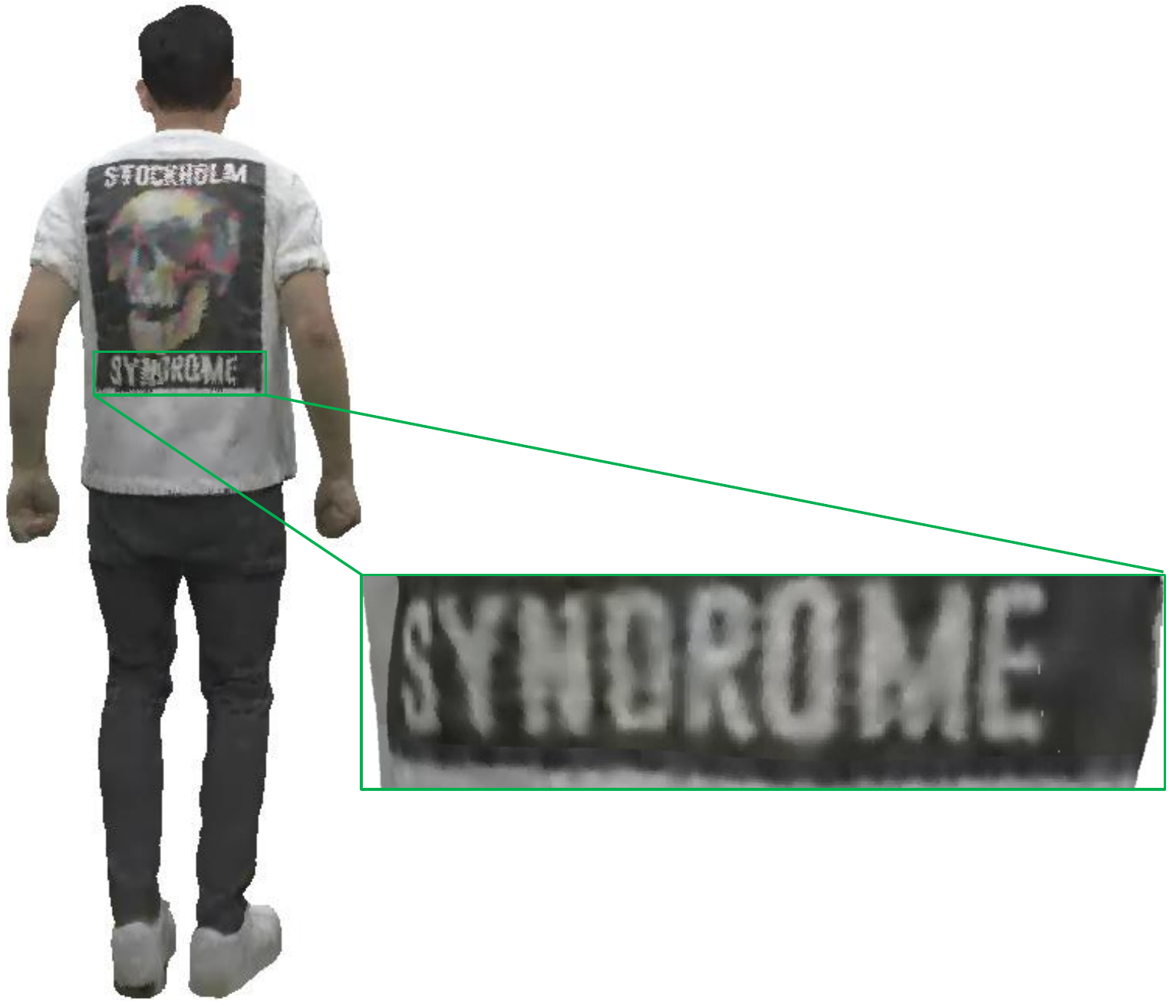}
  }

  \subfigure[Longdress]{
  \includegraphics[width=0.28\linewidth]{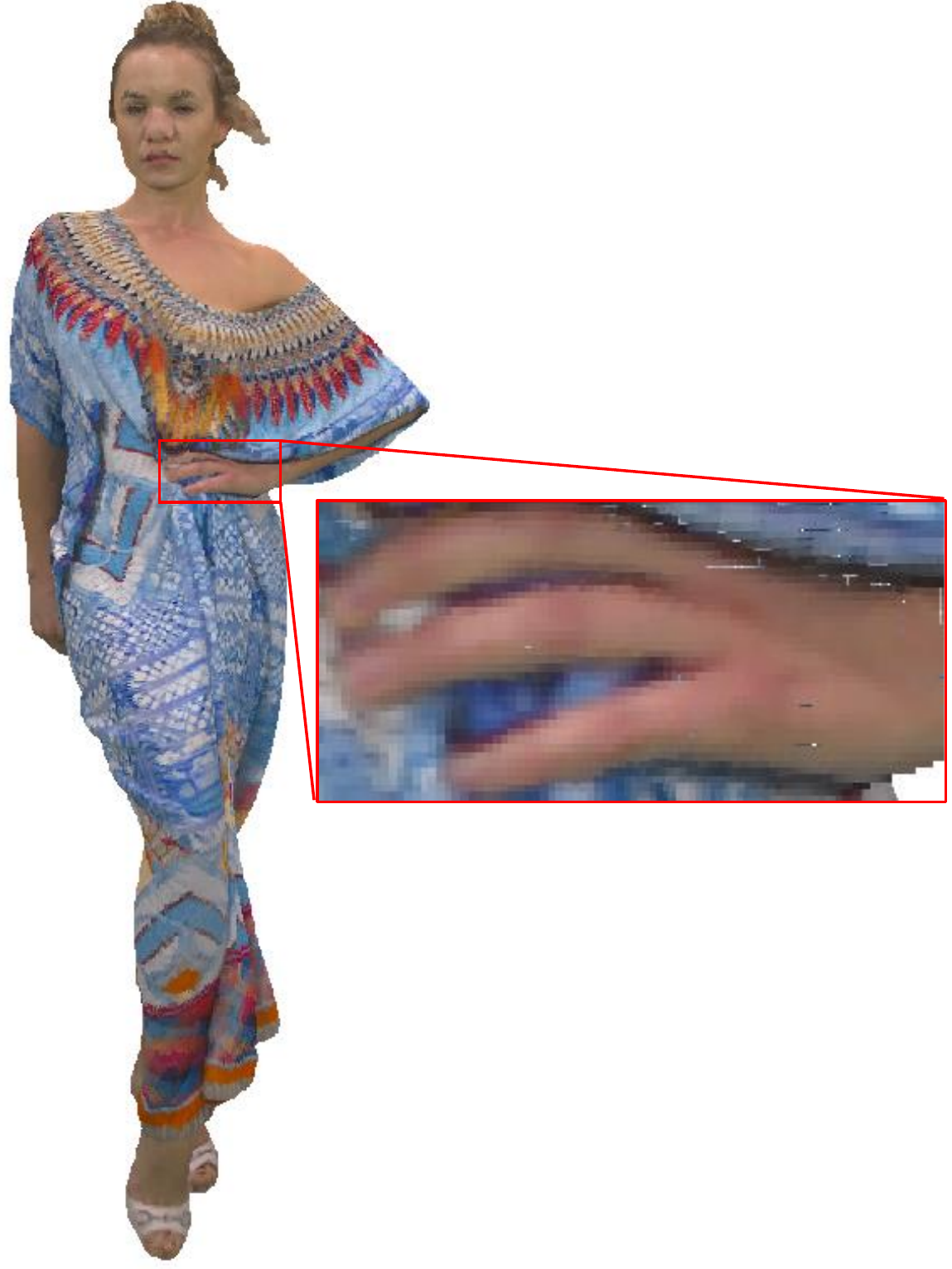}
  }
   \subfigure[bitrate: 0.379 bpp, Y-PSNR: 30.11 dB]{
  \includegraphics[width=0.28\linewidth]{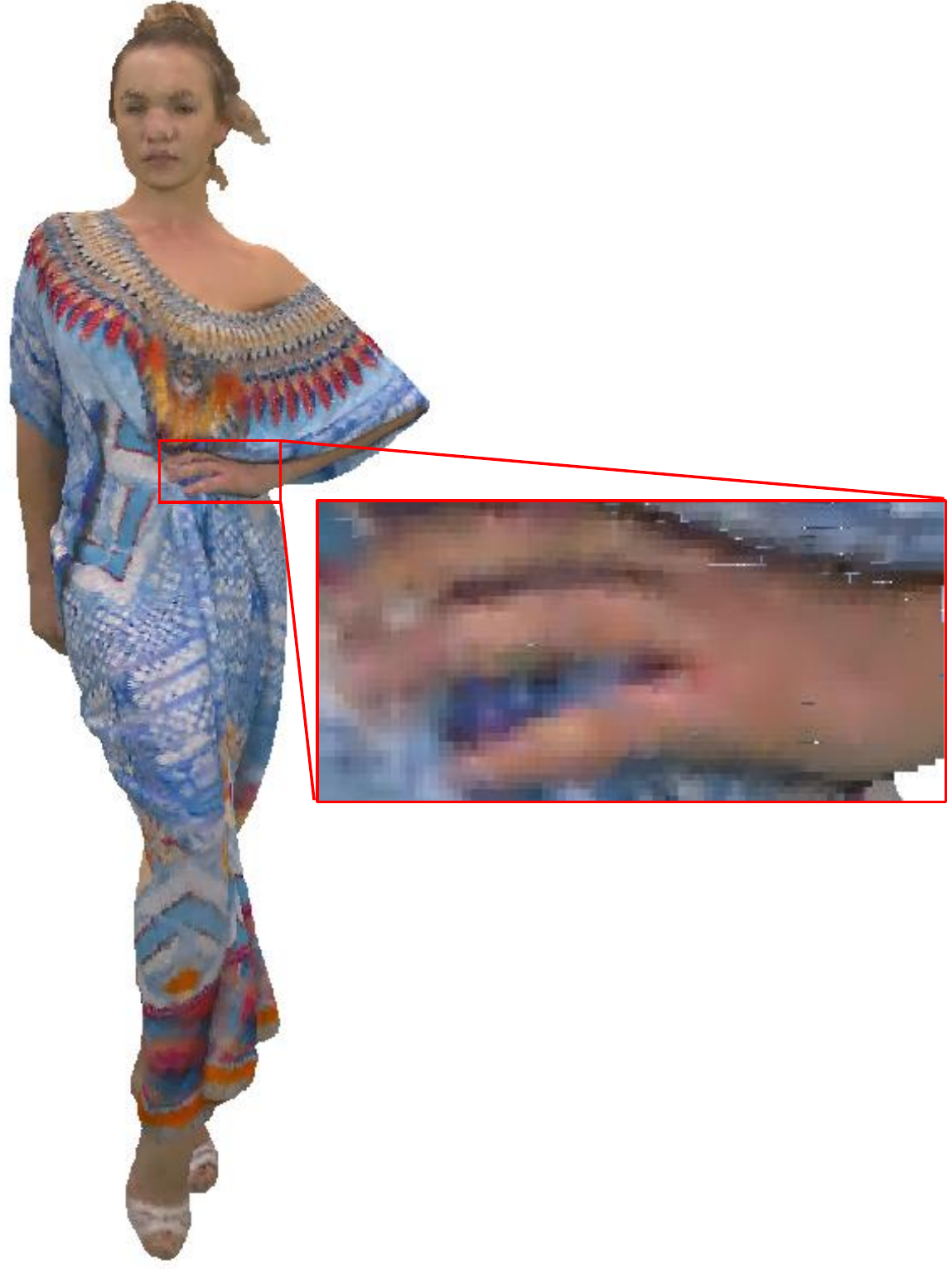}
  }
   \subfigure[bitrate: 0.379 bpp, Y-PSNR: 30.67 dB]{
  \includegraphics[width=0.28\linewidth]{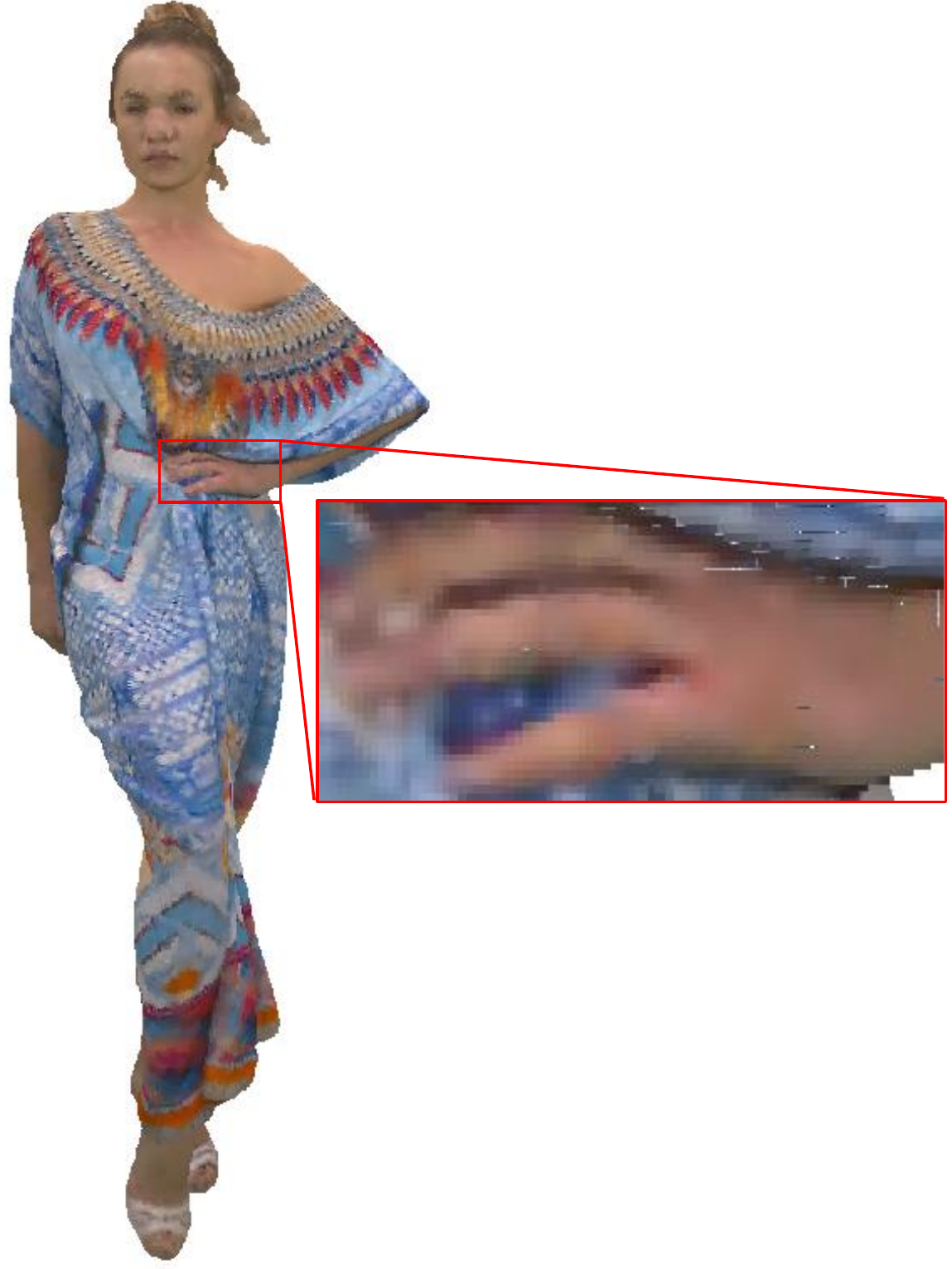}
  }
   \subfigure[Phil]{
  \includegraphics[width=0.3\linewidth]{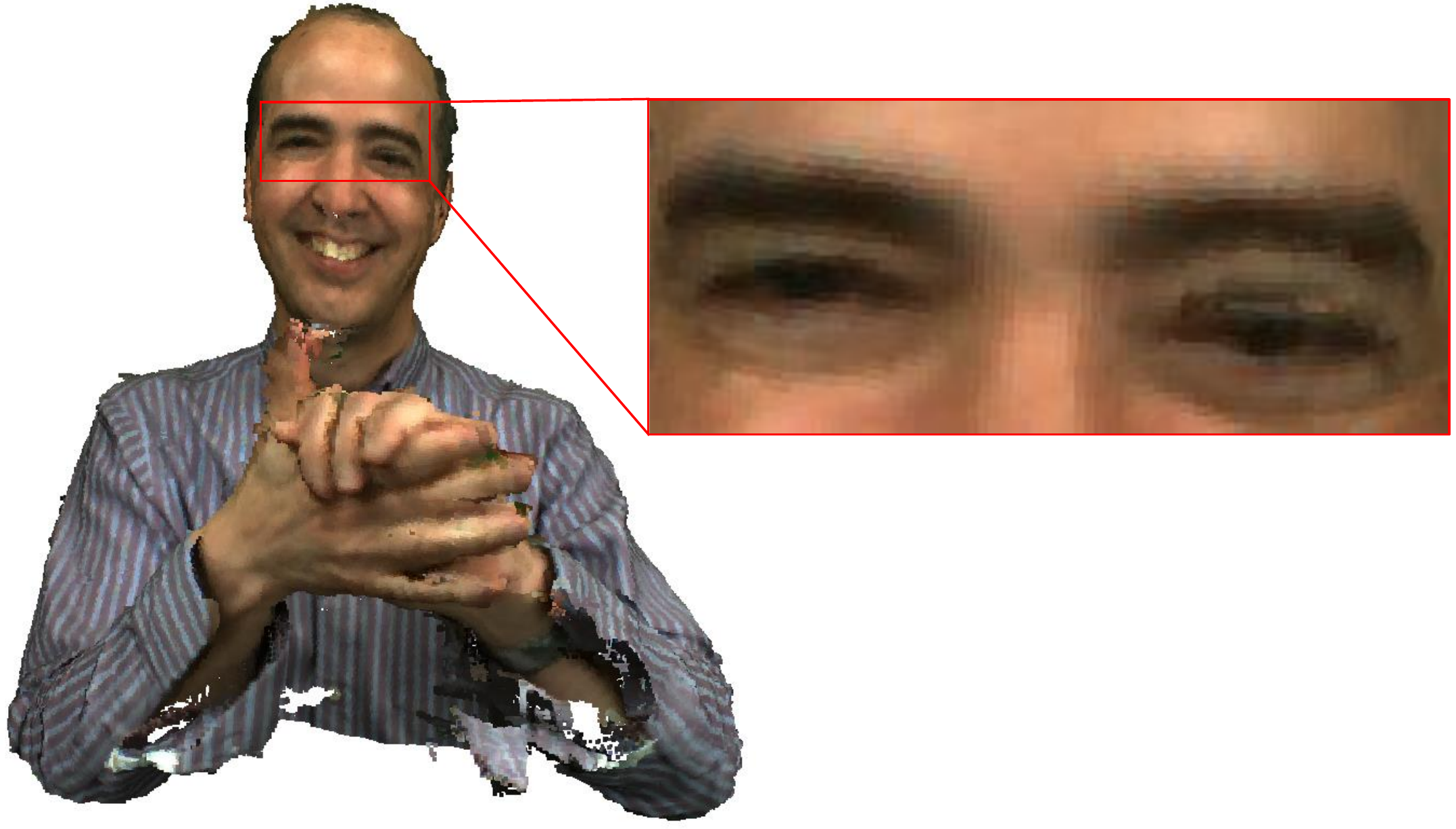}
  }
   \subfigure[bitrate: 0.273 bpp, Y-PSNR: 30.78 dB]{
  \includegraphics[width=0.3\linewidth]{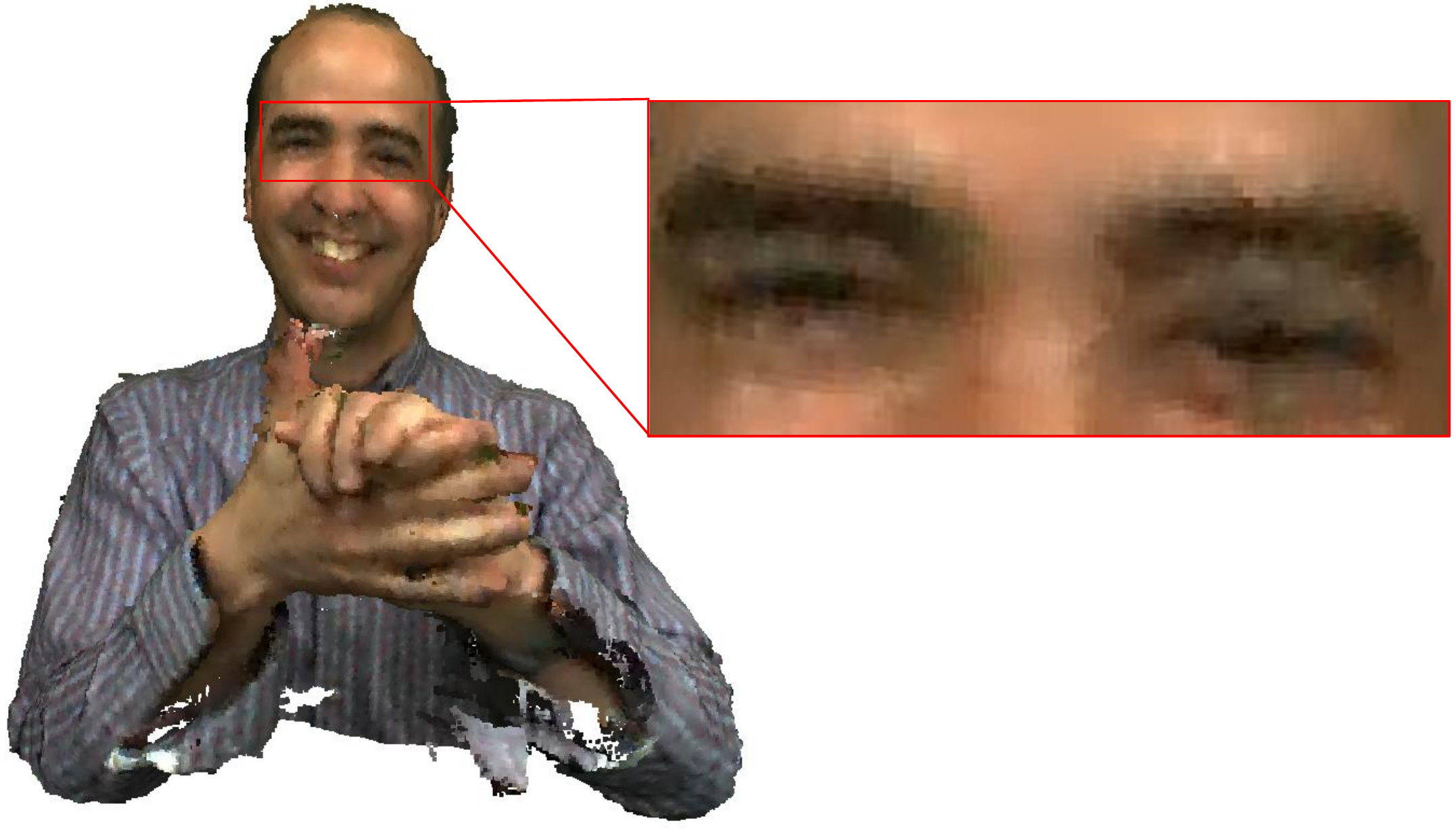}
  }
   \subfigure[bitrate: 0.273 bpp, Y-PSNR: 31.43 dB]{
  \includegraphics[width=0.3\linewidth]{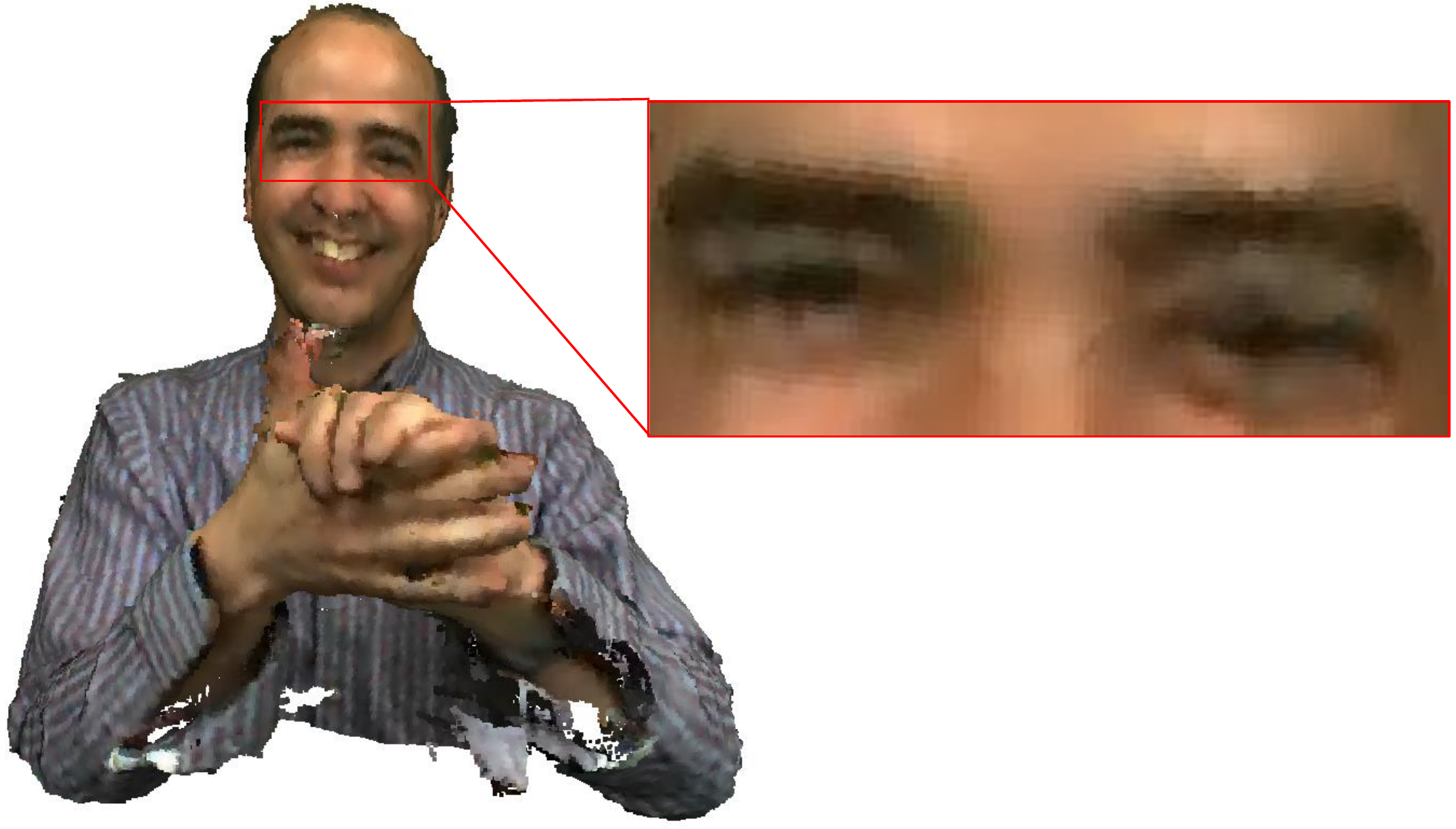}
  }
\end{center}
  \caption{Subjective quality comparison results. Left: Original point clouds. Center: Point clouds compressed by Predlift. Right: Predlift-compressed point clouds restored by our proposed MS-GAT.}
  \label{fig:subjective}
\end{figure*}

\subsection{Objective Quality Comparison}
The objective quality comparisons between the Predlift with and without the proposed artifacts removal algorithm are listed in Table~\ref{table:bdrate_Predlift}. 
We can see that the proposed method obtains 10.97\%, 2.10\%, and 2.81\% BD-rate reduction for the Y, U, and V components, respectively.
On average, a 9.74\% compound YUV BD-rate reduction is achieved.
Fig.~\ref{fig:RDcurve_Y_Predlift} and Fig.~\ref{fig:RDcurve_YUV_Predlift} show the R-D curves with different quality metrics on the test dataset. 
The objective quality comparisons between RAHT with and without MS-GAT are listed in Table~\ref{table:bdrate_RAHT}. The proposed method obtains 10.76\%, 6.14\%, and 8.83\% BD-rate reduction for the Y, U, and V components, respectively. A 10.13\% compound YUV BD-rate reduction is achieved. Fig.~\ref{fig:RDcurve_Y_RAHT} and Fig.~\ref{fig:RDcurve_YUV_RAHT} show the corresponding R-D curves with different quality metrics.

\subsection{Subjective Quality Comparison}
To show the benefits of the proposed MS-GAT in terms of subjective quality, we compare the visual qualities of the compressed point clouds with that of the restored point clouds, as shown in Fig.~\ref{fig:subjective}. We take the point cloud compressed by Predlift as examples. 
At lower bitrates, the point cloud attributes compressed by Predlift suffer severe color shift and blurring. 
For example, the colors of the dress of the compressed ``Model'' are completely changed while the color shift of the restored ``Model'' is reduced. In addition, the letters of the shirts of the compressed ``Dancer'' are blurred  while the restored ones are clear.
At higher bitrates, quantization noise becomes the main compression artifact. 
For example, the hand of compressed ``Longdress"  and the eyes of compressed ``Phil" present more noise. The noise is effectively reduced in the point clouds restored by our method.
\begin{figure*}
\begin{center}
  \includegraphics[width=\linewidth]{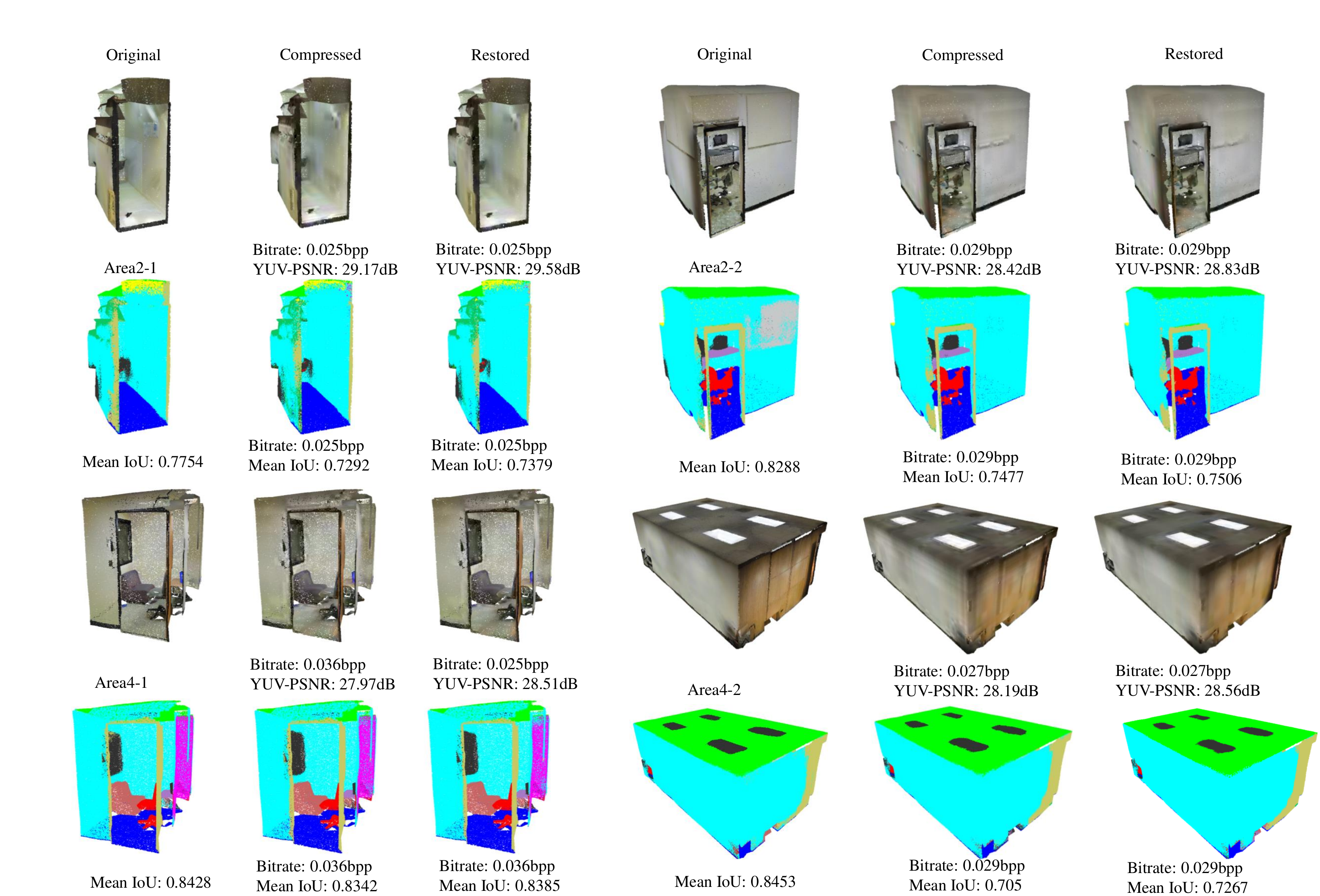}
\end{center}
 \caption{Subjective and semantic segmentation results of original point clouds, point clouds compressed by RAHT, and point clouds restored by our proposed.}
  \label{fig:segmentation}
\end{figure*}
\begin{figure*}
\centering
  \includegraphics[width=0.234\linewidth]{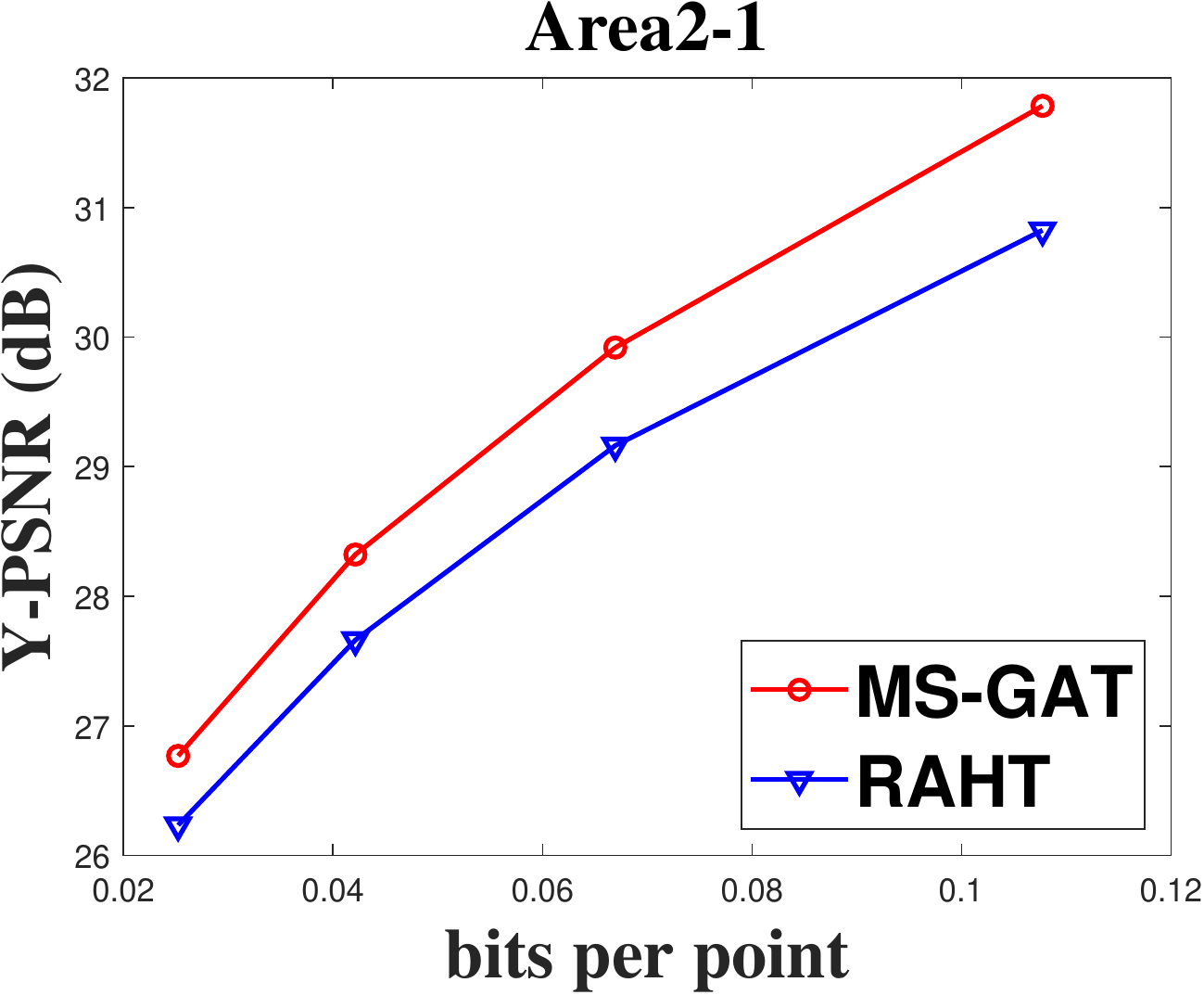}
  \includegraphics[width=0.234\linewidth]{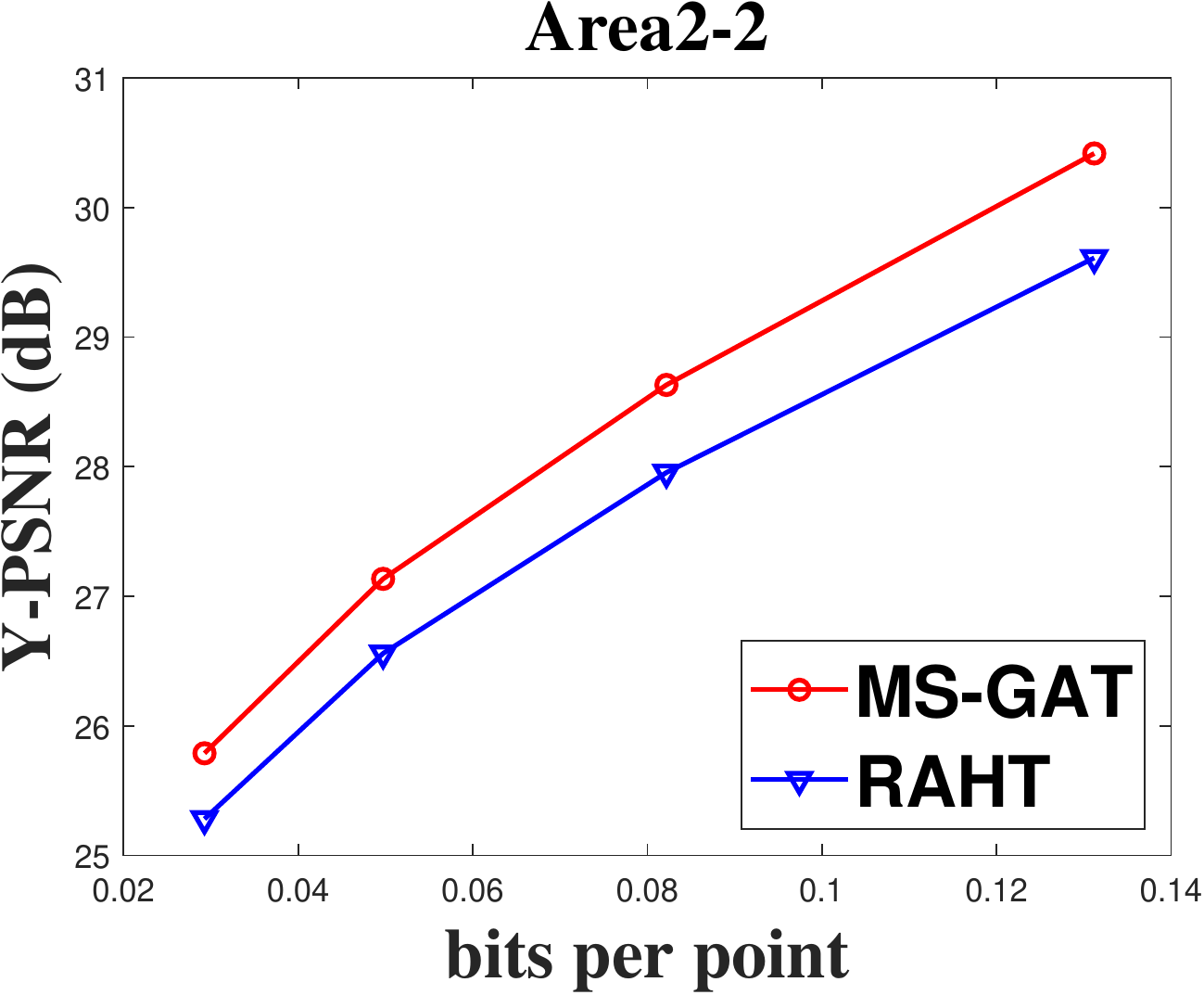}
  \includegraphics[width=0.234\linewidth]{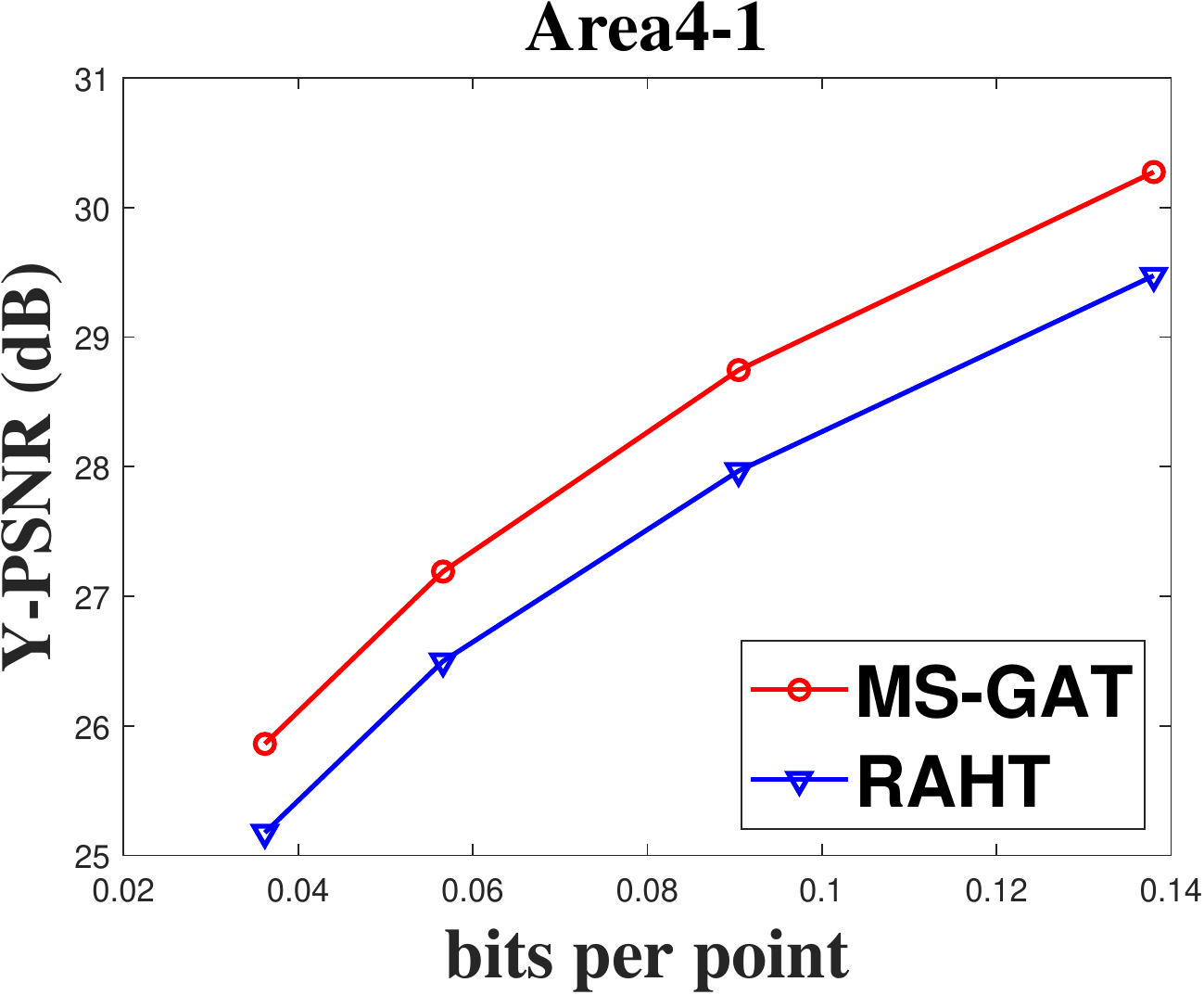}
  \includegraphics[width=0.234\linewidth]{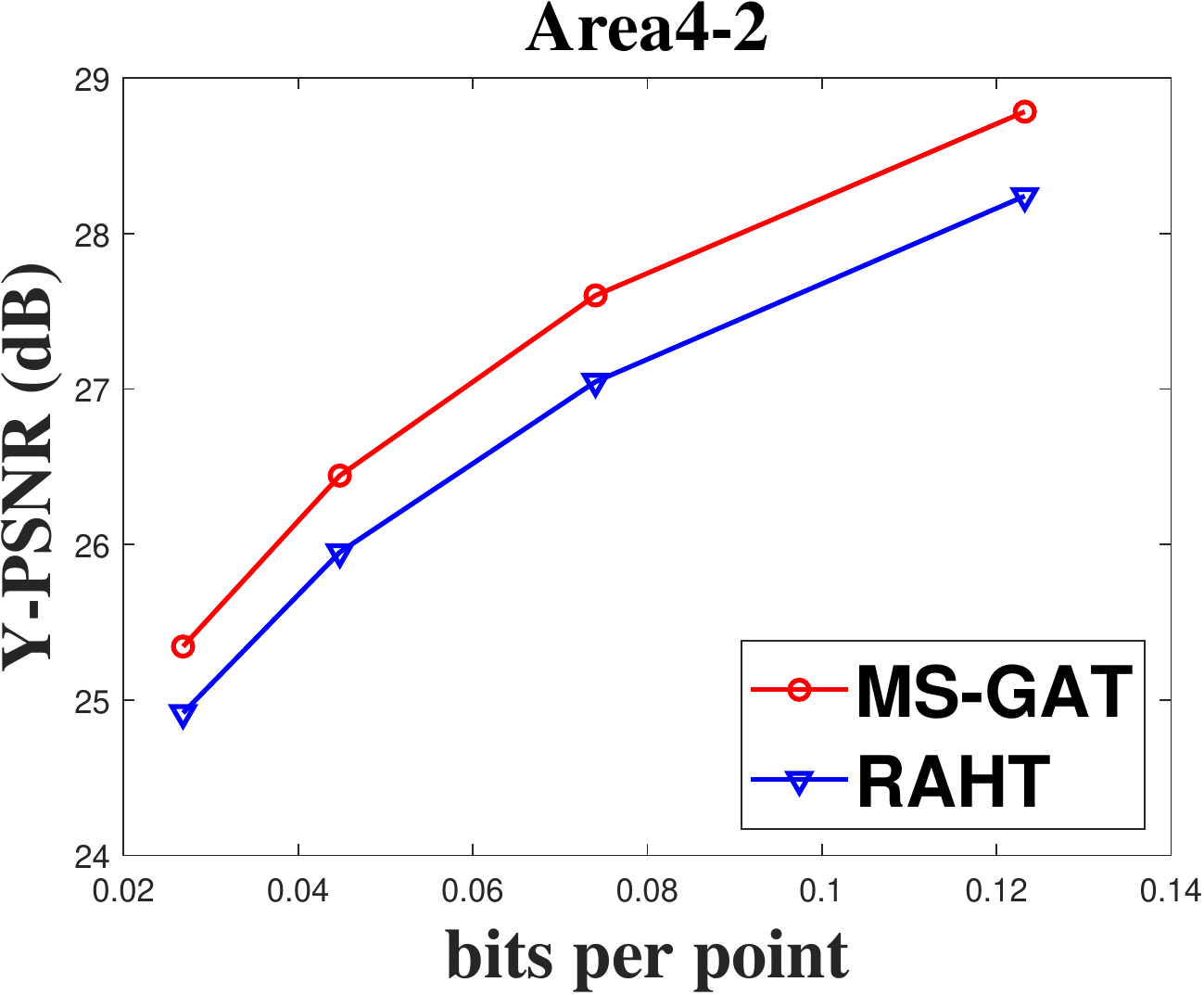}
  \caption{Rate-distortion curves for RAHT with and without our proposed algorithm on the Area2 and Area4 test datasets. The quality is measured by Y PSNR.}
  \label{fig:RDcurve_Area24_Y}
\end{figure*}
\begin{figure*}
\centering
  \includegraphics[width=0.234\linewidth]{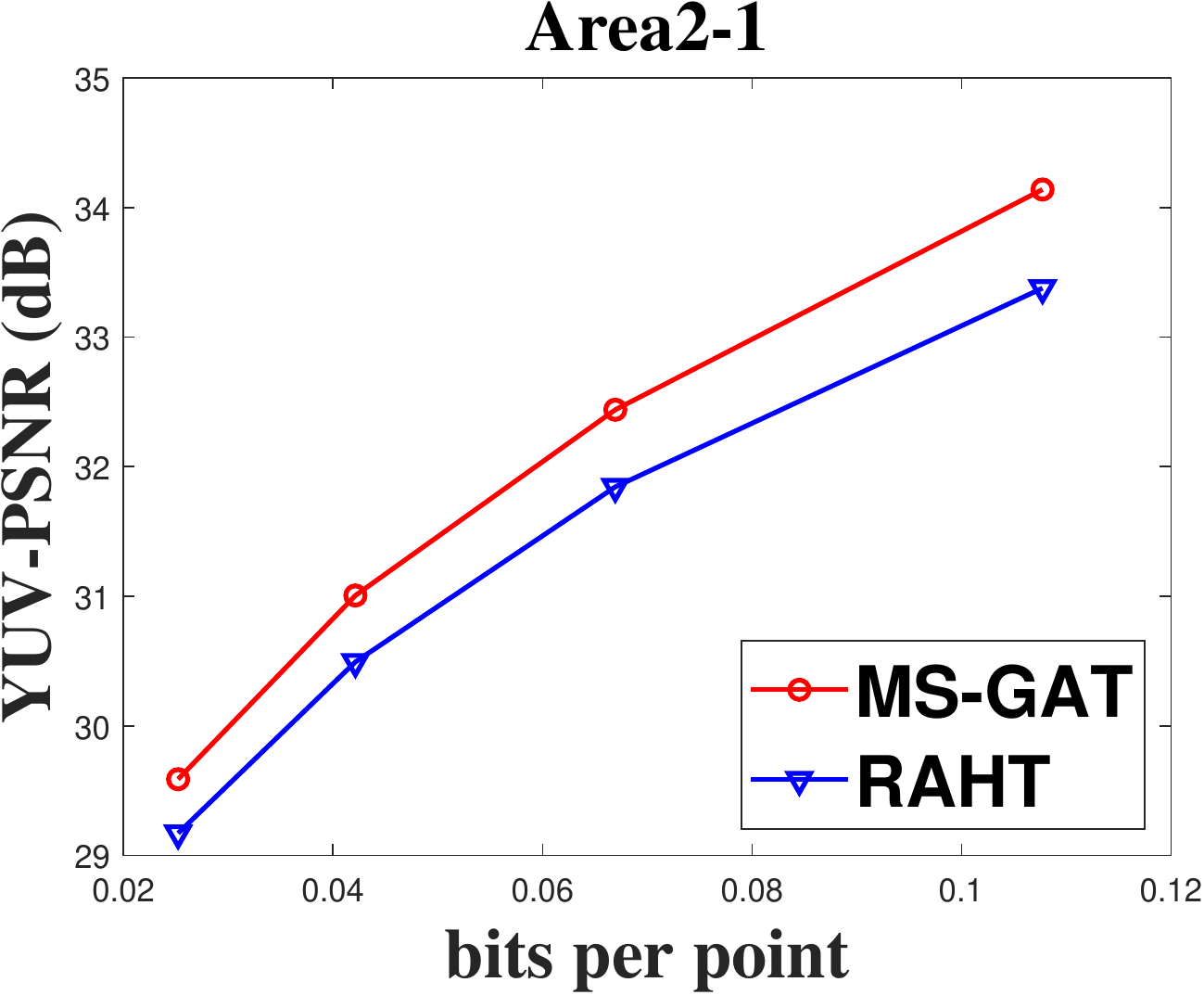}
  \includegraphics[width=0.234\linewidth]{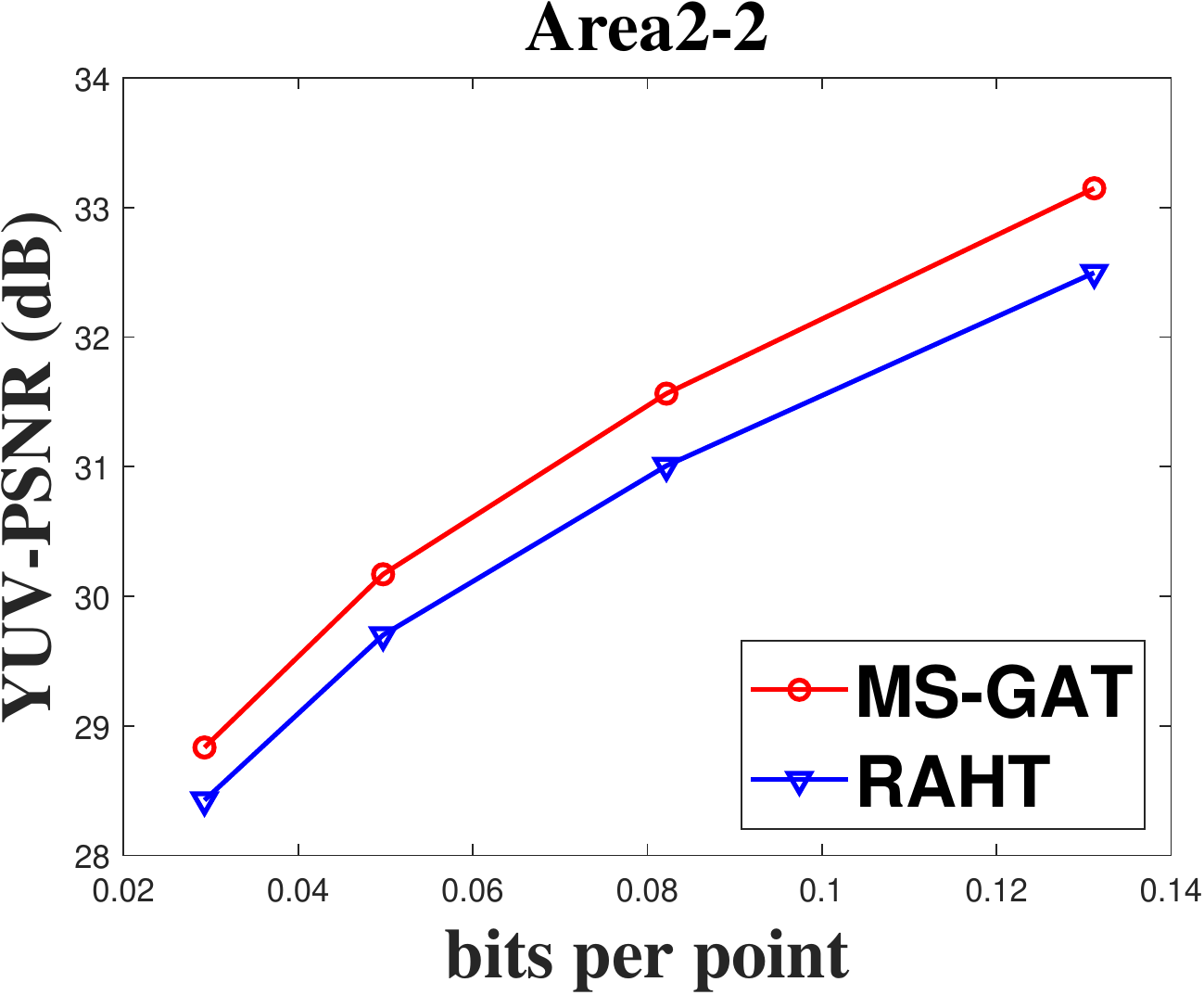}
  \includegraphics[width=0.234\linewidth]{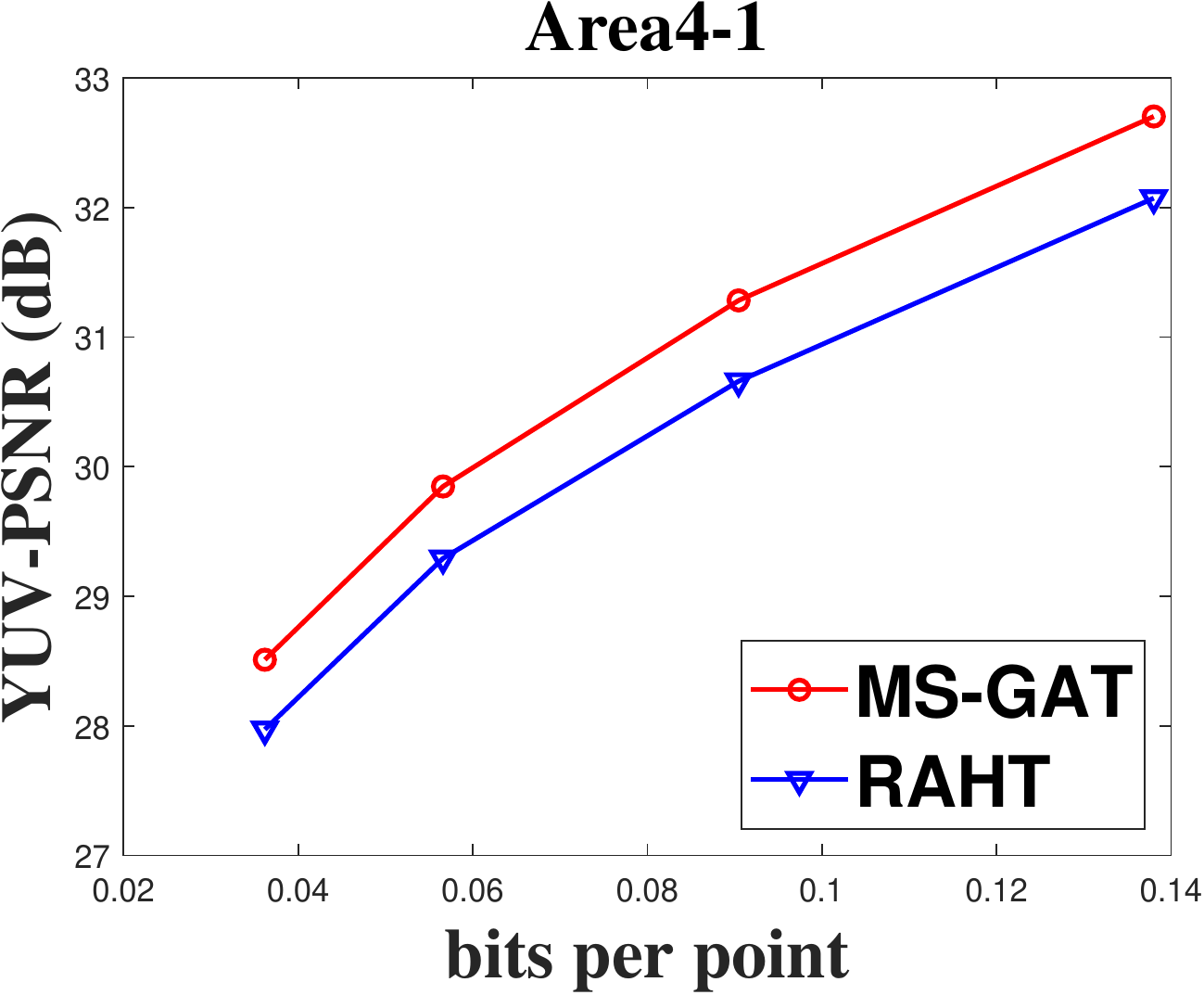}
  \includegraphics[width=0.234\linewidth]{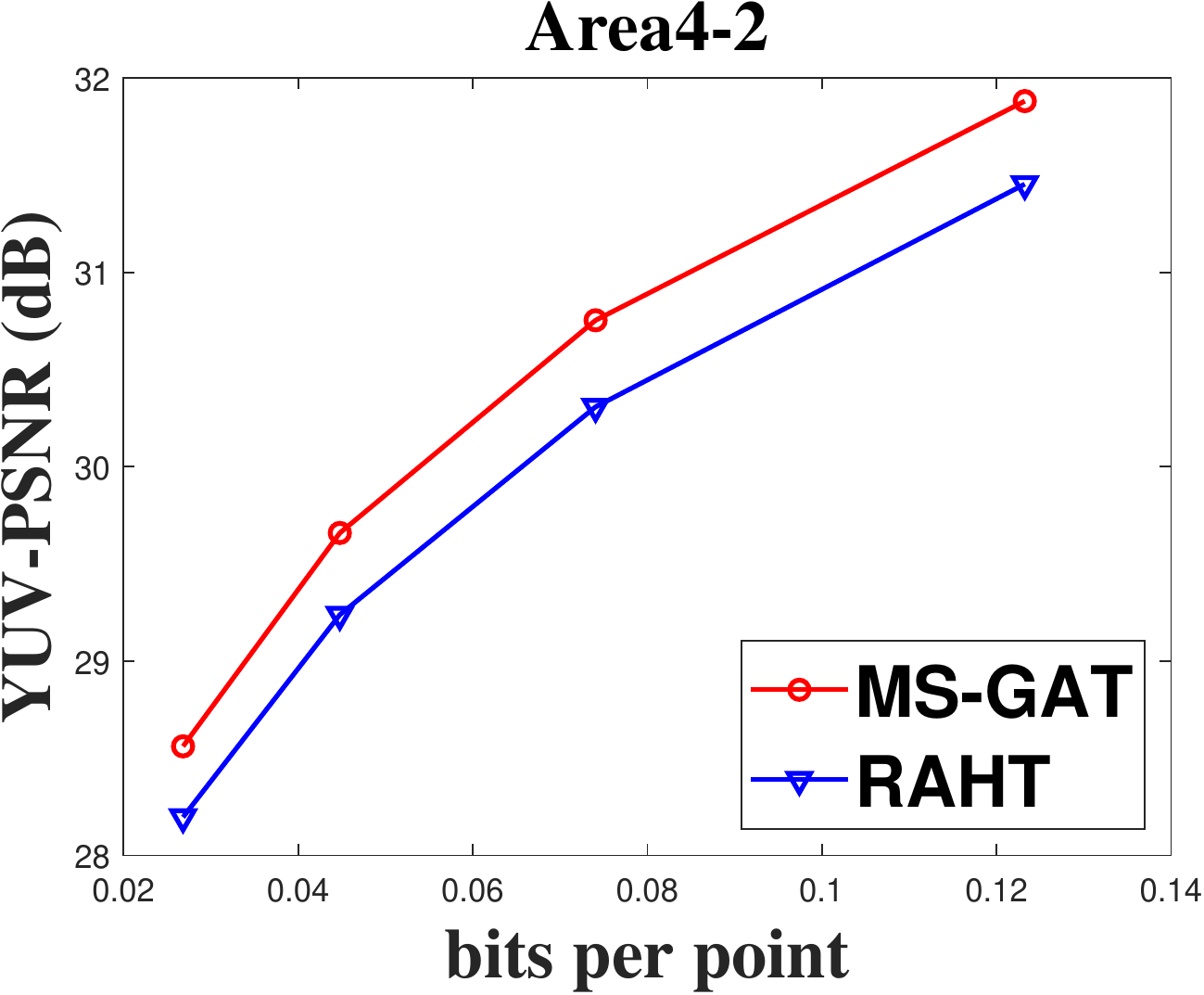}
  \caption{Rate-distortion curves for RAHT with and without our proposed algorithm on the Area2 and Area4 test datasets. The quality is measured by YUV PSNR.}
  \label{fig:RDcurve_Area24_YUV}
\end{figure*}
\subsection{Application to Point Cloud Segmentation}
The data compression may affect the performance of the downstream tasks, which has been demonstrated in image and video compression. 
As we know, point clouds are widely used in 3D segmentation, 3D detection, and other downstream 3D high-level tasks. 
Therefore, the effect on the performance of 3D downstream tasks is an important metric for point cloud compression. 
We take the point cloud semantic segmentation task as an example and quantify the effect on it to demonstrate the benefits of our proposed MS-GAT. 
We use ``Area1'', ``Area3'' and ``Area6'' in Stanford 3D Large-Scale Indoor Spaces Datasets~\cite{armeni20163d} to train a new model and test on ``Area2'' and ``Area4''. 
We use the pre-trained PointNet++~\cite{qi2017pointnet++} model\footnote{https://github.com/yanx27/Pointnet\_Pointnet2\_pytorch.} to segment the point clouds in ``Area2'' and ``Area4'' compressed by RAHT in TMC13v12 and those restored by our MS-GAT. The segmentation performance is measured by mean intersection-over-union (IoU).

The semantic segmentation and subjective results are illustrated in Fig.~\ref{fig:segmentation}. From the results, we observe that RAHT leads to banding and blurring artifacts to the point cloud attributes. The artifacts of the point cloud attributes bring a significant drop to the semantic segmentation performance, even though the geometry keeps unchanged. For example, the board colored by grey on the wall of the ``Area2-2" is  wrongly predicted to be a part of the wall since the compressed board is blurred. Comparing the mean IoU values and the illustrations of semantic segmentation, our proposed method can slightly alleviate the impact of point cloud attribute compression on semantic segmentation. For example, the door colored by brown of the compressed ``Area2-1" is wrongly predicted to the column which is colored by purple. While the wrong prediction is alleviated in the restored point cloud. We also present the corresponding objective results in Fig.~\ref{fig:RDcurve_Area24_Y} and Fig.~\ref{fig:RDcurve_Area24_YUV} to verify the authenticity of the experimental results. 
This experiment reveals that the point cloud attributes play an important role in downstream point cloud high-level tasks. 
It is suggested that researchers should consider both the compression performance and downstream task performance when designing point cloud attribute compression methods.

\subsection{Analysis of Computational Complexity}
To analyze the computational complexity quantitatively, we compare the run time of Predlift in TMC13v12 and our proposed MS-GAT on all point clouds in the test dataset, as shown in Table~\ref{table:time}. For Predlift, we count the color encoding and decoding time. For MS-GAT, we count the block partition time, block combination time, and the artifacts removal time for Y, U, and V components.
For a fair comparison, the run time does not include the time for reading and writing point cloud files. 
We accumulate the run time of all blocks although they can be processed in parallel. 
The proposed MS-GAT is implemented with TensorFlow 1.13 in eager mode. 
The number of total parameters of MS-GAT is 1.98M. 
It is run on one NVIDIA GeForce 1080 Ti GPU. 
The results show that processing each component separately significantly increases the computational cost. 
Processing each block serially is another reason for excessive runtime.
We will explore processing each component simultaneously and process the partitioned blocks in parallel in the future.

\begin{table}[]
\centering
\caption{Influence of the number of scales of multi-scale feature extraction module. }
\label{table:bdrate_scale}
\begin{tabular}{c|c|c|c}
\hline
PointCloud  & OneScale & TwoScale & ThreeScale \\ \hline
Longdress   & -11.23   & -11.78   & -12.28    \\ 
Redandblack & -6.63    & -7.51    & -7.36    \\ 
Soldier     & -10.81   & -11.13   & -11.61    \\ 
Dancer      & -13.79   & -13.33   & -13.46    \\
Model       & -13.27   & -13.07   & -13.30    \\ 
Andrew      & -1.51    & -5.46    & -8.23    \\ 
David       & -4.50    & -6.91    & -7.97    \\
Phil        & -9.45   & -12.66   & -13.54     \\
Sarah       & -10.54   & -11.27   & -11.96    \\
Ricardo     & -3.75    & -6.13    & -7.53    \\ \hline
Average     & -8.55   & -9.93   & -10.74    \\ \hline
\end{tabular}
\end{table}

\begin{table}[]
\centering
\caption{Effectiveness of the quantization steps. }
\label{table:bdrate_qs}
\begin{tabular}{c|c|c|c}
\hline
PointCloud  & WQS      & WoQSiA   & WoQS \\ \hline
Longdress   & -12.28   & -12.56   & -10.14    \\ 
Redandblack & -7.36    & -8.13    & -8.37    \\ 
Soldier     & -11.61   & -11.50   & -11.45    \\ 
Dancer      & -13.46   & -13.24   & -15.17    \\
Model       & -13.30   & -13.06   & -13.89    \\ 
Andrew      & -8.23    & -6.61    & -5.32    \\ 
David       & -7.97    & -7.37    & -6.10    \\
Phil        & -13.54   & -11.72   & -11.49     \\
Sarah       & -11.96   & -11.97   & -10.47    \\
Ricardo     & -7.53    & -7.38    & -5.06    \\ \hline
Average     & -10.74   & -10.36   & -9.75    \\ \hline
\end{tabular}
\end{table}
\subsection{Ablation Study}
\subsubsection{Effectiveness of Multi-Scale Scheme}
In the point cloud attribute coding of G-PCC, the points from coarser-granularity levels are used to predict those from finer-granularity levels. 
Therefore, one point may be related to points in both near or far distances. Therefore, we propose a multi-scale scheme to capture the short- and long-range correlations between the current point and both near and distant points. 
To demonstrate the benefits of the proposed multi-scale scheme, we design three variants which have only one, two, and three scales. 
We denote the variants to OneScale, TwoScale, and ThreeScale, respectively.
Table~\ref{table:bdrate_scale} illustrates the BD-rate and R-D curves of different methods on the test dataset. 
From the comparison results, we find out that the R-D performance improves along with the increase of the number of scales. 
This is because the network can capture the correlation between various points in both near and far distances, which results in a more accurate estimation of the compression artifacts.

\subsubsection{Effectiveness of Quantization Steps}
Different quantization step per point leads to different degrees of artifacts. 
Therefore, we introduce the quantization step per point as an extra input to the proposed network. In addition, we add the quantization steps into the graph attentional layers as the global guiding weights, which can help the network learn weighted attention to further focus on the points with more artifacts.
To verify the effectiveness of the quantization steps, we design two variants. Specifically, we first remove the quantization steps in the graph attentional layers, which is denoted as WoQSiA. Then we further remove the input quantization steps of the network, which is referred to WoQS. Table~\ref{table:bdrate_qs} presents the BD-rate and R-D curves of different methods on the test dataset. 
We can see that the introduction of quantization steps is beneficial for the point clouds with more complex textures and richer colors, such as ``Longdress" and ``Soldier". 
This is because the quantization steps give the network a global prior to handle different degrees of artifacts for each point to a certain extent. 

\begin{table}[]
\centering
\caption{Effectiveness of the weighted graph attention layer. }
\label{table:bdrate_attention}
\begin{tabular}{c|c|c|c}
\hline
PointCloud  & WAtt     & WSAtt   & WoAtt  \\ \hline
Longdress   & -12.28   & -12.56 & -10.67  \\ 
Redandblack & -7.36    & -8.13  & -5.42   \\ 
Soldier     & -11.61   & -11.50 & -9.74  \\ 
Dancer      & -13.46   & -13.24 & -9.85  \\
Model       & -13.30   & -13.06 & -10.34  \\ 
Andrew      & -8.23    & -6.61  & -4.28   \\ 
David       & -7.97    & -7.37  & -5.40   \\
Phil        & -13.54   & -11.72 & -11.42  \\
Sarah       & -11.96   & -11.97 & -8.61  \\
Ricardo     & -7.53    & -7.38  & -4.27   \\ \hline
Average     & -10.74   & -10.36 & -8.00  \\ \hline
\end{tabular}
\end{table}

\begin{table}[]
\centering
\caption{Influence of the number of layers of the MLP in the weighted graph attention layer and bottleneck layer. }
\label{table:bdrate_mlp}
\begin{tabular}{c|c|c|c}
\hline
PointCloud  & OneLayer & TwoLayer & ThreeLayer \\ \hline
Longdress   & -12.28   & -12.73   & -11.34    \\ 
Redandblack & -7.36    & -8.08    & -7.21    \\ 
Soldier     & -11.61   & -11.72   & -10.77    \\ 
Dancer      & -13.44   & -13.18   & -13.20    \\
Model       & -13.46   & -13.23   & -13.19    \\ 
Andrew      & -8.23    & -7.02    & -5.60    \\ 
David       & -7.97    & -8.33    & -7.42    \\
Phil        & -13.54   & -13.73   & -11.86     \\
Sarah       & -11.96   & -12.75   & -11.20    \\
Ricardo     & -7.53    & -8.93    & -6.71    \\ \hline
Average     & -10.74   & -10.97   & -9.85    \\ \hline
\end{tabular}
\end{table}

\subsubsection{Effectiveness of Weighted Graph Attentional Layer}
Considering that the attributes of various points suffer from different degrees of compression artifacts, we incorporate a weighted graph attentional layer into the network to further pay more attention to the points with larger distortions. 
To verify the effectiveness of the weighted graph attentional layer, we design two variants. We first remove the quantization steps in the weighted graph attentional layer to use a simple graph attentional layer, which is denoted to WSAtt. Then we remove all the graph attentional layers, which is denoted to WoAtt. Table~\ref{table:bdrate_attention} shows the BD-rate and R-D curves of different methods on the test dataset. 
The results show that the graph attentional layer increases the qualities of the decoded point cloud attributes. Compared with the simple graph attentional layer, the weighted graph attentional layer can achieve an additional performance improvement, which is owing to the global guiding weights which help the network focus on the points with more artifacts.
\subsubsection{The Number of Layers of MLP}\label{mlp}
The main component of the weighted graph attentional layers and bottleneck layers is MLP. In our paper, the MLP is a $N$-layer perceptron and each layer is followed by the non-linear activation function ReLU. To evaluate the influence of the number of layers $N$ of a MLP, we design three variants whose MLP contains one, two, and three layers. The variants are denoted to OneLayer, TwoLayer, and ThreeLayer, respectively. Table~\ref{table:bdrate_mlp} presents the BD-rate and R-D curves of the three variants. As observed, two-layer MLP gets the best performance. Therefore, we set $N$ to 2 in our paper.
\subsubsection{Effectiveness of separate training}
We train three models with the same structure for Y, U, and V components in our algorithm. Each component is processed independently and identically. To explore the effectiveness of the separate training strategy, we design a variant of the ``TwoLayer" model in Section~\ref{mlp}. Specifically, the number of the input channels of the variant becomes 4, e.g., Y, U, V, and quantization steps. The number of the output channels of the variant becomes 3, e.g., Y, U, and V. The variant is jointly trained for the three components. The loss function is the average of~\eqref{E13} for each component. The BD-rate comparison between the Predlift with and without the variant is presented in Table~\ref{table:bdrate_Predlift_joint_training}. From the results, we find that a  significant performance drop is brought to the U and V components. We think it is owing to that the characteristics of the three components of the compressed point cloud attributes are quite different. It is difficult for the network to handle them together.
\begin{table}[]
\centering
\caption{BD-rate comparison between Predlift with and without the proposed algorithm when the Y, U, and V components are jointly trained.}
\label{table:bdrate_Predlift_joint_training}
\begin{tabular}{c|c|c|c|c}
\hline
PointCloud  & BDBR-Y & BDBR-U & BDBR-V & BDBR-YUV \\ \hline
Longdress   & -10.58 & 8.20   & 3.43  & -7.10    \\ 
Redandblack & -5.40 & 0.21   & 1.13  & -4.02    \\ 
Soldier     & -9.67 & 9.51  & 22.40  & -6.76   \\ 
Dancer      & -11.04 & -4.25  & 4.61  & -9.73   \\
Model       & -11.21 & -10.40  & -1.43  & -10.42   \\ 
Andrew      & -5.45  & 48.42  & 28.24  & -2.59    \\ 
David       & -8.16  & 5.13  & 8.15  & -5.72   \\
Phil        & -11.70 & -3.59  & -3.04   & -10.86   \\
Sarah       & -10.89 & 0.65  & 10.35 & -7.56    \\
Ricardo     & -7.59  & 2.13  & 5.18 & -5.89    \\ \hline
Average     & -9.17 & 5.60  & 7.90  & -7.07    \\ \hline
\end{tabular}
\end{table}

\section{Conclusion}\label{conclusion}
In this paper, we propose a Multi-Scale Graph Attention Network (MS-GAT) to remove the artifacts of point cloud attributes compressed by G-PCC.
We first build a graph with the assistant of point cloud geometry information and treat the point cloud attributes as graph signals.
We then utilize the Chebyshev graph convolutions to extract the features of attributes. 
Considering that one point may be related to the points with both near and far distances in the attribute compression, we design a multi-scale scheme to capture the short- and long-range correlation between the current point and near or distant points. 
In addition, we observe that the quantization steps for various points are different. 
Therefore, we utilize the quantization steps as an extra input to guide the network to handle different degrees of artifacts of various points. 
Furthermore, we incorporate a weighted graph attentional layer into the network to further pay more attention to points with larger distortions.
To the best of our knowledge, this is the first exploration of point cloud attribute artifacts removal for G-PCC. 
Experimental results demonstrate that our proposed MS-GAT increases the quality of the compressed point cloud attributes significantly.


\end{document}